\documentclass[10pt,twocolumn,letterpaper]{article}

\usepackage{wacv}              %

\usepackage{graphicx}
\usepackage{amsmath}
\usepackage{amssymb}
\usepackage{booktabs}

\usepackage[dvipsnames]{xcolor}
\usepackage{xcolor,colortbl}
\usepackage{soul}

\usepackage{comment}
\usepackage{multirow}

\usepackage{algorithm}
\usepackage{algorithmic}
\usepackage{tikz}
\usepackage{pgfplots}

\DeclareMathOperator*{\argmax}{argmax}

\usepackage{bm}

\usepackage{alphalph}

\usepackage{pgfplotstable}
\pgfplotsset{compat = newest}

\colorlet{mygold}{yellow!30}
\colorlet{mysilver}{black!10}

\colorlet{my00}{magenta}
\colorlet{my01}{orange}
\colorlet{my03}{ForestGreen}
\colorlet{my10}{cyan}
\colorlet{my30}{violet}

\usepackage[accsupp]{axessibility}

\usepackage[pagebackref,breaklinks,colorlinks]{hyperref}

\usepackage[capitalize]{cleveref}
\crefname{section}{Sec.}{Secs.}
\Crefname{section}{Section}{Sections}
\Crefname{table}{Table}{Tables}
\crefname{table}{Tab.}{Tabs.}

\def\wacvPaperID{61} %
\def\confName{WACV}
\def\confYear{2025}

\newcommand*\samethanks[1][\value{footnote}]{\footnotemark[#1]}

\begin{document}

\title{Counting Guidance for High Fidelity Text-to-Image Synthesis}

\author{
Wonjun Kang$^{1,2}$\thanks{Authors contribute equally.}
\qquad
Kevin Galim$^{2}$\samethanks
\qquad
Hyung Il Koo$^{2,3}$
\qquad
Nam Ik Cho$^{1}$
 \\
$^{1}$ Seoul National University \quad $^{2}$ FuriosaAI \quad $^{3}$ Ajou University \\
{\tt\small \{kangwj1995, kevin.galim, hikoo\}@furiosa.ai, nicho@snu.ac.kr}
}

\maketitle

\begin{abstract}
Recently, there have been significant improvements in the quality and performance of text-to-image generation, largely due to the impressive results attained by diffusion models. However, text-to-image diffusion models sometimes struggle to create high-fidelity content for the given input prompt. One specific issue is their difficulty in generating the precise number of objects specified in the text prompt. For example, when provided with the prompt “five apples and ten lemons on a table," images generated by diffusion models often contain an incorrect number of objects. In this paper, we present a method to improve diffusion models so that they accurately produce the correct object count based on the input prompt. We adopt a counting network that performs reference-less class-agnostic counting for any given image. We calculate the gradients of the counting network and refine the predicted noise for each step. To address the presence of multiple types of objects in the prompt, we utilize novel attention map guidance to obtain high-quality masks for each object. Finally, we guide the denoising process using the calculated gradients for each object. Through extensive experiments and evaluation, we demonstrate that the proposed method significantly enhances the fidelity of diffusion models with respect to object count. Code is available at \url{https://github.com/furiosa-ai/counting-guidance}.
\end{abstract}

\begin{figure*}[h]
\captionsetup[subfigure]{labelformat=empty}
\centering

\begin{minipage}{0.9\linewidth}
\centering
w/o counting guidance
\end{minipage}

\smallskip

\subfloat[]{
\includegraphics[width=.47\columnwidth]{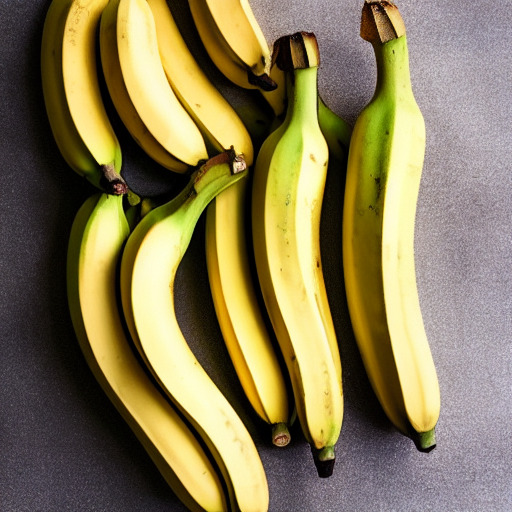}
\label{fig1a}
}
\subfloat[]{
\includegraphics[width=.47\columnwidth]{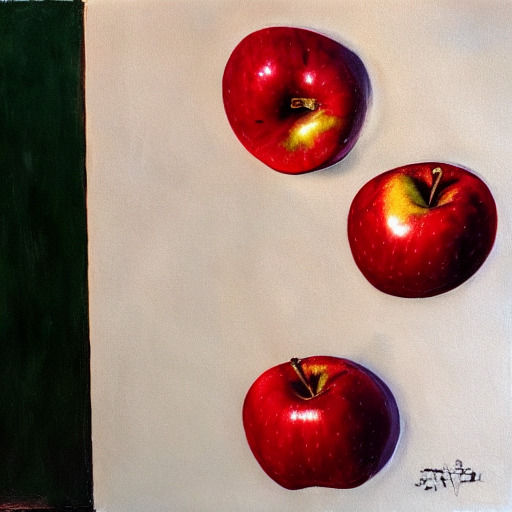}
\label{fig1b}
}
\subfloat[]{
\includegraphics[width=.47\columnwidth]{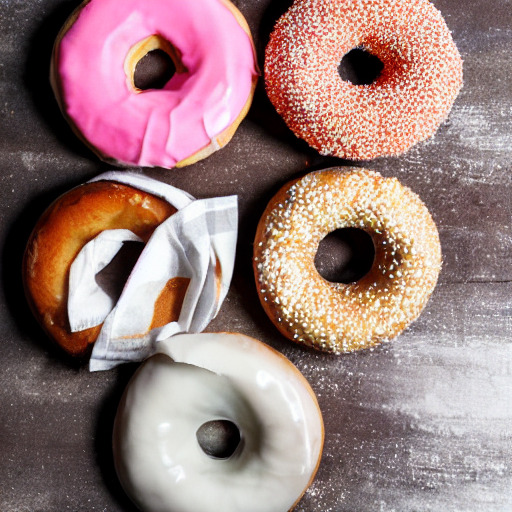}
\label{fig1c}
}
\subfloat[]{
\includegraphics[width=.47\columnwidth]{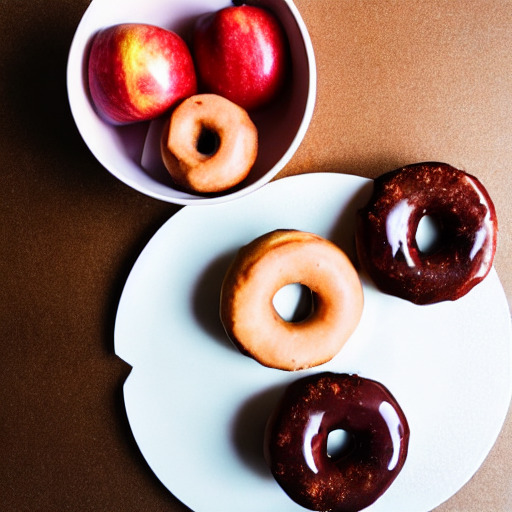}
\label{fig1d}
}

\begin{minipage}{0.9\linewidth}
\centering
w/ counting guidance
\end{minipage}

\smallskip

\subfloat[\textit{“six bananas”}]{
\includegraphics[width=.47\columnwidth]{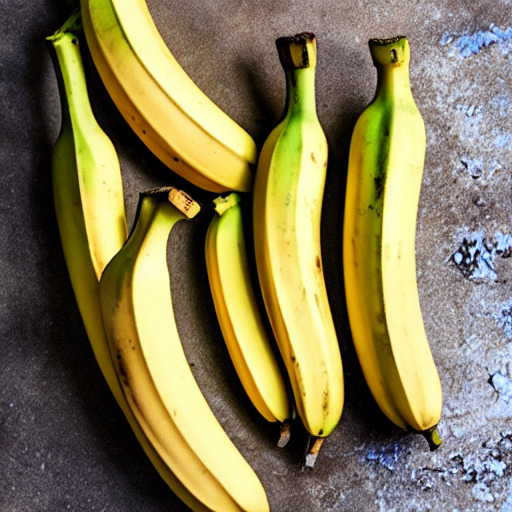}
\label{fig1e}
}
\subfloat[\textit{“five apples”}]{
\includegraphics[width=.47\columnwidth]{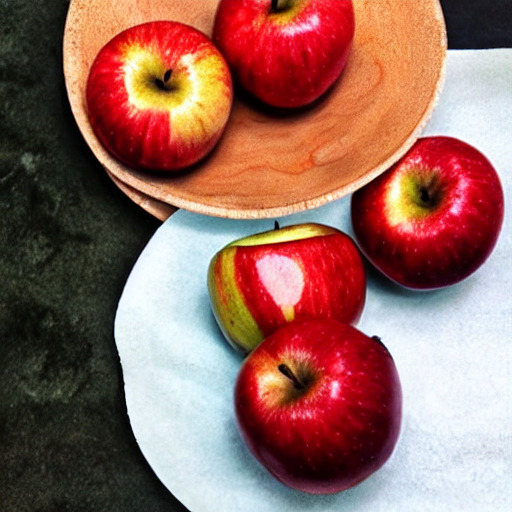}
\label{fig1f}
}
\subfloat[\textit{“four donuts”}]{
\includegraphics[width=.47\columnwidth]{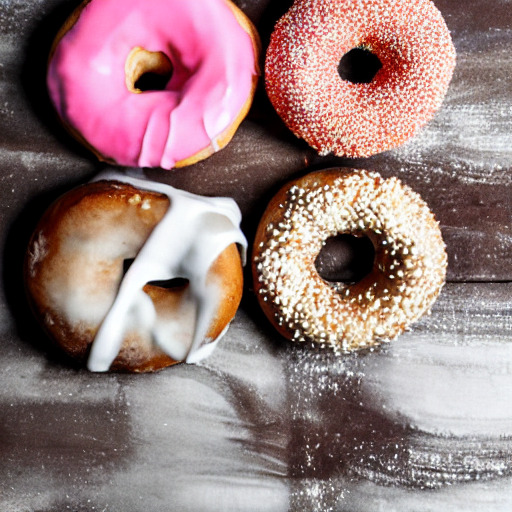}
\label{fig1g}
}
\subfloat[\textit{“three apples and two donuts”}]{
\includegraphics[width=.47\columnwidth]{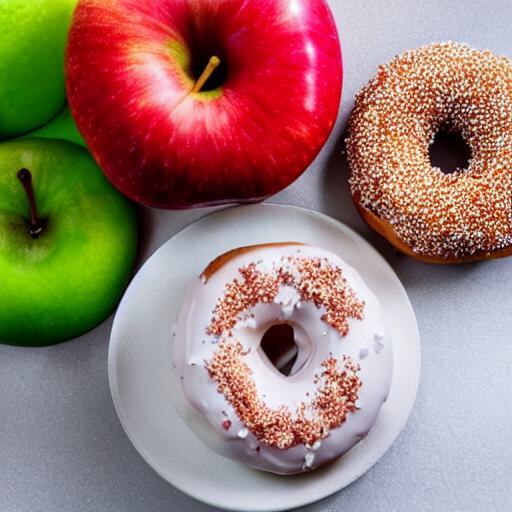}
\label{fig1h}
}

\caption{Counting guidance applied to Stable Diffusion \cite{rombach2022high}. Our proposed counting method generates the exact number of each object for a given prompt.}
\label{fig1}
\end{figure*}

\section{Introduction}
\label{sec:intro}
Text-to-image generation refers to the process of generating high-fidelity images based on a user-specified text prompt. This technology has various applications in digital art, design, and graphics. Traditionally, this was done using Generative Adversarial Networks (GANs) since the early days of deep learning \cite{goodfellow2014generative,karras2019style,karras2020analyzing,karras2021alias, zhang2017stackgan, zhang2018stackgan++, xu2018attngan, xia2021tedigan, patashnik2021styleclip}. However, GANs have limitations such as unstable training and lack of diversity (mode collapse), making them suitable only for generating images in specific domains like faces, animals, or vehicles. Recently, diffusion models \cite{ho2020denoising,song2019generative,song2020score}, a new family of generative models, have shown impressive, high-fidelity, and diverse results with stable training procedures, outperforming GANs, shifting the research focus from GANs to diffusion \cite{nichol2021glide,ramesh2022hierarchical,
saharia2022photorealistic,rombach2022high}. While many diffusion models have been proposed recently, the open source model Stable Diffusion \cite{rombach2022high}, a latent diffusion model trained on large datasets, has become the global standard for text-to-image generation models. Furthermore, Stable Diffusion, with its strong text-to-image generation capability, has been applied to various domains, including image editing \cite{mokady2023null,kang2024eta} and unified multimodal models \cite{ge2023making,sun2023generative,zeng2024can}.

However, there are still unresolved issues with diffusion models and Stable Diffusion. For example, Stable Diffusion sometimes shows poor performance for compositional text-to-image synthesis (e.g., \textit{“an apple and a lemon on the table”}), and various efforts have been made to resolve this problem. \cite{chefer2023attend} proposed Attend-and-Excite, which uses novel attention map guidance for generating two different objects. Several other studies used layout-based methods for compositional text-to-image synthesis \cite{li2023gligen,lian2023llm,phung2023grounded}. While there is considerable interest in compositional text-to-image synthesis, recent studies have focused on synthesizing one object of each kind. This has left the problem of synthesizing multiple instances of each object unsolved, for example, “three apples and five lemons on the table."

In this work, we focus on improving diffusion models to generate the exact number of instances per object, as specified in the input prompt. We propose counting guidance by using gradients of a counting network. Specifically, we use the counting model RCC \cite{hobley2022learning}, which performs reference-less class-agnostic counting for any given image. While most counting networks adopt a heatmap-based approach, RCC retrieves the object count directly via regression, allowing us to obtain its gradient for classifier guidance \cite{dhariwal2021diffusion,bansal2023universal}.

Furthermore, to handle multiple object types, we investigate the semantic information mixing problem of Stable Diffusion. For instance, the text prompt \textit{“three apples and four donuts on the table”} usually causes diffusion models to mix semantic information between apples and donuts leading to poor results and making it hard to enforce the correct object count per object type. We propose novel attention map guidance to separate semantic information between nouns in the prompt by obtaining masks for each object from the corresponding attention map. \cref{fig1} shows the effect of applying our method to Stable Diffusion for single and multiple object types. To the best of our knowledge, our work is the first attempt to generate the exact number of each object using a counting network for text-to-image synthesis. Our contributions can be summarized as follows:

\begin{itemize}
    \item We present counting network guidance to improve pre-trained diffusion models to generate the exact number of objects specified in the prompt. Our approach can be applied to any diffusion model and does not require retraining or finetuning.
    \item We propose novel attention map guidance to solve the semantic information mixing problem and obtain high-fidelity masks for each object.
    \item We demonstrate the effectiveness of our method by qualitative and quantitative comparisons with previous methods.

\end{itemize}

    \begin{figure*}[t]
\centering
\begin{minipage}{.49\textwidth}
\centering
\textit{“ten apples on the table”}
\end{minipage}
\begin{minipage}{.49\textwidth}
\centering
\textit{“fifty apples on the table”}
\end{minipage}

\medskip

\subfloat[w/o guidance ($N=3$)]{
\includegraphics[width=.47\columnwidth]{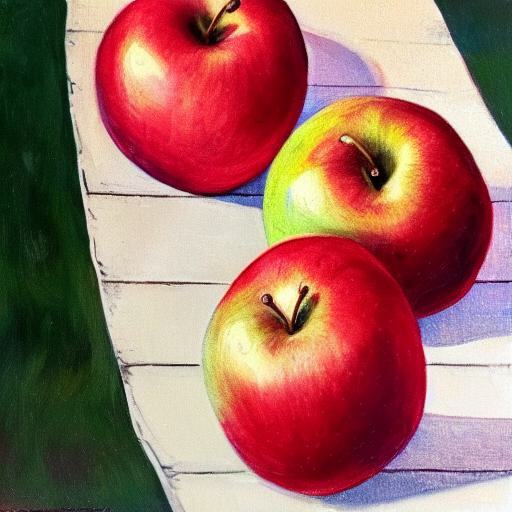}
\label{fig2a}
}
\subfloat[w/ guidance ($N=10$)]{
\includegraphics[width=.47\columnwidth]{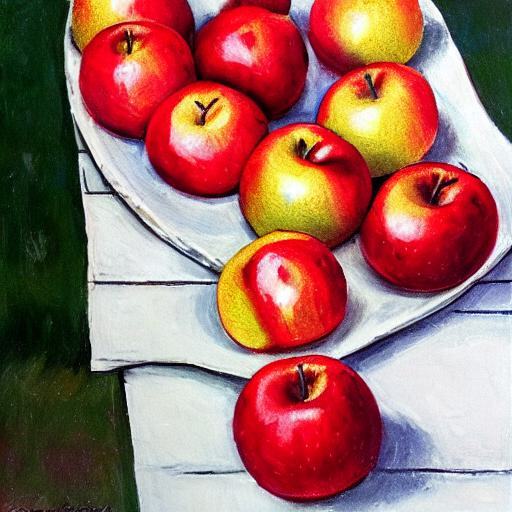}
\label{fig2b}
}
\vrule
\subfloat[w/o guidance ($N=18$)]{
\includegraphics[width=.47\columnwidth]{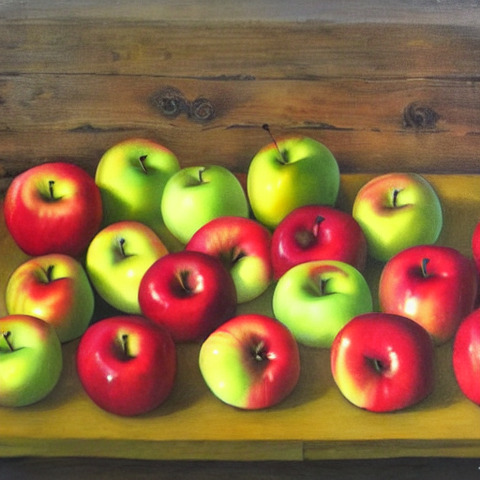}
\label{fig2c}
}
\subfloat[w/ guidance ($N=46$)]{
\includegraphics[width=.47\columnwidth]{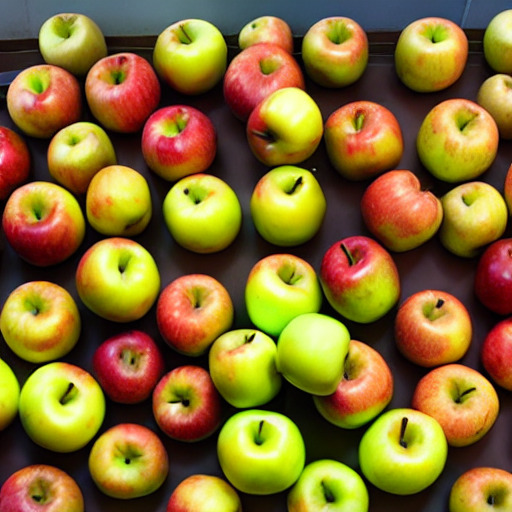}
\label{fig2d}
}
\caption{Effectiveness of counting network guidance. Our method is also effective for large numbers.}
\label{fig2}
\end{figure*}

\section{Related Work}
\label{sec:related}
\subsection{Diffusion Models}
Diffusion models \cite{ho2020denoising,song2019generative,song2020score,dhariwal2021diffusion,rombach2022high} are a new family of generative models that have significantly improved the performance of image synthesis and text-to-image generation. DDPM \cite{ho2020denoising} defined diffusion as a Markov chain process by gradually adding noise, showing the potential of diffusion models for unconditional image generation. Simultaneously, \cite{song2020score} interpreted diffusion models as Stochastic Differential Equations, providing broader insights into their function. One of the problems with DDPM is that it depends on probabilistic sampling and requires about 1,000 steps to obtain high-fidelity results, making the sampling process very slow and computationally intensive. To alleviate this problem, DDIM \cite{song2020denoising} removed the probabilistic factor in DDPM and achieved comparable image quality to DDPM with only 50 denoising steps.

Beyond unconditional image generation, recent papers on diffusion models also started to focus on conditional image generation. \cite{dhariwal2021diffusion} suggested classifier guidance by calculating the gradient of a classifier to perform conditional image generation. However, this method requires a noise-aware classifier and per-step gradient calculation. To avoid this problem, \cite{ho2021classifier} proposed classifier-free guidance, which removes the need for an external classifier by computing each denoising step as an extrapolation, requiring one conditional and one unconditional step. Furthermore, ControlNet \cite{zhang2023adding} proposed a separate control network attached to a pre-trained diffusion model to perform guidance with additional input with feasible training time. Universal Guidance \cite{bansal2023universal} alleviates the problem of requiring a noise-aware classifier by instead calculating the gradient of the predicted clean data point.

One issue of diffusion models is the high inference cost because of repeated inference in pixel-space. To address this problem, Stable Diffusion \cite{rombach2022high} proposed performing the diffusion process in a low dimensional latent space instead of image space, greatly reducing the computational cost.
Despite Stable Diffusion's powerful performance, there are still some remaining problems. For example, Stable Diffusion usually fails to generate multiple objects successfully (e.g., \textit{“an apple and a lemon on the table”}). Attend-and-Excite \cite{chefer2023attend} suggested attention map guidance to activate the attention of all objects in the prompt, but it only focused on a single instance per object, leaving the issue of reliably generating multiple instances per object. In this paper, we explicitly address this issue by introducing counting network guidance and attention map guidance to pre-trained diffusion models.

\cite{paiss2023teaching} and \cite{zhong2023adapter} proposed to generate the exact number of objects using enhanced language models. \cite{paiss2023teaching} trained a counting-aware CLIP model \cite{radford2021learning} and used it to fine-tune the text-to-image diffusion model Imagen \cite{saharia2022photorealistic}. \cite{lee2023aligning} and \cite{fan2023dpok} utilized human feedback to fine-tune text-to-image generation models by supervised learning and reinforcement learning. \cite{phung2023grounded} and \cite{lian2023llm} proposed layout-based text-to-image generation, which requires additional layout input and leverages a large language model (LLM) to generate proper layouts from given prompts. Unlike the above works, our method does not require additional layout input, an LLM, or retraining.

\subsection{Object Counting}
The goal of object counting is to count arbitrary objects in images. Object counting can be divided into few-shot object counting, reference-less counting, and zero-shot object counting. For few-shot object counting \cite{you2023few,shi2022represent}, a few example images of the object to count are provided as input. For reference-less counting \cite{ranjan2022exemplar,hobley2022learning}, example images are not provided and the aim is to count the number of all salient objects in the image. Zero-shot object counting \cite{xu2023zero, jiang2023clip} aims to count arbitrary objects of a user-provided class.

Object counting networks are usually either heatmap-based or regression-based \cite{you2023few,shi2022represent,hobley2022learning}. Since we require gradient calculation through the counting network, we adopt the model RCC \cite{hobley2022learning}, a reference-less regression-based counting model which builds on top of extracted features of a pre-trained ViT \cite{dosovitskiy2020image}.

\section{Preliminaries}
\label{sec:pre}
Denoising Diffusion Probabilistic Models (DDPM) \cite{ho2020denoising} define a forward noising process and a reverse denoising process, each with $T$ steps (e.g., $T=1000$). The forward process $q(x_{t}|x_{t-1})$ is defined as
\begin{equation}
q(x_{t}|x_{t-1})=\mathcal{N}(x_{t};\sqrt{\alpha_{t}}x_{t-1},(1-\alpha_{t})I),
\end{equation}
where $\alpha_{t}$ is the schedule and $x_{t}$ is the data point at time step $t$.
This process can be seen as iteratively adding scaled Gaussian noise.
Thanks to the property of the Gaussian distribution, we can obtain $q(x_{t}|x_{0})$ directly as
\begin{equation}
q(x_{t}|x_{0})=\mathcal{N}(x_{t};\sqrt{\bar{\alpha}_{t}}x_{0},(1-\bar{\alpha}_{t})I),
\end{equation}
and rewrite it as
\begin{equation}
x_{t}= \sqrt{\bar{\alpha}_{t}}x_{0} + \sqrt{1-\bar{\alpha}_{t}}\epsilon,
\end{equation}
where $\bar{\alpha}_{t}=\prod_{i=1}^{t}\alpha_{i}$ and $\epsilon\sim\mathcal{N}(0,I)$.
DDPM $\epsilon_{\theta}(x_{t},t)$ is trained to estimate the noise which was added in the forward process $\epsilon$ at each time step $t$. By iteratively estimating and removing the estimated noise, the original image can be recovered. During inference, images are generated using random noise as starting point.

In practice, however, deterministic DDIM \cite{song2020denoising} sampling is commonly used since it requires significantly fewer sampling steps compared to DDPM. DDIM sampling is performed as
\begin{equation}
x_{t-1}=\sqrt{\bar{\alpha}_{t-1}}(\frac{x_{t}-\sqrt{1-\bar{\alpha}_{t}}\epsilon_{\theta}}{\sqrt{\bar{\alpha}_{t}}})+\sqrt{1-\bar{\alpha}_{t-1}}\epsilon_{\theta}.
\end{equation}
With DDIM sampling, the clean data point $\hat{x}_{0}$ can be obtained by
\begin{equation}
\hat{x}_{0} = \frac{(x_{t}-\sqrt{1-\bar{\alpha}_{t}}\epsilon_{\theta}(x_{t},t))}{\sqrt{\bar{\alpha}_{t}}}.
\end{equation}
To add classifier guidance to DDIM \cite{dhariwal2021diffusion}, the gradient of a classifier is computed and used to retrieve the refined predicted noise $\hat{\epsilon}$ by
\begin{equation}
\hat{\epsilon} = \epsilon - s\sqrt{1-\bar{\alpha}_{t}}\nabla_{x_{t}}\log p_{\phi}(y|x_{t}),
\end{equation}
where $s$ is a scale parameter and $p_{\phi}$ is a classifier. One issue of classifier guidance is that the underlying classifier needs to be noise-aware as it receives outputs from intermediate denoising steps, requiring expensive noise-aware retraining. Universal Guidance \cite{bansal2023universal} addresses this by feeding the predicted clean data point $\hat{x}_{0}$ instead of the noisy $x_{t}$ to the classifier which can be expressed as
\begin{equation}
\hat{\epsilon} = \epsilon - s\sqrt{1-\bar{\alpha}_{t}}\nabla_{x_{t}}\log p_{\phi}(y|\hat{x}_{0}).
\end{equation}

    \begin{figure*}[t]
\captionsetup[subfigure]{labelformat=empty}
\centering
\begin{minipage}{.45\textwidth}
\centering
w/o attention map guidance
\end{minipage}

\smallskip

\subfloat[]{
\includegraphics[width=.38\columnwidth]{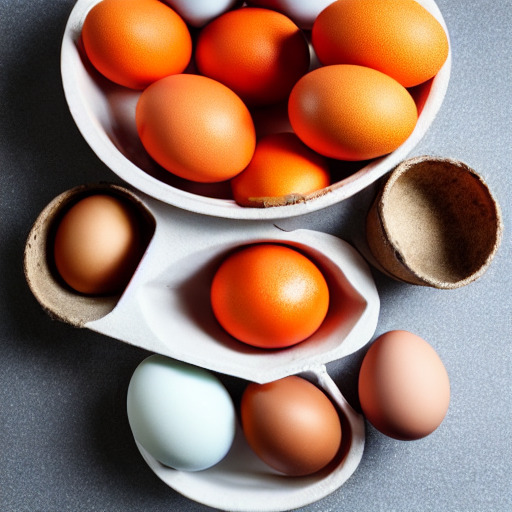}
\label{fig3a}
}
\subfloat[]{
\includegraphics[width=.38\columnwidth]{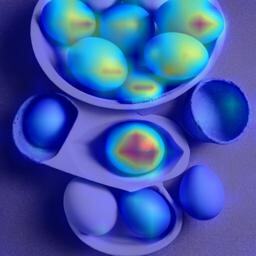}
\label{fig3b}
}
\subfloat[]{
\includegraphics[width=.38\columnwidth]{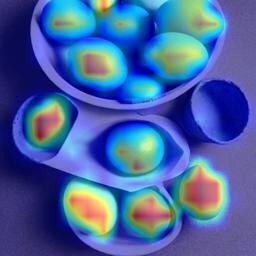}
\label{fig3c}
}
\subfloat[]{
\includegraphics[width=.38\columnwidth]{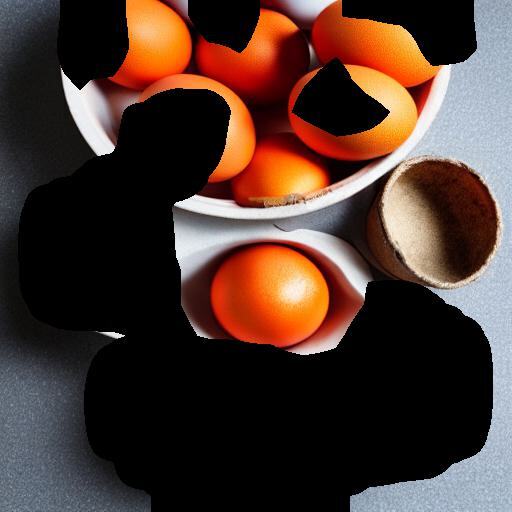}
\label{fig3d}
}
\subfloat[]{
\includegraphics[width=.38\columnwidth]{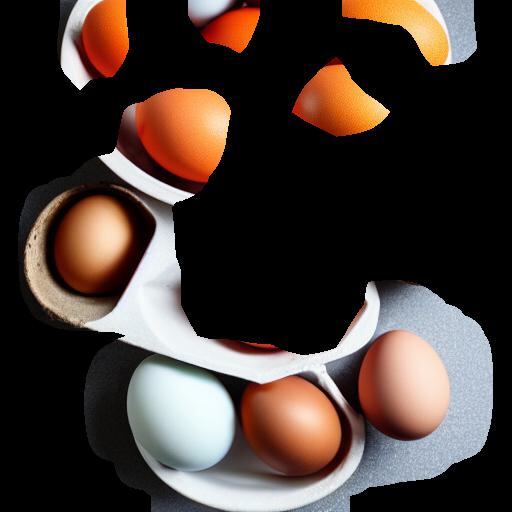}
\label{fig3e}
}

\begin{minipage}{.45\textwidth}
\centering
w/ attention map guidance
\end{minipage}

\smallskip

\subfloat[generated image]{
\includegraphics[width=.38\columnwidth]{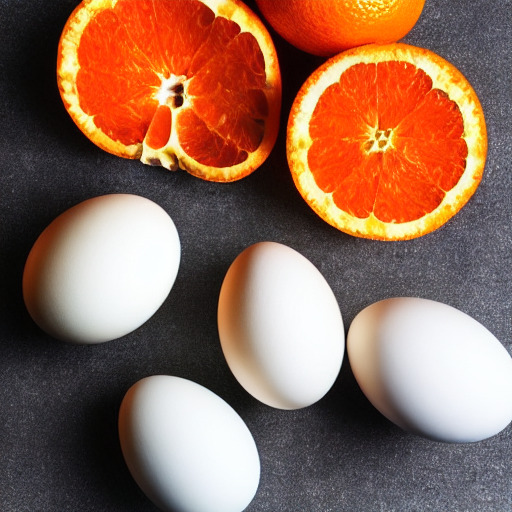}
\label{fig3f}
}
\subfloat[attention map of \textit{“oranges”}]{
\includegraphics[width=.38\columnwidth]{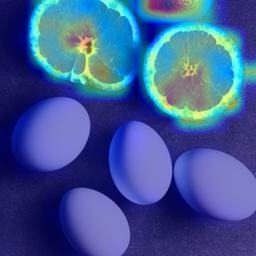}
\label{fig3g}
}
\subfloat[attention map of \textit{“eggs”}]{
\includegraphics[width=.38\columnwidth]{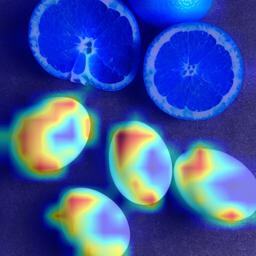}
\label{fig3h}
}
\subfloat[mask of \textit{“oranges”}]{
\includegraphics[width=.38\columnwidth]{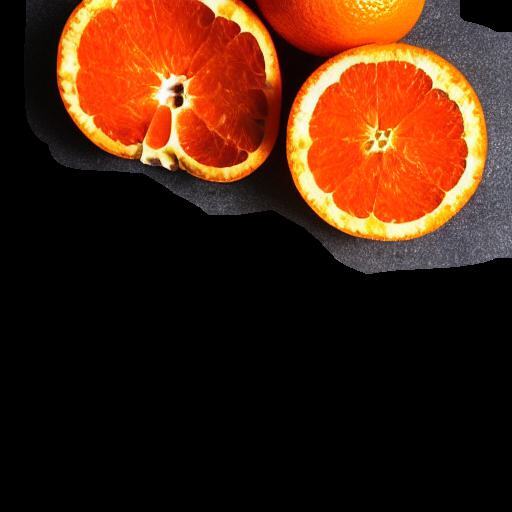}
\label{fig3i}
}
\subfloat[mask of \textit{“eggs”}]{
\includegraphics[width=.38\columnwidth]{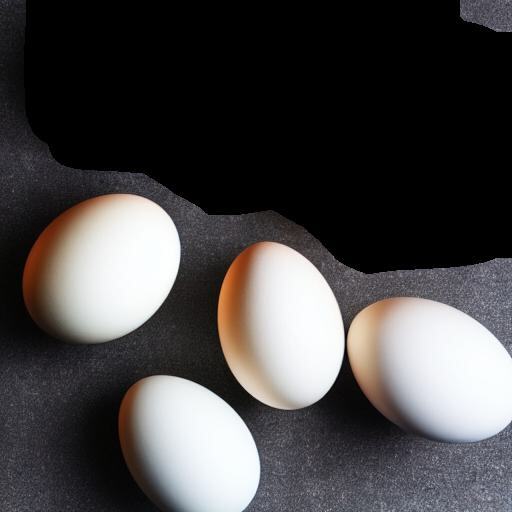}
\label{fig3j}
}
\caption{Effectiveness of attention map guidance for the prompt \textit{“three oranges and four eggs on the table.”} The first row shows the results of Stable Diffusion without attention map guidance, and the second row shows the results with attention map guidance.}
\label{fig3}
\end{figure*}

\section{Method}

\begin{algorithm}[tb]
\caption{Counting guidance for single object type}
\label{alg:algorithm1}
\textbf{Input}: time step $t$, denoising network $\epsilon_{\theta}(\cdot,\cdot)$, decoder $Decoder(\cdot)$, counting network $Count(\cdot)$, number of object $N$\\
\textbf{Parameter}: scale parameter $s_{count}$\\
\textbf{Output}: clean latent $z_{0}$
\begin{algorithmic}[1] %
\FOR{$t = T, T-1, ..., 1$}
\STATE $\epsilon\leftarrow \epsilon_{\theta}(z_{t},t)$
\STATE $\hat{z}_{0} \leftarrow (z_{t}-\sqrt{1-\bar{\alpha}_{t}}\epsilon)/\sqrt{\bar{\alpha}_{t}}$
\STATE $\hat{x}_{0} \leftarrow Decoder(\hat{z}_{0})$
\STATE $L_{count} \leftarrow |(Count(\hat{x}_{0})-N)/N|^{2}$
\STATE $\epsilon \leftarrow \epsilon+s_{count}\sqrt{1-\bar{\alpha}_{t}}\nabla_{z_{t}}L_{count}$
\STATE $z_{t-1} \leftarrow Sample(z_{t},\epsilon)$
\ENDFOR
\STATE \textbf{return} $z_{0}$
\end{algorithmic}
\end{algorithm}

\begin{algorithm}[tb]
\caption{Counting guidance for multiple object types}
\label{alg:algorithm2}
\textbf{Input}: time step $t$, denoising network $\epsilon_{\theta}$, decoder $Decoder$, counting network $Count$, number of $i$th object $N_{i}$\\
\textbf{Parameter}: scale parameter $s_{max}$, $s_{attention}$, $s_{count,i}$\\
\textbf{Output}: clean latent $z_{0}$
\begin{algorithmic}[1] %
\FOR{$t = T, T-1, ..., 1$}
\STATE $\epsilon, M \leftarrow \epsilon_{\theta}(z_{t},t)$
\STATE $L_{min} \leftarrow \sum_{j,k}\min_{i}({M_{i,j,k}})$
\STATE $L_{max} \leftarrow \sum_{j,k}\max_{i}({M_{i,j,k}})$
\STATE $L_{attention} \leftarrow L_{min}-s_{max}L_{max}$
\STATE $\epsilon \leftarrow \epsilon+s_{attention}\sqrt{1-\bar{\alpha}_{t}}\nabla_{z_{t}}L_{attention}$
\STATE $\hat{z}_{0} \leftarrow (z_{t}-\sqrt{1-\bar{\alpha}_{t}}\epsilon)/\sqrt{\bar{\alpha}_{t}}$
\STATE $\hat{x}_{0} \leftarrow Decoder(\hat{z}_{0})$
\FOR{i}
\STATE $\hat{x}_{0,i} \leftarrow Mask(\hat{x}_{0},M_{i})$
\STATE $L_{count, i} \leftarrow |(Count(\hat{x}_{0,i})-N_{i})/N_{i}|^{2}$
\STATE $\epsilon \leftarrow \epsilon+s_{count,i}\sqrt{1-\bar{\alpha}_{t}}\nabla_{z_{t}}L_{count,i}$
\ENDFOR
\STATE $z_{t-1} \leftarrow Sample(z_{t},\epsilon)$
\ENDFOR
\STATE \textbf{return} $z_{0}$
\end{algorithmic}
\end{algorithm}

In this section, we first demonstrate how to control the number of a single object type using counting network guidance and then expand this method to accommodate multiple object types. For multiple object types, we address the semantic information mixing problem of Stable Diffusion with attention map guidance and introduce masked counting network guidance for successful generation.

\subsection{Counting Guidance for a Single Object Type}

To avoid retraining the counting network on noisy images, we perform counting network guidance following Universal Guidance \cite{bansal2023universal}. For a given number of $N$ objects, we define the counting loss $L_{count}$ as
\begin{equation}
L_{count} = \left| \frac{Count(\hat{x}_{0})-N}{N} \right|^{2},
\end{equation}
where $Count(\cdot)$ is the pre-trained counting network RCC \cite{hobley2022learning} and $\hat{x}_{0}$ is the predicted clean image at each time step. We update the predicted noise $\epsilon$ using the gradient of the counting network as
\begin{equation}
\epsilon \leftarrow \epsilon+s_{count}\sqrt{1-\bar{\alpha}_{t}}\nabla_{z_{t}}L_{count},
\end{equation}
where $s_{count}$ is an additional scale parameter to control the strength of counting guidance.

\cref{fig2a} and \cref{fig2b} show the effectiveness of our proposed counting network guidance method. For the prompt \textit{“ten apples on the table,”} Stable Diffusion with counting
network guidance generates ten apples, while vanilla Stable Diffusion generates only three apples. We find that \cref{fig2a} and \cref{fig2b} have similar textures and backgrounds, indicating that counting guidance maintains the original properties of Stable Diffusion while only influencing the object count.

Counting guidance is also effective for generating a large number of objects. Due to a lack of images containing a large number of objects in Stable Diffusion's training dataset, it often fails to create plausible results for such cases.
\cref{fig2c} and \cref{fig2d} show the effectiveness of counting guidance  on large numbers. For the given text prompt \textit{“fifty apples on the table,”} Stable Diffusion with counting network guidance generates 46 apples, while vanilla Stable Diffusion generates only 18 apples.

\subsection{Counting Guidance for Multiple Object Types}

\subsubsection{Semantic Information Mixing Problem}

When dealing with multiple object classes, it is important to count each class individually. While a class-aware counting network could be used, the clean image predicted during the early denoising steps is of too low quality for the counting network to accurately identify each object instance. Hence, we have chosen to use a class-agnostic counting network instead. For each object type to count, we obtain a mask using the underlying self-attention maps of Stable Diffusion's UNet model similar to \cite{hertz2022prompt, chefer2023attend,kang2024eta} and feed the masked image of each object type to the counting network separately.

\subsubsection{Attention Map Guidance}

We have noticed that Stable Diffusion often tends to produce attention maps that do not accurately correspond to the correct location of each object. The first row of \cref{fig3} demonstrates this semantic information mixing problem. For the prompt \textit{“three oranges and four eggs on the table,”} we find that the attention map of \textit{“oranges”} and the attention map of \textit{“eggs”} share a large part of pixels resulting in the generation of orange-colored eggs instead of oranges and eggs. To solve the semantic information mixing problem, we first obtain each object's attention map following \cite{chefer2023attend}. Similarly, we exclude the $\langle sot \rangle$ token, re-weigh using Softmax, and then Gaussian-smooth to receive the attention map $M_{i}$ for each object $i$. Finally, we normalize each object's attention map as 
\begin{equation}
{\hat{M}_{i,j,k}} = \frac{{M_{i,j,k}} - \min_{j,k}({M_{i,j,k}})}{\max_{j,k}({M_{i,j,k}}) - \min_{j,k}({M_{i,j,k}})},
\end{equation}
where $M_{i,j,k}$ is the attention value of coordinate $(j,k)$ of object $i$'s attention map.

We then ensure that each pixel coordinate is only referred to by the attention of a single object by calculating each coordinate's minimum attention value and summate them to $L_{min}$ where a low $L_{min}$ indicates that each coordinate is only activated by a single object: 
\begin{equation}
L_{min} = \sum_{j,k}\min_{i}({\hat{M}_{i,j,k}}).
\end{equation}
Similar to $L_{min}$, we define $L_{max}$ to ensure that at least one object activates each pixel as
\begin{equation}
L_{max} = \sum_{j,k}\max_{i}({\hat{M}_{i,j,k}}).
\end{equation}
Finally, we calculate the total attention loss $L_{attention}$ as
\begin{equation}
L_{attention} = L_{min}-s_{max}L_{max},
\end{equation}
where $s_{max}$ is a scale parameter. The predicted noise $\epsilon$ is then updated as
\begin{equation}
\epsilon \leftarrow \epsilon+s_{attention}\sqrt{1-\bar{\alpha}_{t}}\nabla_{z_{t}}L_{attention}.
\end{equation}  %

The second row of \cref{fig3} demonstrates the effectiveness of our attention map guidance. We find that the attention map for \textit{“oranges”} focuses solely on oranges, and the attention map for \textit{“eggs”} focuses solely on eggs, resulting in a correctly synthesized output. Moreover, we observe that high-fidelity object masks are generated from the corresponding attention maps.

\subsubsection{Masked Counting Guidance}

For each object $i$, we binarize its attention map to receive the binary mask $M^b_{i}$ as 
\begin{equation}
M^b_{i,j,k} = 
    \begin{cases}
          1, & \textrm{if}\ i = \argmax_{i}(M_{i,j,k}) \\
          0, & \textrm{otherwise}
    \end{cases}
\end{equation}
and then generate a masked clean image $\hat{x}_{0,i}$ using element-wise multiplication:
\begin{equation}
\hat{x}_{0,i} = \hat{x}_{0} \odot M^b_{i}.
\end{equation}  %
For the $i$-th object count of  object $N_{i}$, each masked counting guidance $L_{count,i}$ is defined as
\begin{equation}
L_{count,i} = \left| \frac{Count(\hat{x}_{0,i})-N_{i}}{N_{i}} \right|^{2}.
\end{equation}
Finally, we update the noise $\epsilon$ as
\begin{equation}
\epsilon \leftarrow \epsilon+\sum_{i} s_{count,i}\sqrt{1-\bar{\alpha}_{t}}\nabla_{z_{t}}L_{count,i},
\end{equation}
where $s_{count,i}$ is an additional scaling parameter per object.

    \begin{figure*}[h]
\captionsetup[subfigure]{labelformat=empty}
\centering

\begin{minipage}{.5\textwidth}
\centering
Stable Diffusion \cite{rombach2022high}
\end{minipage}

\smallskip

\subfloat[]{
\includegraphics[width=.31\columnwidth]{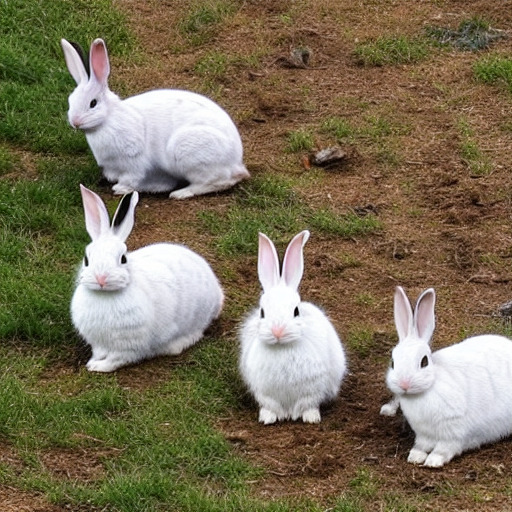}
\label{fig4a}
}
\subfloat[]{
\includegraphics[width=.31\columnwidth]{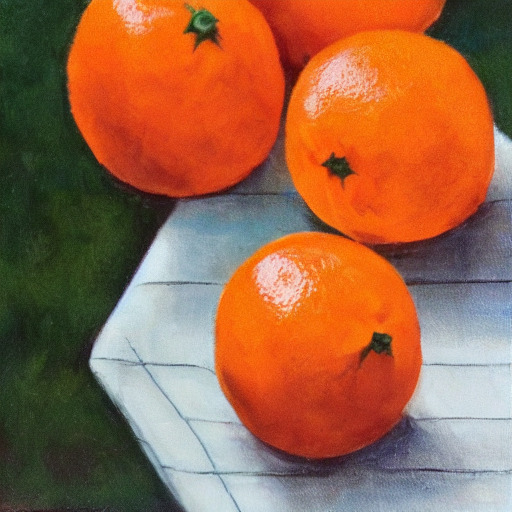}
\label{fig4b}
}
\subfloat[]{
\includegraphics[width=.31\columnwidth]{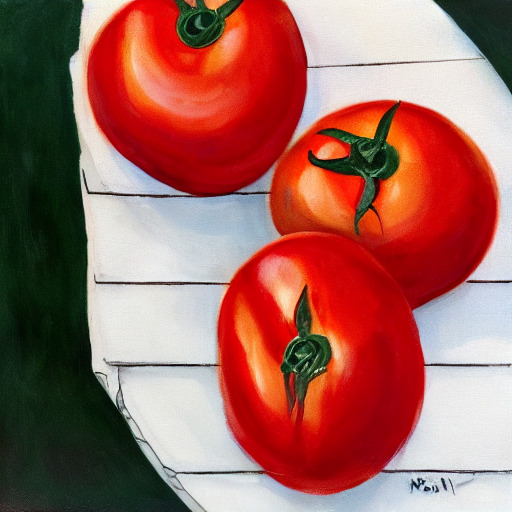}
\label{fig4c}
}
\subfloat[]{
\includegraphics[width=.31\columnwidth]{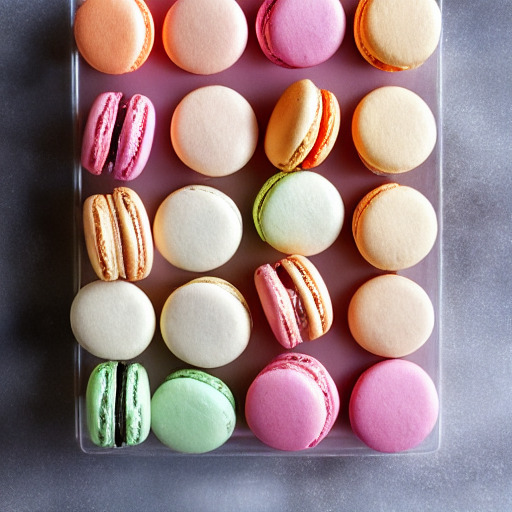}
\label{fig4d}
}
\subfloat[]{
\includegraphics[width=.31\columnwidth]{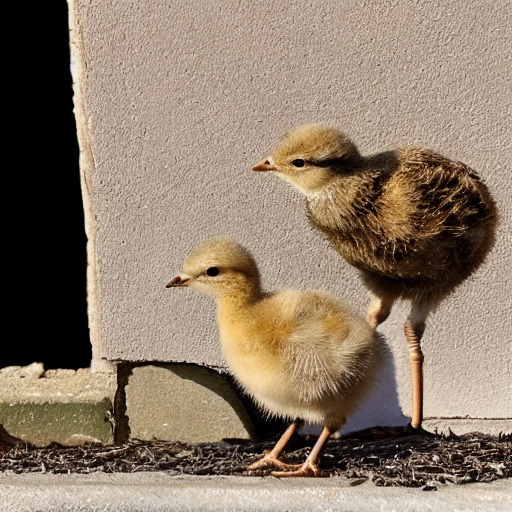}
\label{fig4e}
}
\subfloat[]{
\includegraphics[width=.31\columnwidth]{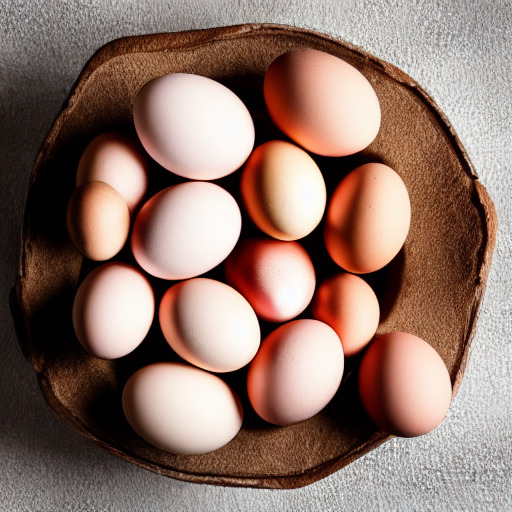}
\label{fig4f}
}

\begin{minipage}{.5\textwidth}
\centering
Attend-and-Excite \cite{chefer2023attend}
\end{minipage}

\smallskip

\subfloat[]{
\includegraphics[width=.31\columnwidth]{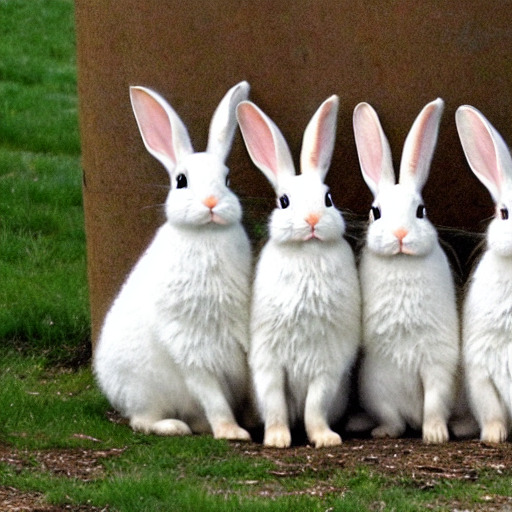}
\label{fig4g}
}
\subfloat[]{
\includegraphics[width=.31\columnwidth]{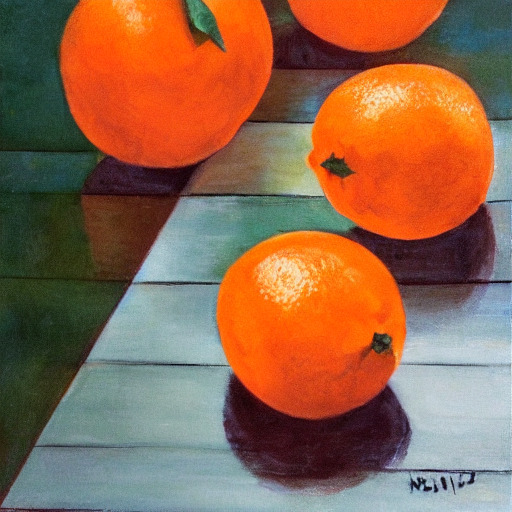}
\label{fig4h}
}
\subfloat[]{
\includegraphics[width=.31\columnwidth]{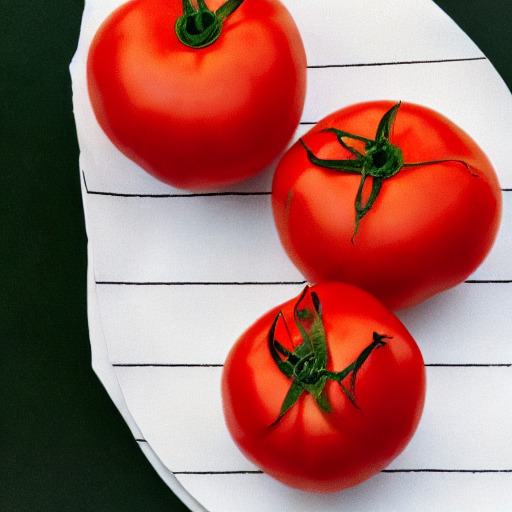}
\label{fig4i}
}
\subfloat[]{
\includegraphics[width=.31\columnwidth]{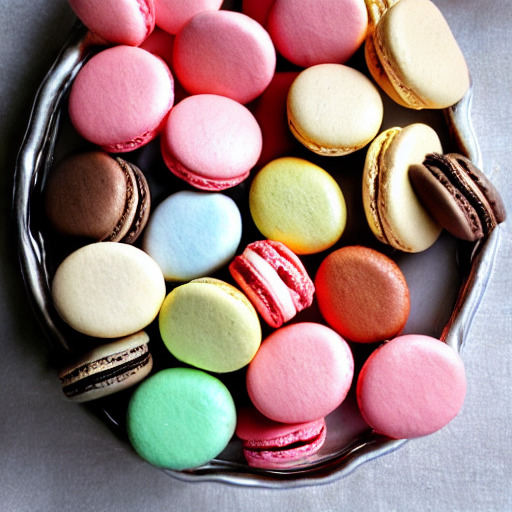}
\label{fig4j}
}
\subfloat[]{
\includegraphics[width=.31\columnwidth]{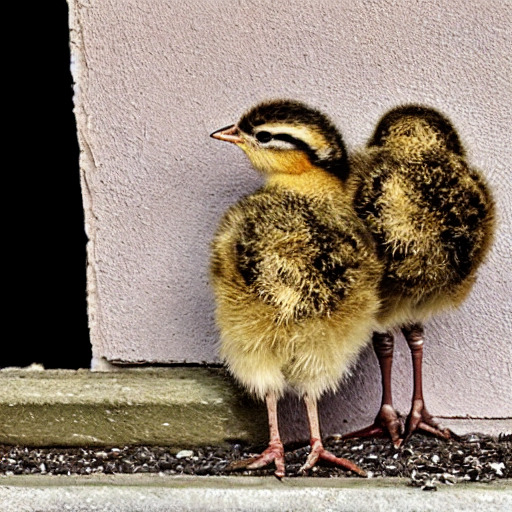}
\label{fig4k}
}
\subfloat[]{
\includegraphics[width=.31\columnwidth]{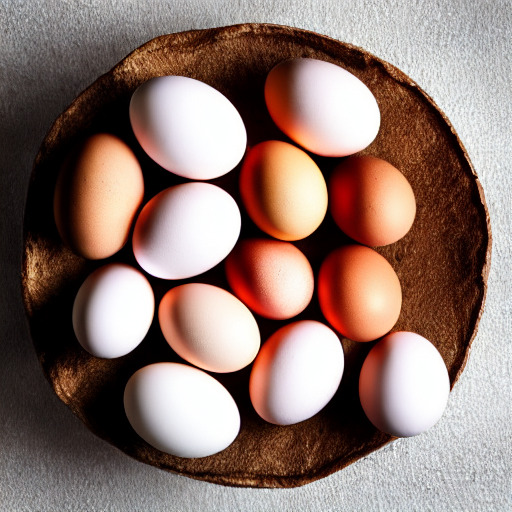}
\label{fig4l}
}

\begin{minipage}{.5\textwidth}
\centering
Ours
\end{minipage}

\smallskip

\subfloat[\textit{“five rabbits”}]{
\includegraphics[width=.31\columnwidth]{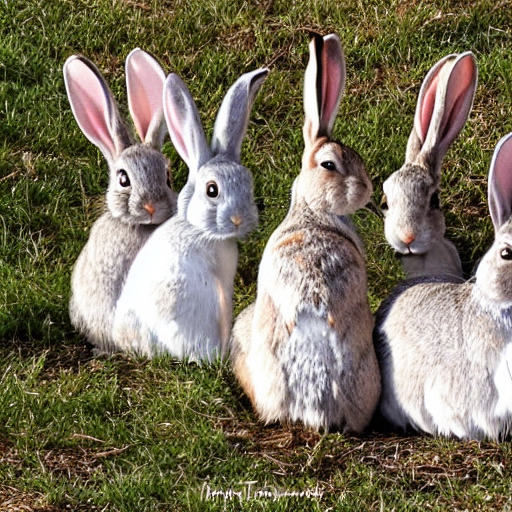}
\label{fig4m}
}
\subfloat[\textit{“ten oranges”}]{
\includegraphics[width=.31\columnwidth]{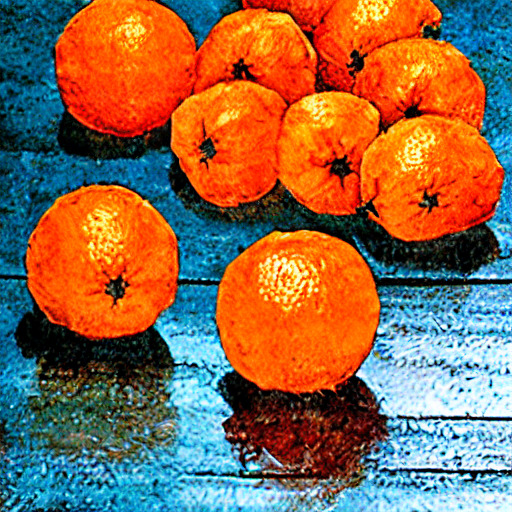}
\label{fig4n}
}
\subfloat[\textit{“four tomatoes”}]{
\includegraphics[width=.31\columnwidth]{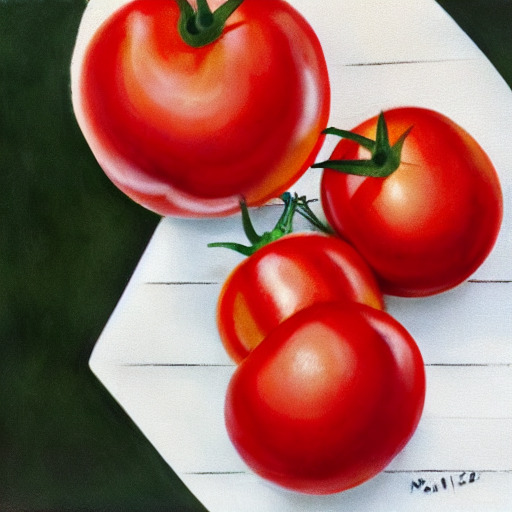}
\label{fig4o}
}
\subfloat[\textit{“twelve macarons”}]{
\includegraphics[width=.31\columnwidth]{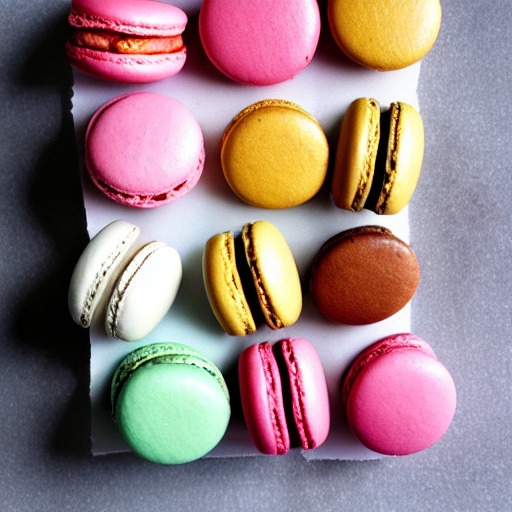}
\label{fig4p}
}
\subfloat[\textit{“three chicks”}]{
\includegraphics[width=.31\columnwidth]{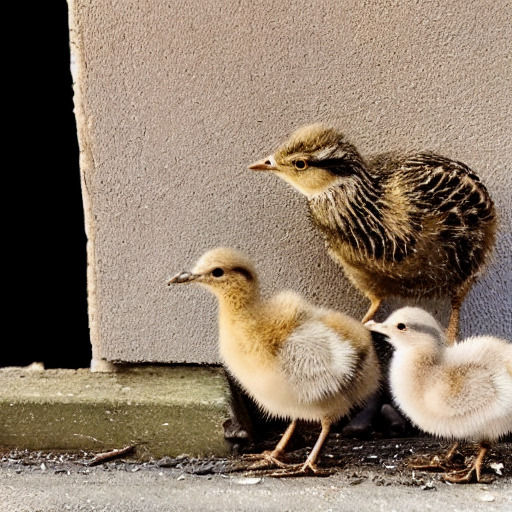}
\label{fig4q}
}
\subfloat[\textit{“seven eggs”}]{
\includegraphics[width=.31\columnwidth]{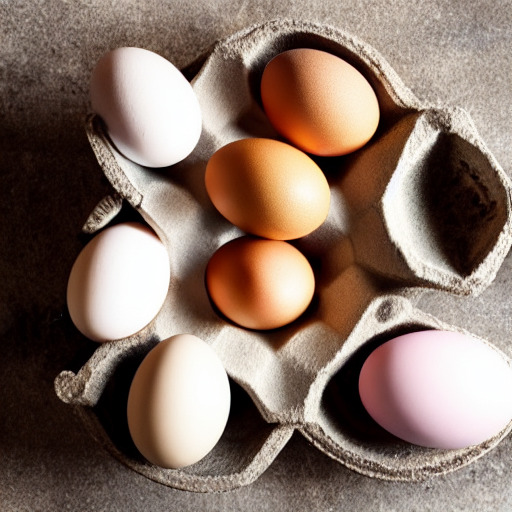}
\label{fig4r}
}
\caption{Qualitative comparison for single object type. The first row shows the results of Stable Diffusion \cite{rombach2022high}, the second row shows the results of Attend-and-Excite \cite{chefer2023attend} and the last row shows the results of our method.}
\label{fig4}
\end{figure*}

\section{Experiments}
We borrow the state-of-the-art text-to-image generation model Stable Diffusion (v1.4 and v2.1) for our experiments. We use DDIM sampling with 50 steps and set the scale parameter for $L_{max}$ to $s_{max}=0.1$ by default. We create a modified dataset based on the object classes from Attend-and-Excite \cite{chefer2023attend} to evaluate and compare our approach with previous methods. Specifically, we remove the color category and add more animals and objects for a total of 34 object classes. We compare our method with Stable Diffusion \cite{rombach2022high}, Attend-and-Excite \cite{chefer2023attend}, and SUR-Adpater \cite{zhong2023adapter}.

\subsection{Quantitative Results}

For quantitative comparison, we count the number of given objects using the object detection network Grounding DINO \cite{liu2023grounding}. We create a dataset of 680 prompts using our 34 predefined object classes with counts ranging from 1-20 (e.g., \textit{“ten apples”}) and measure the normalized MAE (Mean Absolute Error) and RMSE (Root Mean Squared Error). In our evaluation of $s_{count}$, we explored both constant and linearly scheduled approaches. For the constant scenario, we fixed $s_{count} = 1$. However, when implementing a linear schedule, we discovered that $s_{count} = \max(0.01, 0.2N - 1)$ resulted in markedly improved performance. This formulation allows $s_{count}$ to increase incrementally with $N$, providing a more dynamic adjustment compared to the static nature of the constant value (see supplementary materials for detailed hyperparameter analysis).

\cref{tab1} presents a detailed quantitative comparison of counting performance. Our method (linear) achieves the best scores for both MAE and RMSE while maintaining comparable or better CLIP similarity to vanilla Stable Diffusion (\cref{tab1_a,tab1_d}). Our method (constant) achieves the second-best score for both MAE and RMSE, demonstrating the effectiveness of our method with fixed $s_{count}$. For the user study, we conducted 330 comparisons on our dataset. In non-tie cases, our method is preferred about 1.9 times more than vanilla Stable Diffusion (\cref{tab1_b}).

Despite our method demonstrating superior performance across various metrics, CLIP alone is insufficient to fully reflect image quality, and user studies lack scalability. To address these issues, we incorporate GPT-4V~\cite{openai2024gpt4technicalreport} evaluation to further validate the effectiveness of our approach (as shown in \cref{tab1_b}). The results indicate that GPT also favors our method over Stable Diffusion, reinforcing the advantages of our strategy.

We also show the effectiveness of our attention map guidance by evaluating text-image and text-text CLIP \cite{radford2021learning} similarities. We generate 1122 multiple object prompts using our 34 object classes by combining two object classes with a random count for each prompt (e.g., \textit{“eight lemons and seventeen onions”}). We measure text-image CLIP similarities for all prompts and text-text CLIP similarities for generated captions by BLIP \cite{li2022blip} following \cite{chefer2023attend}. We fix the scale parameter to $s_{attention}=1$. \cref{tab1_c} presents the quantitative results for both metrics. Attend-and-Excite achieves the best text-image similarity, while our method achieves the best text-text similarity.

\subsection{Qualitative Results}

\paragraph{Results for Single Object Type}

\begin{figure}[t]
\captionsetup[subfigure]{labelformat=empty}
\centering
\begin{minipage}{.45\textwidth}
\centering
\textit{“two apples and three donuts on the table”}
\end{minipage}

\smallskip

\subfloat[]{
\includegraphics[width=.3\columnwidth]{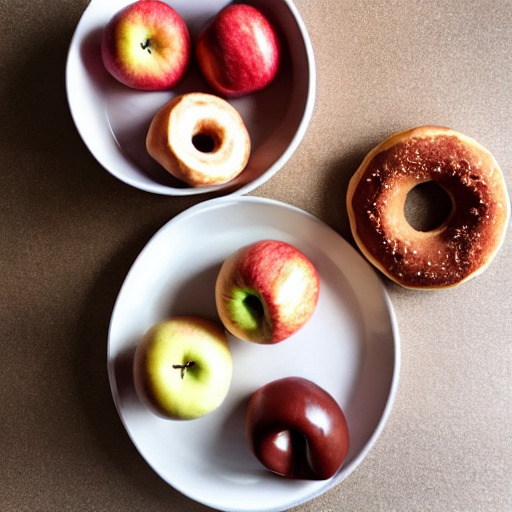}
\label{fig5a}
}
\subfloat[]{
\includegraphics[width=.3\columnwidth]{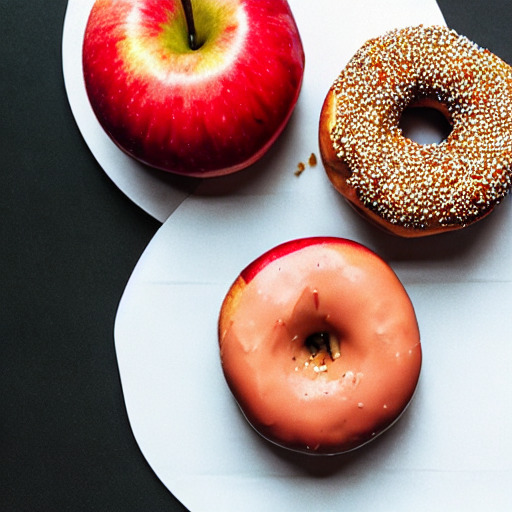}
\label{fig5b}
}
\subfloat[]{
\includegraphics[width=.3\columnwidth]{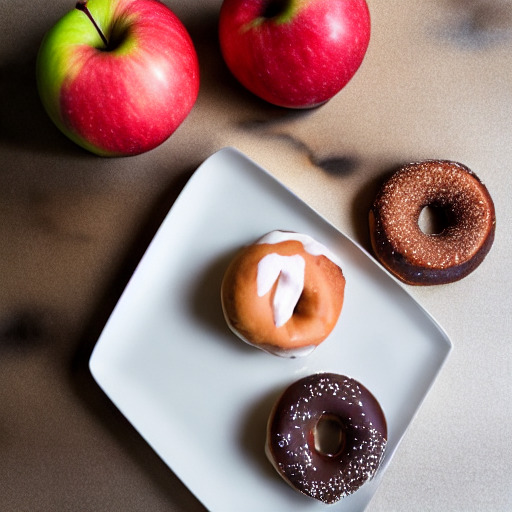}
\label{fig5c}
}

\begin{minipage}{.45\textwidth}
\centering
\textit{“three lemons and one bread on the table”}
\end{minipage}

\smallskip

\subfloat[]{
\includegraphics[width=.3\columnwidth]{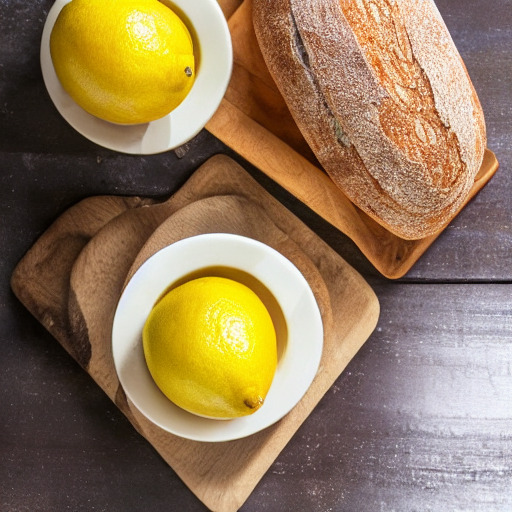}
\label{fig5d}
}
\subfloat[]{
\includegraphics[width=.3\columnwidth]{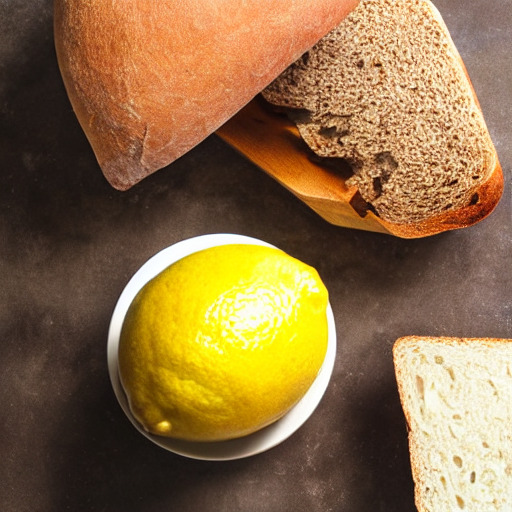}
\label{fig5e}
}
\subfloat[]{
\includegraphics[width=.3\columnwidth]{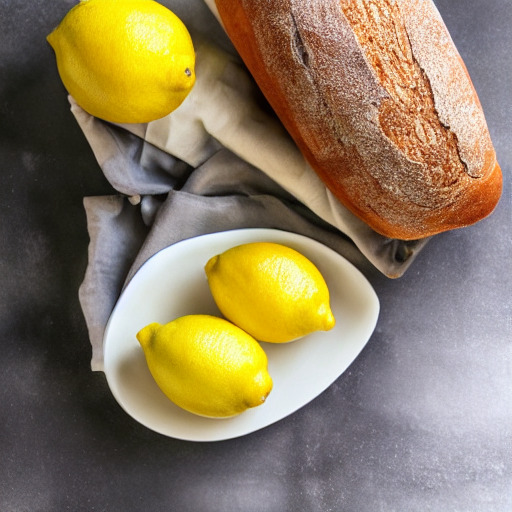}
\label{fig5f}
}

\begin{minipage}{.45\textwidth}
\centering
\textit{“two onions and two tomatoes on the table”}
\end{minipage}

\smallskip

\subfloat[Stable Diffusion]{
\includegraphics[width=.3\columnwidth]{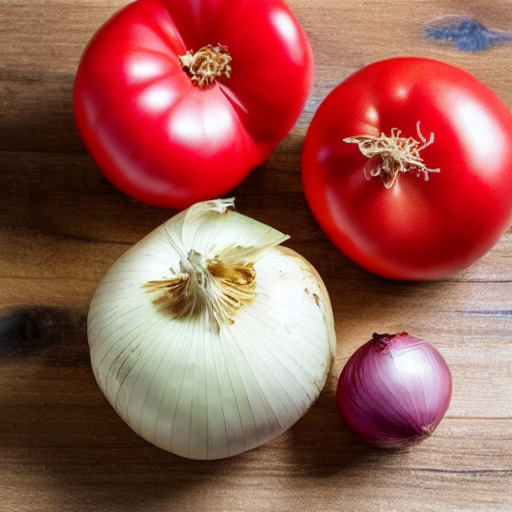}
\label{fig5g}
}
\subfloat[Attend-and-Excite]{
\includegraphics[width=.3\columnwidth]{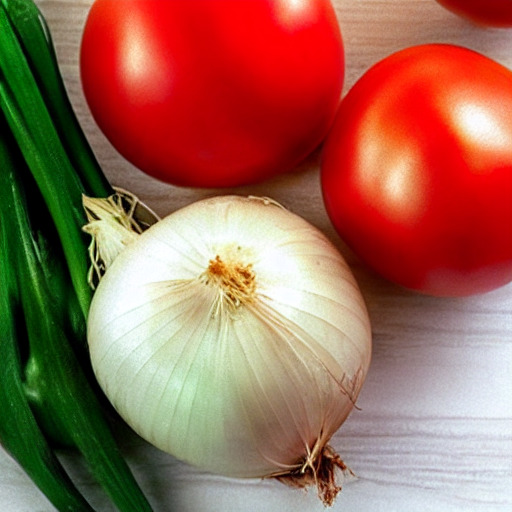}
\label{fig5h}
}
\subfloat[Ours]{
\includegraphics[width=.3\columnwidth]{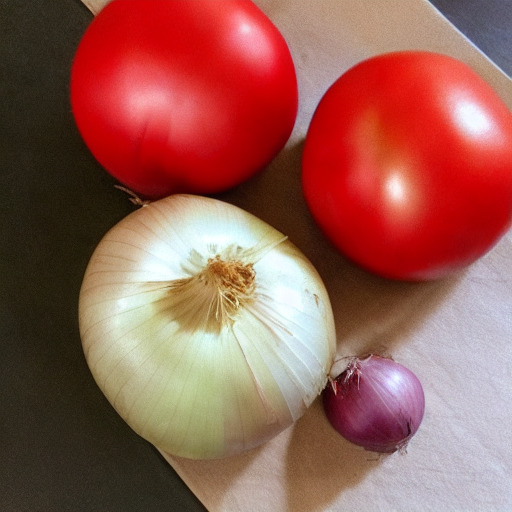}
\label{fig5i}
}
\caption{Qualitative comparison for multiple object types. The first column shows the results of Stable Diffusion, the second column shows the results of Attend-and Excite, and the last column shows the results of our method.}
\label{fig5}
\end{figure}

\cref{fig4} shows a qualitative comparison for the single object type scenario. While Stable Diffusion and Attend-and-Excite fail to generate the right number of objects as specified in the prompt, our method generates the correct number. For the prompt \textit{“four tomatoes on the table,”} Stable Diffusion generates only three tomatoes without counting guidance. With counting guidance, the tomato at the bottom is successfully divided into two tomatoes, while the rest of the image is consistent with the original result. The text prompt \textit{“ten oranges on the table,”} causes Stable Diffusion to only generate four oranges compared to our solution that creates the correct amount of ten. The big difference in object count between Stable Diffusion and the target prompt causes large gradients, making our result severely differ from the original.

Our method also works well for more complex categories, such as animals. Considering the prompt \textit{“three chicks on the road”}, Stable Diffusion and Attend-and-Excite synthesize only two chicks, unlike our method which generates one additional chick while maintaining the other two chicks' appearance. For the prompt \textit{“five rabbits on the yard”} Stable Diffusion and Attend-and-Excite generate only four rabbits, while our method generates one more rabbit but fails to maintain the other rabbits' appearance. That is because of the difference between the background and the rabbit colors. It is hard to generate a white rabbit from a brown yard, so Stable Diffusion with counting guidance changes the overall structure and recreates five rabbits.

\paragraph{Results for Multiple Object Types}

\cref{fig5} shows a qualitative comparison for multiple object types. For \textit{“three lemons and one bread on the table”}, Stable Diffusion successfully generates one bread but fails with three lemons, while Attend-and-Excite fails in both cases. With masked counting guidance, our method correctly generates three lemons and one bread. The result shows that the lemon at the bottom is divided into two lemons thanks to masked counting guidance while maintaining the bread's shape.

For \textit{“two onions and two tomatoes on the table”}, Stable Diffusion suffers from the semantic information mixing problem and generates red onions instead of tomatoes. Due to our attention map guidance, our method creates realistic tomatoes. As Attend-and-Excite is also based on attention map optimization, it successfully generates realistic tomatoes but fails to generate the exact number of onions.

\begin{table}[t]
        \centering
        \scriptsize
        \centering
        \setlength{\tabcolsep}{6pt}
        \begin{subtable}[t]{0.47\textwidth}
        \centering
\begin{tabular}{ll|ccc}
        \toprule
        Baseline & Method & MAE $\downarrow$ & RMSE $\downarrow$ & CLIP $\uparrow$ \\
        \midrule
        \multirow{5}{*}{Stable Diffusion}&Vanilla & 0.599 & 0.746 & \cellcolor{mygold}0.316 \\
        &Attend-and-Excite & 0.601 & 0.709 & 0.313 \\
        &SUR-Adapter & 0.903 & 0.924 & 0.236 \\
        &Ours (constant) & \cellcolor{mysilver}0.585 & \cellcolor{mysilver}0.696 & 0.311 \\
        &\textbf{Ours (linear)}& \cellcolor{mygold}0.567 & \cellcolor{mygold}0.692 & \cellcolor{mysilver}0.315 \\
         \bottomrule
        \end{tabular}
                \caption{\textbf{Counting error and CLIP similarity.} Tested with Stable Diffusion.}
                        \label{tab1_a}
                \end{subtable}
                
                \medskip
        \begin{subtable}[t]{0.47\textwidth}
        \centering
\begin{tabular}{ll|ccc}
        \toprule
        Baseline & Evaluation &Tie & Vanilla & \textbf{Ours (linear)} \\
        \midrule
        \multirow{2}{*}{Stable Diffusion}&User study & 64.9\% & 12.1\% & 23.0\% \\
        &GPT evaluation & 38.5\% & 26.2\% & 35.3\% \\
         \bottomrule
        \end{tabular}
                \caption{\textbf{User study and GPT evaluation.} Tested with Stable Diffusion.}
                        \label{tab1_b}
                \end{subtable}
   
                \medskip
        \begin{subtable}[t]{0.47\textwidth}
        \centering
\begin{tabular}{ll|cc}
        \toprule
        Baseline &Method & Text-Image $\uparrow$ & Text-Caption $\uparrow$ \\
        \midrule
        \multirow{4}{*}{Stable Diffusion}& Vanilla & 0.324 & 0.722 \\
        &Attend-and-Excite & \cellcolor{mygold}0.330 & \cellcolor{mysilver}0.731 \\
        &SUR-Adpater & 0.238 & 0.563 \\
        &\textbf{Ours} & \cellcolor{mysilver}0.329 & \cellcolor{mygold}0.732 \\
         \bottomrule
        \end{tabular}
                \caption{\textbf{Effectiveness of attention map guidance.} Tested with Stable Diffusion.}
                        \label{tab1_c}
                \end{subtable}
   
                \medskip
        \begin{subtable}[t]{0.47\textwidth}
        \centering
\begin{tabular}{ll|ccc}
        \toprule
        Baseline &Method & MAE $\downarrow$ & RMSE $\downarrow$ & CLIP $\uparrow$ \\
        \midrule
        \multirow{2}{*}{Stable Diffusion 2}& Vanilla & 0.473 & 0.607 & 0.324 \\
        &\textbf{Ours (linear)} & \cellcolor{mygold}0.461 & \cellcolor{mygold}0.593 & \cellcolor{mygold}0.326 \\
         \bottomrule
        \end{tabular}
                \caption{\textbf{Counting error and CLIP similarity.} Tested with Stable Diffusion~2.}
                        \label{tab1_d}
                \end{subtable}
        \caption{\footnotesize \textbf{Quantitative results.} Evaluated on 680 images.}
        \label{tab1}
        \end{table}

\begin{figure}[h]
\centering
\begin{minipage}{.45\textwidth}
\centering
\textit{“eighteen suitcases”}
\end{minipage}

\smallskip
\subfloat[$s_{count}=0$]{
\includegraphics[width=.22\columnwidth]{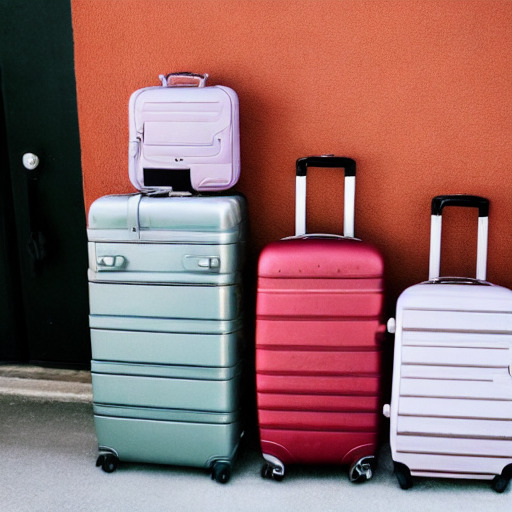}
\label{fail_0}
}
\subfloat[$s_{count}=1$]{
\includegraphics[width=.22\columnwidth]{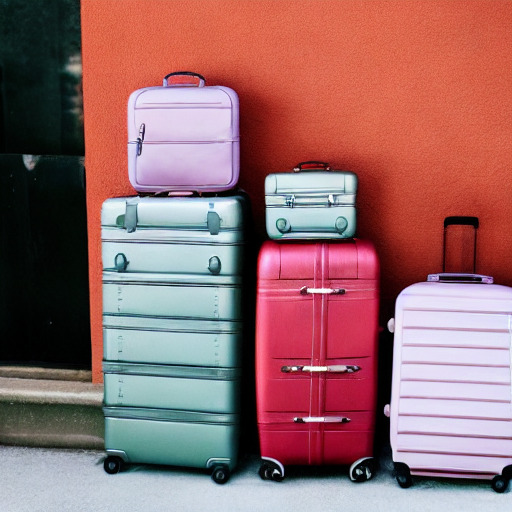}
\label{fail_1}
}
\subfloat[$s_{count}=3$]{
\includegraphics[width=.22\columnwidth]{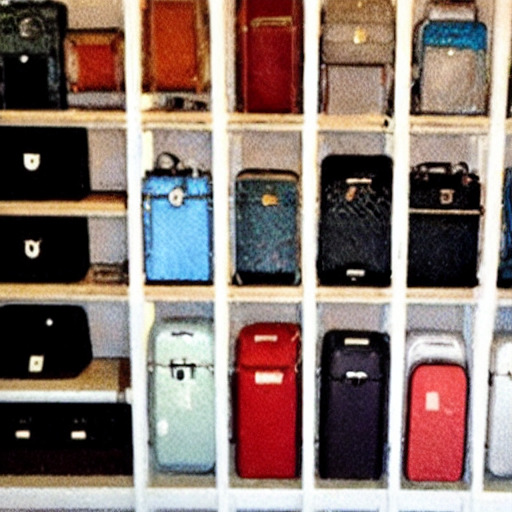}
\label{fail_3}
}
\subfloat[$s_{count}=10$]{
\includegraphics[width=.22\columnwidth]{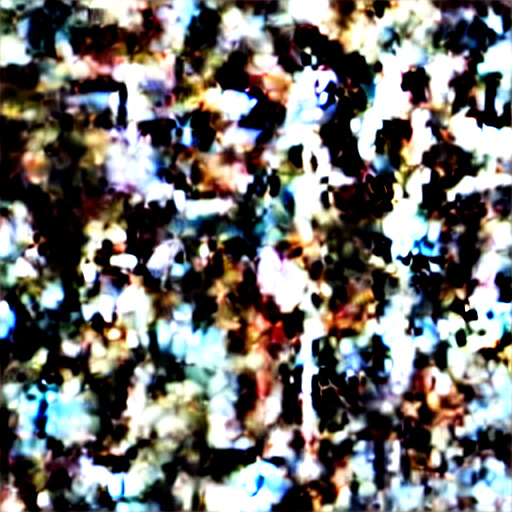}
\label{fail_10}
}
\caption{\textbf{Failure Cases.}}
\label{failure}
\end{figure}

\paragraph{Failure Cases.}
\cref{failure} highlights some failure cases of our method concerning the selection of $s_{count}$. For the prompt \textit{“eighteen suitcases”}, the vanilla Stable Diffusion generates only four suitcases. Given the large gap between eighteen and four, with $s_{count}=1$, our method adds only one additional suitcase. Increasing $s_{count}$ to 3 results in more suitcases, but it compromises the structure and quality of the image. At $s_{count}=10$, the image becomes significantly distorted. These results emphasize the critical importance of careful hyperparameter selection.

\section{Limitations}
As our results show, our method aids in generating the exact number of each object. However, it is often necessary to tune the scale parameters of the counting network guidance for a specific text prompt (\cref{failure}). Although constant or linear scheduling of $s_{count}$ can help to control the number of objects to a certain degree, generating the exact number of each object may require tuning the underlying scale parameters.

\section{Conclusions}
\label{sec:conclusion}
In this paper, we proposed counting guidance, which, to our knowledge, is the first attempt to guide Stable Diffusion with a counting network to generate the correct number of objects. For a single object type, we calculate the gradient of a counting network and refine the estimated noise at every step. For multiple object types, we discuss the semantic information mixing problem and propose attention map guidance to alleviate it. Finally, we obtain masks of each object from the corresponding attention map and calculate the counting network's gradient for each masked image separately. We demonstrated that our method effectively controls the number of objects. For future work, we will aim to remove the occasional need for hyperparameter tuning and ensure the framework works more robustly for any prompt.%

\section*{Acknowledgements}
This work was supported in part by Institute of Information \& communications Technology Planning \& Evaluation (IITP) under the Artificial Intelligence Convergence Innovation Human Resources Development (IITP-2024-RS-2023-00255968) grant funded by the Korea government (MSIT) ITRC and by the MSIT (Ministry of Science and ICT), Korea, under the ITRC (Information Technology Research Center) support program (IITP-2024-2020-0-01461) supervised by the IITP (Institute for Information \& communications Technology Planning \& Evaluation).

{\small
\bibliographystyle{ieee_fullname}
\bibliography{egbib}
}

\clearpage
\appendix
\section{Supplementary}

This supplementary section provides more information about our experiments, evaluation methods and additional quantitative and qualitative results. We describe in detail how we generate our two evaluation datasets and how we calculate the counting performance of our and previous approaches. We additionally provide more quantitative results to visualize the impact of the choice of our hyperparameters. Finally, we provide a further rich qualitative comparison of our method, Stable Diffusion and Attend-and-Excite to show that our approach outperforms existing ones in various scenarios.

\subsection{Dataset}
We create two separate datasets for measuring counting loss guidance evaluated by our counting metric and attention loss guidance evaluated by text-image/text-text similarity.
The dataset for counting evaluation consists of prompts of a single object with a specific object count. We utilize the 34 object classes from \cref{tab3}, providing a good balance between simpler to generate objects like fruits and more complex objects like animals. We cover a broad range of object counts ranging from 1-20 per object class to test and compare our method to previous ones. We generate 680 prompts (20 different counts times 34 objects) with the template of the form “\{count\} \{object\}” to construct prompts like “one apple”, “three lemons” and “six onions”.

For evaluating our attention loss guidance we use the same 34 objects and build prompts containing two object classes per prompt. Specifically, we form object pairs by combining each object with each other disregarding order and create two prompts per pair with a random count for each object ranging from 1-20. This results in a total of 1122 prompts. We use the template “\{count\_a\} \{object\_a\} and \{count\_b\} \{object\_b\}” yielding examples like “ten cats and five birds”, “nineteen birds and eight lemons” and “five elephants and twelve chicks”.

\setlength{\tabcolsep}{4pt}
\begin{table}[htb!]
\caption{Dataset}
\begin{center}

\begin{tabular}{|c||c|}
\hline
Animals & \begin{tabular}{c} cat, dog, bird, bear, lion, horse, elephant, \\ monkey, frog, turtle, rabbit, mouse, chick \end{tabular} \\
\hline
Objects & \begin{tabular}{c} backpack, glasses, crown, suitcase, chair, \\ balloon, bow, car, bowl, bench, clock, apple, \\ banana, donut, orange, egg, tomato, lemon, \\ macaron, bread, onion\end{tabular} \\
\hline
\end{tabular}
\end{center}

\label{tab3}
\end{table}
\setlength{\tabcolsep}{1.4pt}

\subsection{Testing Environment}
For our experiments, we use PyTorch \cite{pytorch} with a single NVIDIA Tesla V100 32GB GPU. It takes about 12 seconds to generate one image with vanilla Stable Diffusion, while our method takes about 26.9 seconds when using counting guidance for a single object. For two object classes it takes 15 seconds when using attention map guidance only and 37.6 seconds when using both attention map guidance and counting guidance.

\subsection{Counting Metric}
To calculate our counting metric, we use the state of the art pretrained object detection model Grounding DINO \cite{liu2023grounding} with Swin-T \cite{liu2021swin} backbone to detect bounding boxes in the generated images. We use the fact that Grounding DINO is able to perform object detection with arbitrary class labels specified as prompts and thus use the objects in the prompt as detection classes. After detection, we count the number of output boxes per object class and compare it with the ground truth count in the prompt. To balance the influence of small and large object counts on the final metric, we additionally normalize our metric by the ground truth object count. Our normalized MAE metric for one object class is given as 
\begin{equation}
\textit{MAE} = \frac{1}{n} \sum_{i=1}^{n} \left|\frac{\hat{y_i}-y_i}{y_i}\right|,
\end{equation}
while our normalized RMSE metric is defined as
\begin{equation}
\textit{RMSE} = \sqrt{\frac{1}{n} \sum_{i=1}^{n} \left(\frac{\hat{y_i}-y_i}{y_i}\right)^2},
\end{equation}
where $y_i$ is the ground truth object count from the prompt and $\hat{y_i}$ is the number of detected bounding boxes in the generated image for the respective class.

\subsection{Hyperparameter Analysis}
\paragraph{Counting Loss Scale}
To determine the ideal counting loss scale, we run our method with various scales on our 680 prompts counting dataset and plot the resulting \textit{MAE} and \textit{RMSE} metrics in \cref{subfig:hyper_a,subfig:hyper_b}. We choose $s_{count}=1$ for our method (constant) since it provides a good value for both \textit{MAE} and \textit{RMSE}. As $s_{count}$ increases, the counting error initially decreases but subsequently rises, exhibiting the behavior of a \textbf{convex function.} While excessive gradient guidance can negatively impact image generation, we demonstrate that increasing counting guidance up to a certain threshold can effectively reduce the counting error.

\cref{subfig:hyper_c} shows the counting error (MAE) versus the number of objects $N$ in the prompt for five $s_{count}$ values, and \cref{subfig:hyper_d} depicts its linear trend. As $s_{count}$ increases, the slope of the linear trend gradually decreases. As a result, for small $N$, the performance is better when the $s_{count}$ is smaller, while for large N, the performance improves as the $s_{count}$ increases. This observed trend aligns with the intuition that increasing $N$ poses greater challenges for accurate generation, thereby necessitating a larger $s_{count}$.

Our analysis yielded $s_{count} = \max (0.01, 0.2N - 1)$, which is a simple increasing function of $N$ that significantly improves performance compared to a constant value.

\paragraph{Attention Loss Scale}
Similarly, we visualize the text-text and text-image similarity on our 1122 multi object class dataset for various attention loss scales in \cref{clip_plot}. We notice a strong peak of text-text similarity at the value 1 and thus choose our attention loss scale for our experiments as 1.

\subsection{Additional Qualitative Results}

\cref{fig8}, \cref{fig9} and \cref{fig10} show additional results for our counting guidance with various prompts and varying object count for Stable Diffusion, Attend and Excite and ours. Even though we need to tweak our counting guidance scale hyperparameter for some prompts, our counting guidance method consistently creates the correct amount or, when dealing with large count, a similar amount of objects, whereas Stable Diffusion and Attend and Excite fail in many cases. When the object count grows, it becomes more challenging to generate the exact amount, however, our method nevertheless outperforms the other two tested methods.

\cref{fig11} visualizes the attention map per object for several prompts for Stable Diffusion and our attention map guidance. We note that our attention maps capture the spatial location of each object more accurately than Stable Diffusion, while reducing the overlap between different objects.

\subsection{Template for User Study and GPT Evaluation}
\paragraph{User Study}

Compare the first and second images provided, and select the one that more closely aligns with the given prompt. Pay particular attention to the object count.

\paragraph{GPT Evaluation Prompt}

Compare the first and second images provided, and select the one that more closely aligns with the given prompt. Pay particular attention to the accuracy of the object count. Your selection can be subjective. Your final output score must be either 0 (if the first image is best), 0.5 ('Tie'), or 1 (if the second image is best). You have to give your output in this way (Keep your reasoning concise and short.  Give your intermediate thinking step by step.)

\begin{figure*}[h]
\centering
\begin{subfigure}{0.49\linewidth}
\centering
\begin{tikzpicture}[scale=0.8]
\begin{axis}[
    xmode=log,
    xlabel={counting loss scale $s_{count}$},
    ylabel={counting error (RMSE)},
    xmin=0.1, xmax=10,
    ymin=0.65, ymax=0.8,
    xtick={0.1,1,10},
    ytick={0.7,0.8},
    legend pos=south west,
    ymajorgrids=true,
    grid style=dashed,
    every axis plot/.append style={ultra thick}
]

\addplot[
    color=red,
    mark=square,
    ]
    coordinates {
    (0.1,0.7334471699585519)
    (0.3,0.7039613598022211)
    (0.5,0.7128141200299508)
    (0.7,0.6971769249336324)
    (1,0.6959092241241479)
    (3,0.6871137157401223)
    (5,0.7125691504550064)
    (7,0.7448865156157942)
    (10,0.7623909901497313)
    };
    \addlegendentry{RMSE}
\addplot[mark=none, dashed, red, domain=0.1:10]{0.7457674031354116};
\end{axis}

\end{tikzpicture}
\caption{\footnotesize Effect of $s_{count}$ on RMSE}
\label{subfig:hyper_a}
\end{subfigure}
\begin{subfigure}{0.49\linewidth}
\centering
\begin{tikzpicture}[scale=0.8]

\begin{axis}[
    xmode=log,
    xlabel={counting loss scale $s_{count}$},
    ylabel={counting error (MAE)},
    xmin=0.1, xmax=10,
    ymin=0.55, ymax=0.70,
    xtick={0.1,1,10},
    ytick={0.6,0.7},
    legend pos=south west,
    ymajorgrids=true,
    grid style=dashed,
    every axis plot/.append style={ultra thick}
]
\addplot[
    color=blue,
    mark=square,
    ]
    coordinates {
    (0.1,0.6015676607357312)
    (0.3,0.5862615980687599)
    (0.5,0.5883252656798573)
    (0.7,0.5822178382762747)
    (1,0.5845010453538044)
    (3,0.5917011984896364)
    (5,0.6228075450331756)
    (7,0.6599791872707212)
    (10,0.6878658033572895)
    };
    \addlegendentry{MAE}

\addplot[mark=none, dashed, blue, domain=0.1:10]{0.598861050401924};

\end{axis}
\end{tikzpicture}
\caption{Effect of $s_{count}$ on MAE}
\label{subfig:hyper_b}
\end{subfigure}

\bigskip

\begin{subfigure}{0.49\linewidth}
\centering
\begin{tikzpicture}[scale=0.8]

\begin{axis}[
    xlabel={number of object $N$},
    ylabel={counting error (MAE)},
    xmin=2, xmax=20,
    ymin=0.35, ymax=0.85,
    xtick={5,10,15,20},
    ytick={0.4,0.5,0.6,0.7,0.8},
    legend pos=south east,
    ymajorgrids=true,
    grid style=dashed,
    every axis plot/.append style={ultra thick}
]

\addplot[
    color=my00,
    mark=square,
    ]
    coordinates {
(2,0.47058823529411764)
(3,0.37254901960784315)
(4,0.4411764705882353)
(5,0.4294117647058823)
(6,0.45588235294117646)
(7,0.4831932773109243)
(8,0.5220588235294118)
(9,0.611111111111111)
(10,0.7088235294117646)
(11,0.6818181818181818)
(12,0.6740196078431372)
(13,0.6651583710407242)
(14,0.7205882352941176)
(15,0.7137254901960784)
(16,0.7316176470588235)
(17,0.759515570934256)
(18,0.8071895424836601)
(19,0.7647058823529411)
(20,0.7485294117647059)
    };
    \addlegendentry{$s_{count}=0.0$}
\addplot[
    color=my01,
    mark=square,
    ]
    coordinates {
(2,0.5)
(3,0.38235294117647056)
(4,0.47794117647058826)
(5,0.44705882352941173)
(6,0.4607843137254902)
(7,0.5084033613445378)
(8,0.5220588235294118)
(9,0.5980392156862745)
(10,0.6676470588235294)
(11,0.6951871657754011)
(12,0.6617647058823529)
(13,0.6719457013574662)
(14,0.7016806722689076)
(15,0.707843137254902)
(16,0.7095588235294118)
(17,0.759515570934256)
(18,0.8039215686274509)
(19,0.760061919504644)
(20,0.7308823529411765)
    };
    \addlegendentry{$s_{count}=0.1$}
\addplot[
    color=my03,
    mark=square,
    ]
    coordinates {
(2,0.4852941176470588)
(3,0.39215686274509803)
(4,0.4485294117647059)
(5,0.46470588235294114)
(6,0.4901960784313726)
(7,0.5252100840336134)
(8,0.5330882352941176)
(9,0.5784313725490196)
(10,0.638235294117647)
(11,0.6684491978609626)
(12,0.6715686274509803)
(13,0.658371040723982)
(14,0.6974789915966386)
(15,0.607843137254902)
(16,0.6985294117647058)
(17,0.7110726643598617)
(18,0.7859477124183006)
(19,0.7554179566563468)
(20,0.7088235294117647)
    };
    \addlegendentry{$s_{count}=0.3$}
\addplot[
    color=my10,
    mark=square,
    ]
    coordinates {
(2,0.5441176470588235)
(3,0.4215686274509804)
(4,0.4411764705882353)
(5,0.4882352941176471)
(6,0.5196078431372549)
(7,0.5672268907563024)
(8,0.5367647058823529)
(9,0.5718954248366013)
(10,0.6352941176470588)
(11,0.5828877005347595)
(12,0.5612745098039216)
(13,0.6832579185520363)
(14,0.6428571428571429)
(15,0.6666666666666666)
(16,0.6691176470588235)
(17,0.7024221453287198)
(18,0.7500000000000001)
(19,0.760061919504644)
(20,0.7102941176470589)
    };
    \addlegendentry{$s_{count}=1.0$}
\addplot[
    color=my30,
    mark=square,
    ]
    coordinates {
(2,0.5588235294117647)
(3,0.4411764705882352)
(4,0.5882352941176471)
(5,0.5470588235294117)
(6,0.6078431372549019)
(7,0.6302521008403361)
(8,0.5919117647058824)
(9,0.5980392156862745)
(10,0.6470588235294117)
(11,0.6470588235294118)
(12,0.6053921568627451)
(13,0.6515837104072398)
(14,0.6512605042016807)
(15,0.6431372549019607)
(16,0.6378676470588235)
(17,0.6384083044982698)
(18,0.7058823529411765)
(19,0.7136222910216719)
(20,0.6705882352941177)
    };
    \addlegendentry{$s_{count}=3.0$}

\end{axis}
\end{tikzpicture}
\caption{Effect of $N$ on MAE}
\label{subfig:hyper_c}
\end{subfigure}
\begin{subfigure}{0.49\linewidth}
\centering
\begin{tikzpicture}[scale=0.8]

\begin{axis}[
    xlabel={number of object $N$},
    ylabel={counting error (MAE)},
    xmin=2, xmax=20,
    ymin=0.35, ymax=0.85,
    xtick={5,10,15,20},
    ytick={0.4,0.5,0.6,0.7,0.8},
    legend pos=south east,
    ymajorgrids=true,
    grid style=dashed,
    every axis plot/.append style={ultra thick}
]

        \addplot [ultra thick, my00] table[y={create col/linear regression}]{
2 0.47058823529411764
3 0.37254901960784315
4 0.4411764705882353
5 0.4294117647058823
6 0.45588235294117646
7 0.4831932773109243
8 0.5220588235294118
9 0.611111111111111
10 0.7088235294117646
11 0.6818181818181818
12 0.6740196078431372
13 0.6651583710407242
14 0.7205882352941176
15 0.7137254901960784
16 0.7316176470588235
17 0.759515570934256
18 0.8071895424836601
19 0.7647058823529411
20 0.7485294117647059
        };
        \addplot [ultra thick, my01] table[y={create col/linear regression}]{
2 0.5
3 0.38235294117647056
4 0.47794117647058826
5 0.44705882352941173
6 0.4607843137254902
7 0.5084033613445378
8 0.5220588235294118
9 0.5980392156862745
10 0.6676470588235294
11 0.6951871657754011
12 0.6617647058823529
13 0.6719457013574662
14 0.7016806722689076
15 0.707843137254902
16 0.7095588235294118
17 0.759515570934256
18 0.8039215686274509
19 0.760061919504644
20 0.7308823529411765
        };
        \addplot [ultra thick, my03] table[y={create col/linear regression}]{
2 0.4852941176470588
3 0.39215686274509803
4 0.4485294117647059
5 0.46470588235294114
6 0.4901960784313726
7 0.5252100840336134
8 0.5330882352941176
9 0.5784313725490196
10 0.638235294117647
11 0.6684491978609626
12 0.6715686274509803
13 0.658371040723982
14 0.6974789915966386
15 0.607843137254902
16 0.6985294117647058
17 0.7110726643598617
18 0.7859477124183006
19 0.7554179566563468
20 0.7088235294117647
        };
        \addplot [ultra thick, my10] table[y={create col/linear regression}]{
2 0.5441176470588235
3 0.4215686274509804
4 0.4411764705882353
5 0.4882352941176471
6 0.5196078431372549
7 0.5672268907563024
8 0.5367647058823529
9 0.5718954248366013
10 0.6352941176470588
11 0.5828877005347595
12 0.5612745098039216
13 0.6832579185520363
14 0.6428571428571429
15 0.6666666666666666
16 0.6691176470588235
17 0.7024221453287198
18 0.7500000000000001
19 0.760061919504644
20 0.7102941176470589
        };
        \addplot [ultra thick, my30] table[y={create col/linear regression}]{
2 0.5588235294117647
3 0.4411764705882352
4 0.5882352941176471
5 0.5470588235294117
6 0.6078431372549019
7 0.6302521008403361
8 0.5919117647058824
9 0.5980392156862745
10 0.6470588235294117
11 0.6470588235294118
12 0.6053921568627451
13 0.6515837104072398
14 0.6512605042016807
15 0.6431372549019607
16 0.6378676470588235
17 0.6384083044982698
18 0.7058823529411765
19 0.7136222910216719
20 0.6705882352941177
        };
\addlegendentry{$s_{count}=0.0$}
\addlegendentry{$s_{count}=0.1$}
\addlegendentry{$s_{count}=0.3$}
\addlegendentry{$s_{count}=1.0$}
\addlegendentry{$s_{count}=3.0$}
\end{axis}
\end{tikzpicture}
\caption{Effect of $N$ (linear trend)}
\label{subfig:hyper_d}
\end{subfigure}
\caption{\textbf{Hyperparameter study.} Evaluated on 680 images.}
\label{fig:hyper}
\end{figure*}

\begin{figure*}[t]
\captionsetup[subfigure]{labelformat=empty}
\centering
\begin{subfigure}{0.49\linewidth}
\centering
\begin{tikzpicture}[scale=0.8]
\begin{axis}[
    xmode=log,
    xlabel={attention loss scale},
    ylabel={similarity},
    xmin=0.1, xmax=5,
    ymin=0.31, ymax=0.33,
    xtick={0.1,1,5},
    ytick={0.32,0.33},
    legend pos=south west,
    ymajorgrids=true,
    grid style=dashed,
    every axis plot/.append style={ultra thick}
]
\addplot[
    color=red,
    mark=square,
    ]
    coordinates {
    (0.1,0.326372843358689)
    (0.25,0.32823449140622296)
    (0.5,0.32885376155961316)
    (0.75,0.3291451955591324)
    (1,0.32878415980422)
    (1.5,0.32804452375999843)
    (2,0.32736008339806605)
    (3,0.3228102243845214)
    (4,0.3203670132330334)
    (5,0.31546993422040337)
    };
    \addlegendentry{text-image}
\addplot[mark=none, solid, red, domain=0.1:5]{0.3235054808094712};
\end{axis}
\end{tikzpicture}
\end{subfigure}
\begin{subfigure}{0.49\linewidth}
\centering
\begin{tikzpicture}[scale=0.8]
\begin{axis}[
    xmode=log,
    xlabel={attention loss scale},
    ylabel={similarity},
    xmin=0.1, xmax=5,
    ymin=0.7, ymax=0.74,
    xtick={0.1,1,5},
    ytick={0.72,0.74},
    legend pos=south west,
    ymajorgrids=true,
    grid style=dashed,
    every axis plot/.append style={ultra thick}
]
\addplot[
    color=blue,
    mark=square,
    ]
    coordinates {
    (0.1,0.7253394928444293)
    (0.25,0.7256261441826714)
    (0.5,0.7288233029268131)
    (0.75,0.7304018688031757)
    (1,0.7323044803821009)
    (1.5,0.7298874847331714)
    (2,0.7273298607512737)
    (3,0.7231072480395689)
    (4,0.7155876819519567)
    (5,0.7046323355343808)
    };
    \addlegendentry{text-text}
\addplot[mark=none, solid, blue, domain=0.1:5]{0.7220300709318626};
\end{axis}
\end{tikzpicture}
\end{subfigure}
\caption{Effect of attention loss scale on the text-image and text-text CLIP similarity. Evaluated on our 1122 two object prompt dataset.}
\label{clip_plot}
\end{figure*}

\begin{figure*}[h]
\captionsetup[subfigure]{labelformat=empty}
\centering

\begin{minipage}{.5\textwidth}
\centering
\bigskip
Stable Diffusion
\medskip
\end{minipage}

\subfloat[\textit{“an apple”}]{
\includegraphics[width=.31\columnwidth]{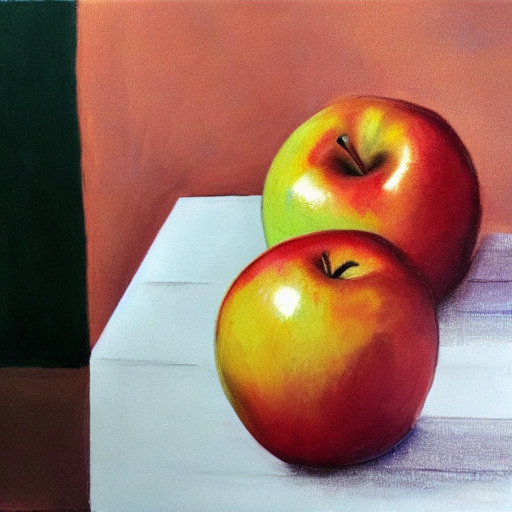}
\label{fig_sd_apple_1}
}
\subfloat[\textit{“seven apples”}]{
\includegraphics[width=.31\columnwidth]{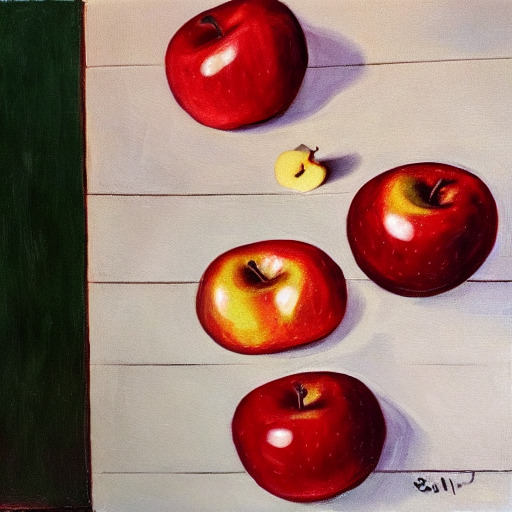}
\label{fig_sd_apple_7}
}
\subfloat[\textit{“eight apples”}]{
\includegraphics[width=.31\columnwidth]{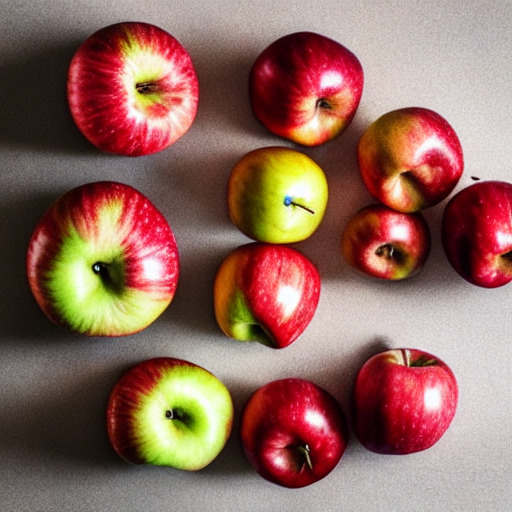}
\label{fig_sd_apple_8}
}
\subfloat[\textit{“nine apples”}]{
\includegraphics[width=.31\columnwidth]{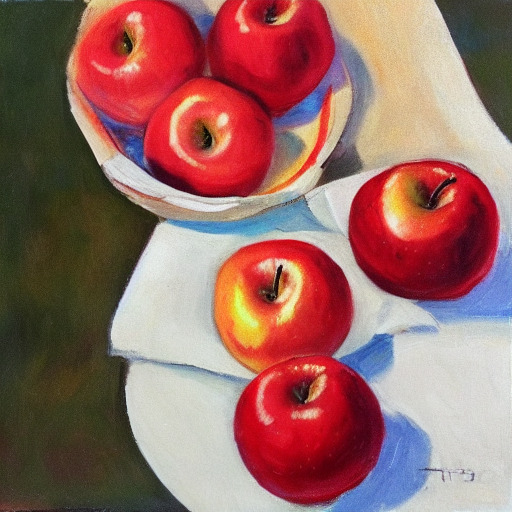}
\label{fig_sd_apple_9}
}
\subfloat[\textit{“ten apples”}]{
\includegraphics[width=.31\columnwidth]{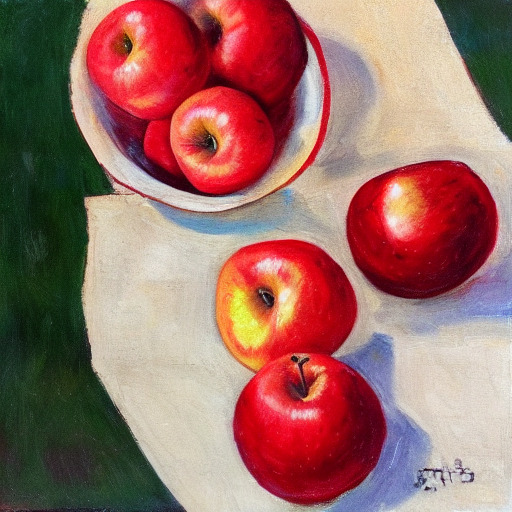}
\label{fig_sd_apple_10}
}
\subfloat[\textit{“thirteen apples”}]{
\includegraphics[width=.31\columnwidth]{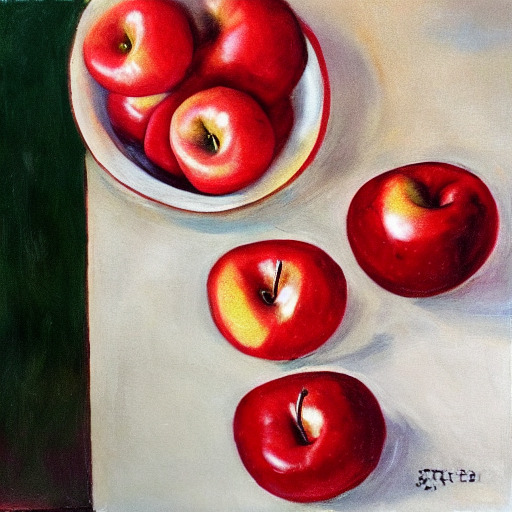}
\label{fig_sd_apple_13}
}

\subfloat[\textit{“two donuts”}]{
\includegraphics[width=.31\columnwidth]{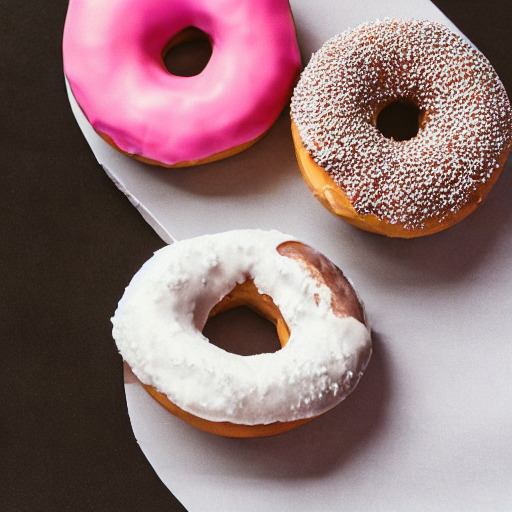}
\label{fig_sd_donut_2}
}
\subfloat[\textit{“five donuts”}]{
\includegraphics[width=.31\columnwidth]{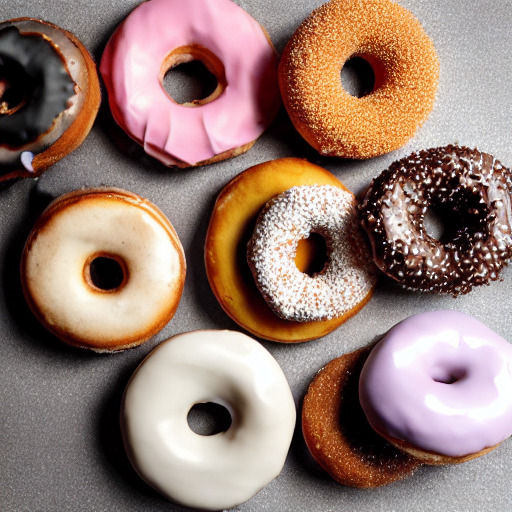}
\label{fig_sd_donut_5}
}
\subfloat[\textit{“six donuts”}]{
\includegraphics[width=.31\columnwidth]{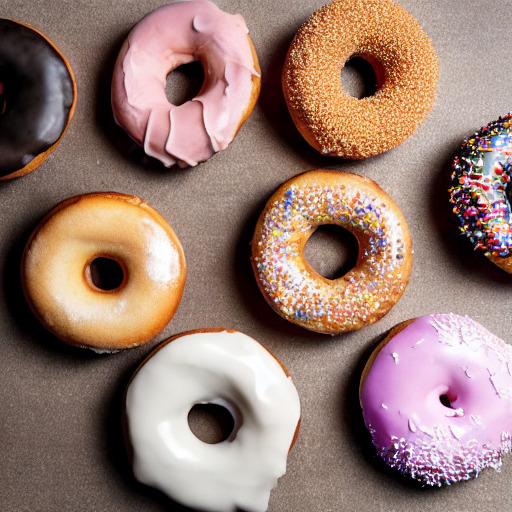}
\label{fig_sd_donut_6}
}
\subfloat[\textit{“seven donuts”}]{
\includegraphics[width=.31\columnwidth]{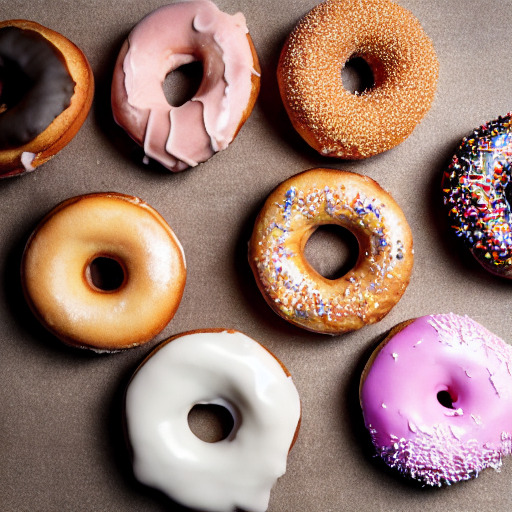}
\label{fig_sd_donut_7}
}
\subfloat[\textit{“eight donuts”}]{
\includegraphics[width=.31\columnwidth]{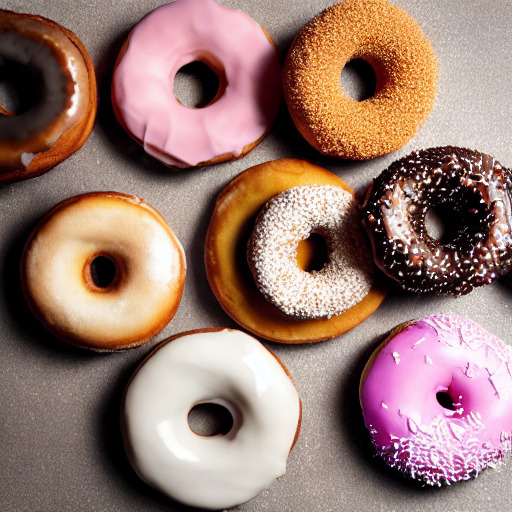}
\label{fig_sd_donut_8}
}
\subfloat[\textit{“eleven donuts”}]{
\includegraphics[width=.31\columnwidth]{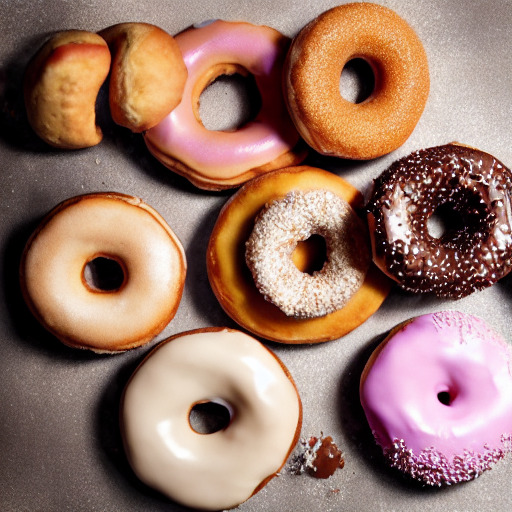}
\label{fig_sd_donut_11}
}

\begin{minipage}{.5\textwidth}
\centering

\bigskip
Attend-and-Excite
\medskip
\end{minipage}

\subfloat[\textit{“an apple”}]{
\includegraphics[width=.31\columnwidth]{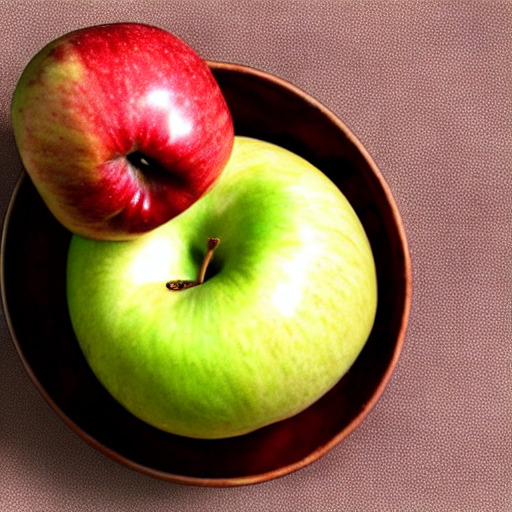}
\label{fig_ae_apple_1}
}
\subfloat[\textit{“seven apples”}]{
\includegraphics[width=.31\columnwidth]{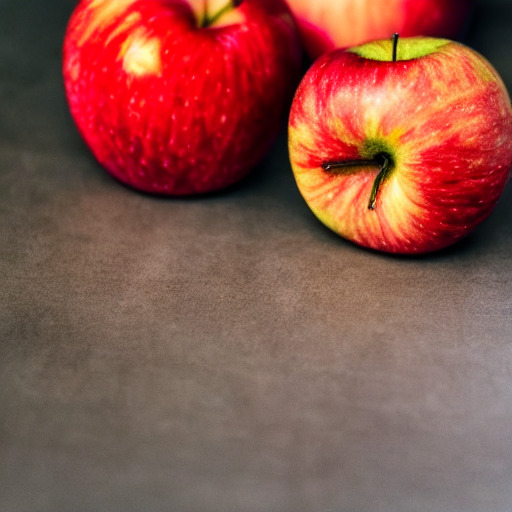}
\label{fig_ae_apple_7}
}
\subfloat[\textit{“eight apples”}]{
\includegraphics[width=.31\columnwidth]{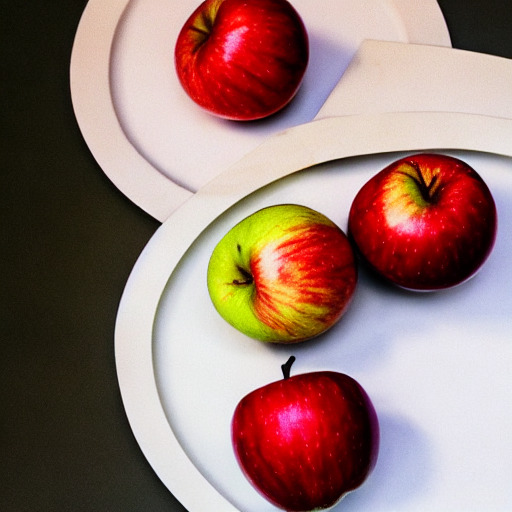}
\label{fig_ae_apple_8}
}
\subfloat[\textit{“nine apples”}]{
\includegraphics[width=.31\columnwidth]{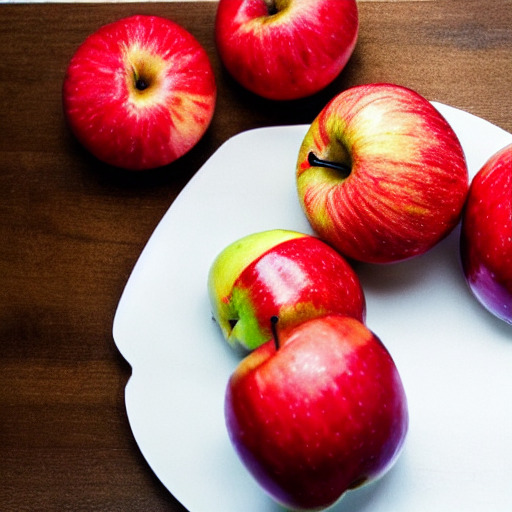}
\label{fig_ae_apple_9}
}
\subfloat[\textit{“ten apples”}]{
\includegraphics[width=.31\columnwidth]{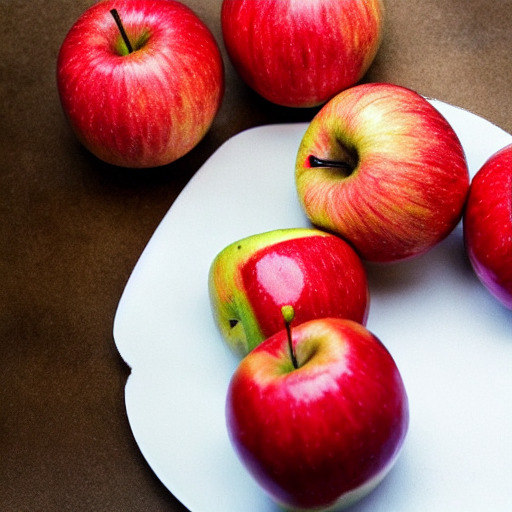}
\label{fig_ae_apple_10}
}
\subfloat[\textit{“thirteen apples”}]{
\includegraphics[width=.31\columnwidth]{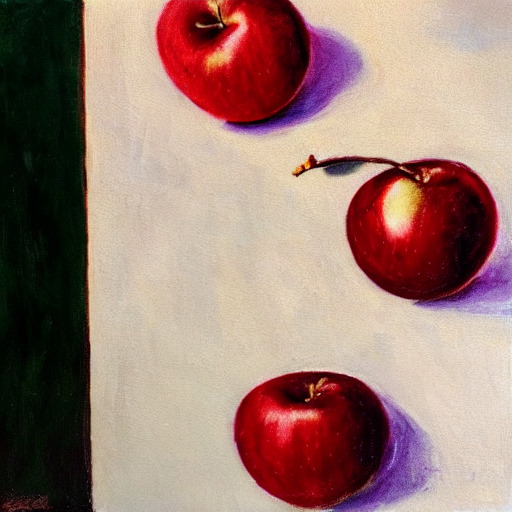}
\label{fig_ae_apple_13}
}

\subfloat[\textit{“two donuts”}]{
\includegraphics[width=.31\columnwidth]{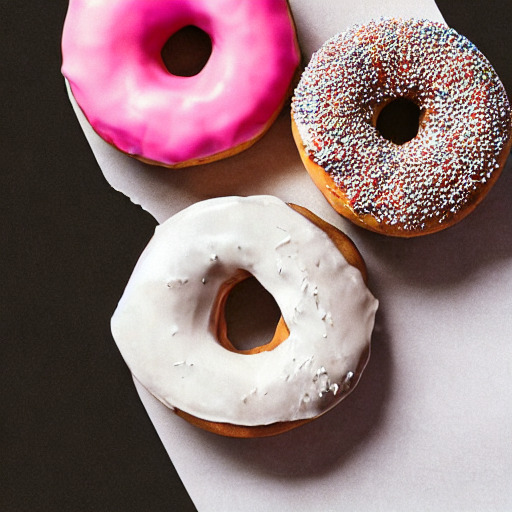}
\label{fig_ae_donut_2}
}
\subfloat[\textit{“five donuts”}]{
\includegraphics[width=.31\columnwidth]{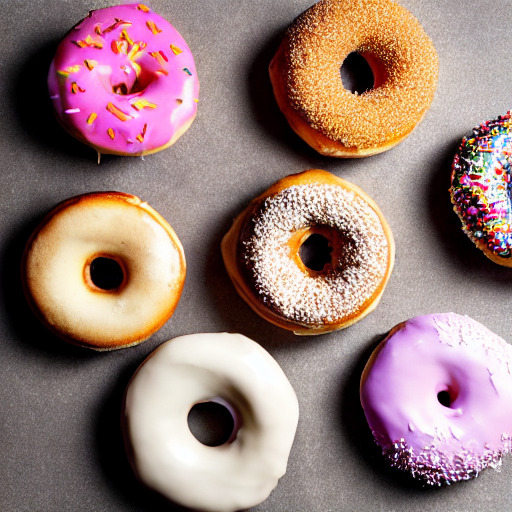}
\label{fig_ae_donut_5}
}
\subfloat[\textit{“six donuts”}]{
\includegraphics[width=.31\columnwidth]{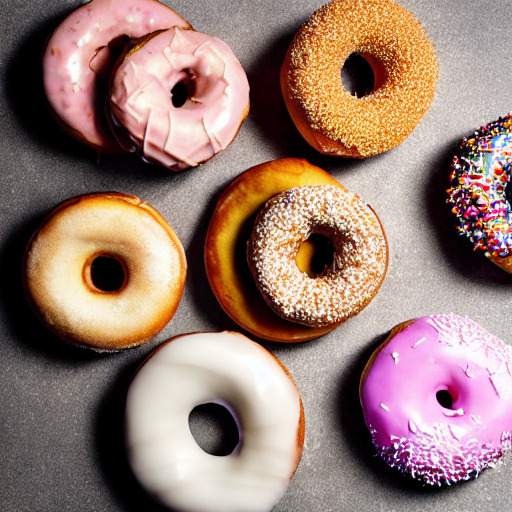}
\label{fig_ae_donut_6}
}
\subfloat[\textit{“seven donuts”}]{
\includegraphics[width=.31\columnwidth]{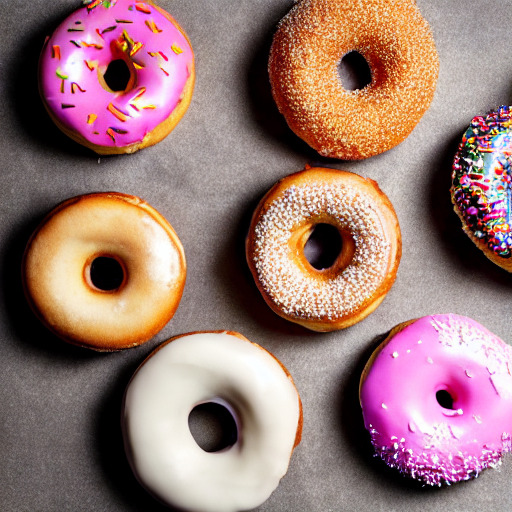}
\label{fig_ae_donut_7}
}
\subfloat[\textit{“eight donuts”}]{
\includegraphics[width=.31\columnwidth]{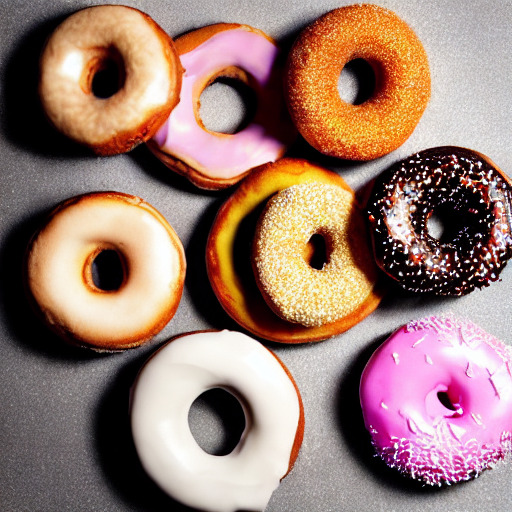}
\label{fig_ae_donut_8}
}
\subfloat[\textit{“eleven donuts”}]{
\includegraphics[width=.31\columnwidth]{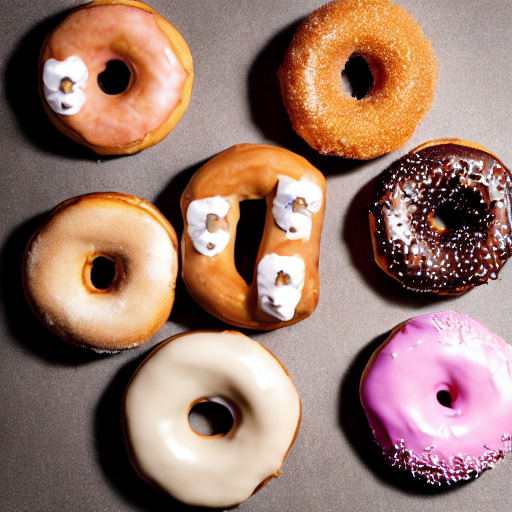}
\label{fig_ae_donut_11}
}

\begin{minipage}{.5\textwidth}
\centering

\bigskip
Ours
\medskip
\end{minipage}

\subfloat[\textit{“an apple”}]{
\includegraphics[width=.31\columnwidth]{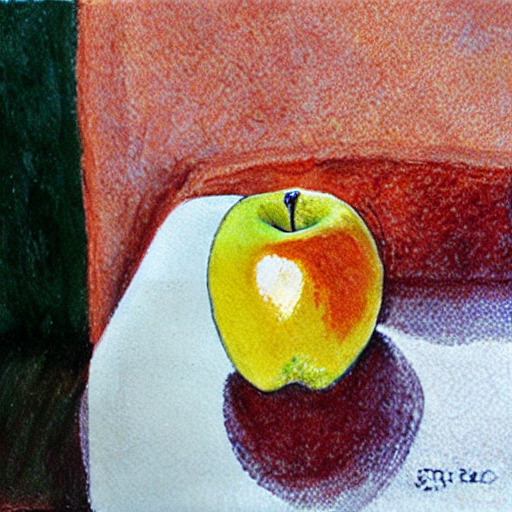}
\label{fig_ours_apple_1}
}
\subfloat[\textit{“seven apples”}]{
\includegraphics[width=.31\columnwidth]{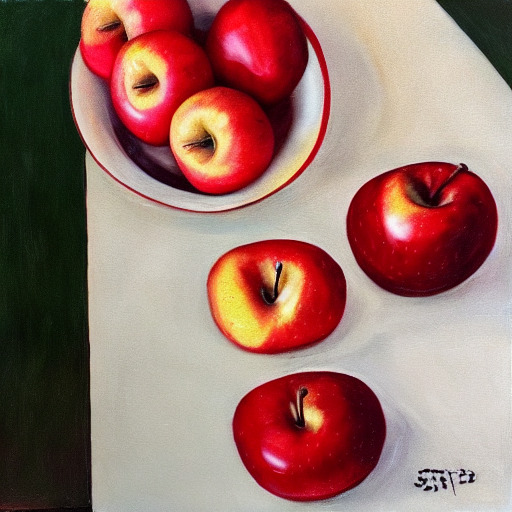}
\label{fig_ours_apple_7}
}
\subfloat[\textit{“eight apples”}]{
\includegraphics[width=.31\columnwidth]{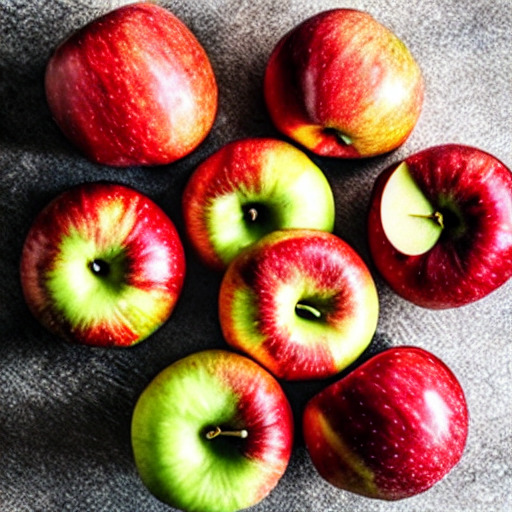}
\label{fig_ours_apple_8}
}
\subfloat[\textit{“nine apples”}]{
\includegraphics[width=.31\columnwidth]{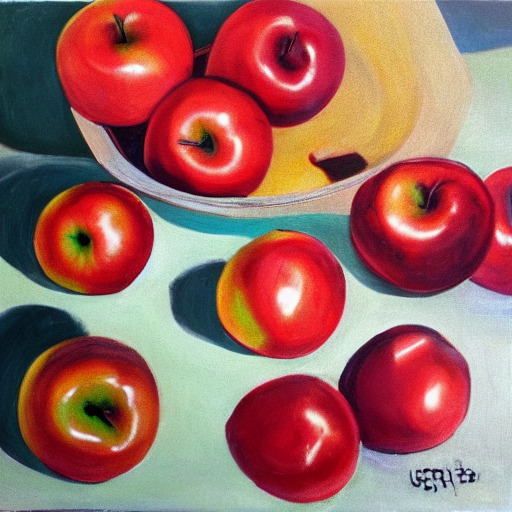}
\label{fig_ours_apple_9}
}
\subfloat[\textit{“ten apples”}]{
\includegraphics[width=.31\columnwidth]{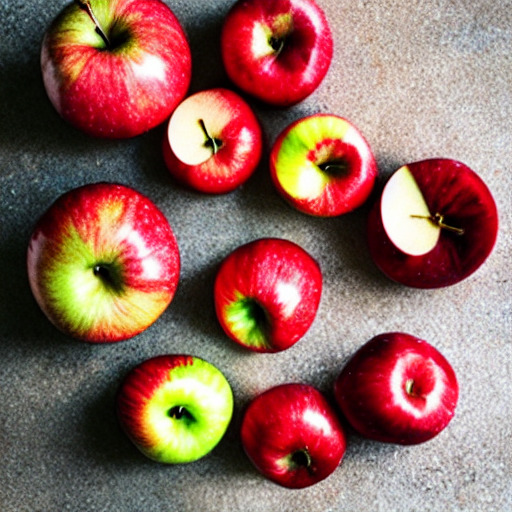}
\label{fig_ours_apple_10}
}
\subfloat[\textit{“thirteen apples”}]{
\includegraphics[width=.31\columnwidth]{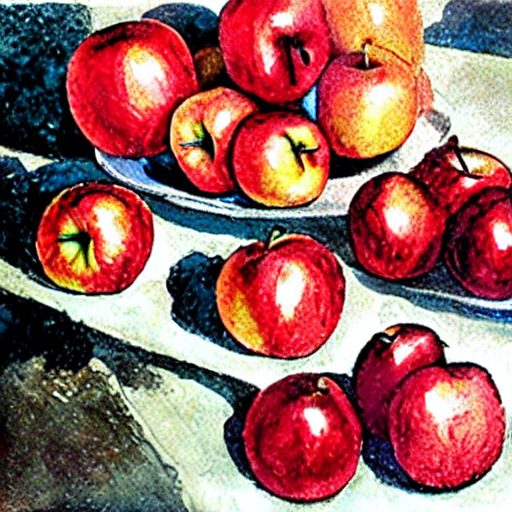}
\label{fig_ours_apple_13}
}

\subfloat[\textit{“two donuts”}]{
\includegraphics[width=.31\columnwidth]{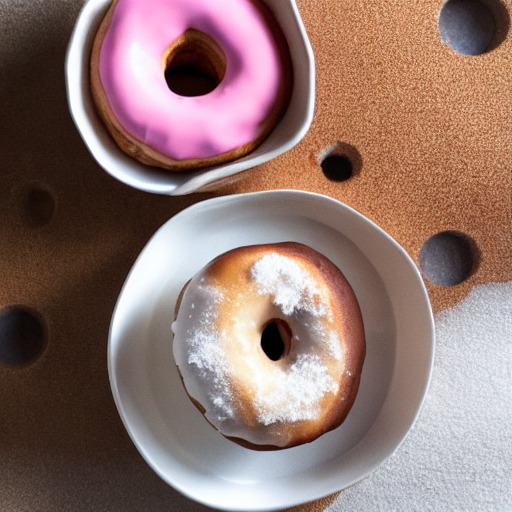}
\label{fig_ours_donut_2}
}
\subfloat[\textit{“five donuts”}]{
\includegraphics[width=.31\columnwidth]{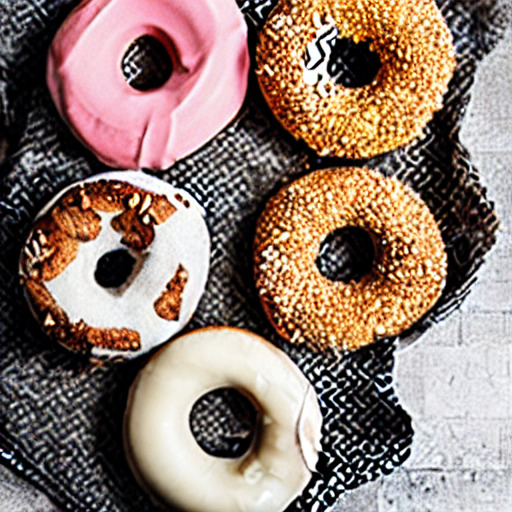}
\label{fig_ours_donut_5}
}
\subfloat[\textit{“six donuts”}]{
\includegraphics[width=.31\columnwidth]{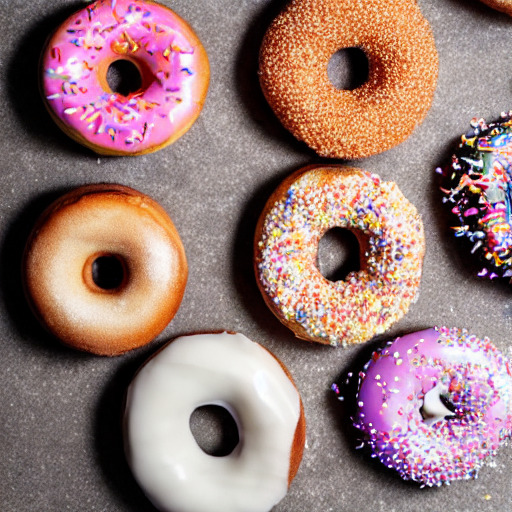}
\label{fig_ours_donut_6}
}
\subfloat[\textit{“seven donuts”}]{
\includegraphics[width=.31\columnwidth]{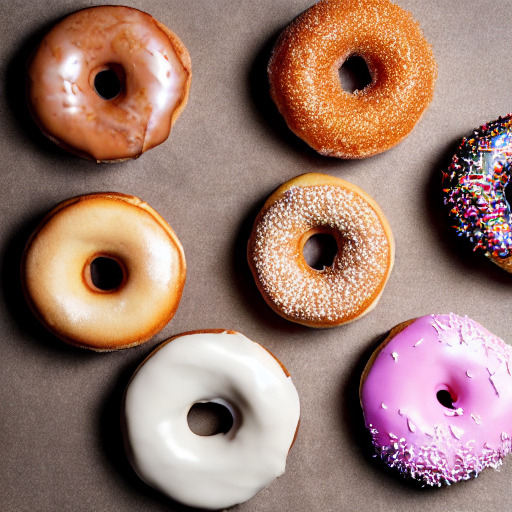}
\label{fig_ours_donut_7}
}
\subfloat[\textit{“eight donuts”}]{
\includegraphics[width=.31\columnwidth]{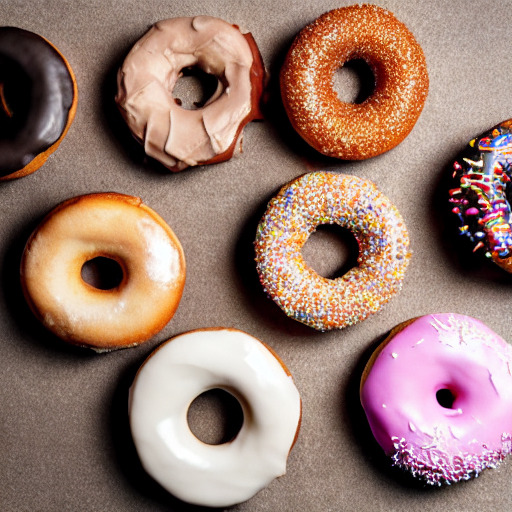}
\label{fig_ours_donut_8}
}
\subfloat[\textit{“eleven donuts”}]{
\includegraphics[width=.31\columnwidth]{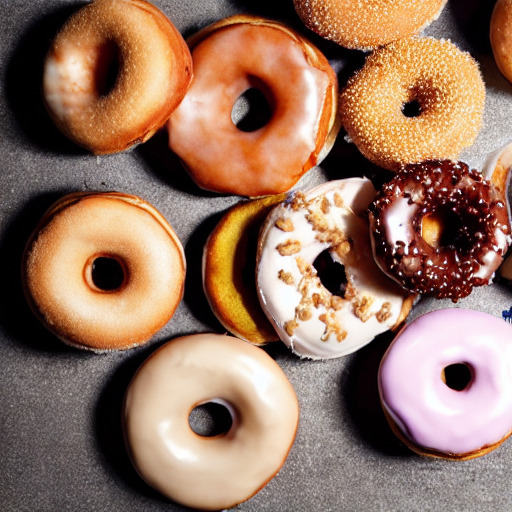}
\label{fig_ours_donut_11}
}

\caption{Additional qualitative results (1)}
\label{fig8}
\end{figure*}

\begin{figure*}[h]
\captionsetup[subfigure]{labelformat=empty}
\centering

\begin{minipage}{.5\textwidth}
\centering
\bigskip
Stable Diffusion
\medskip
\end{minipage}

\subfloat[\textit{“a macaron”}]{
\includegraphics[width=.31\columnwidth]{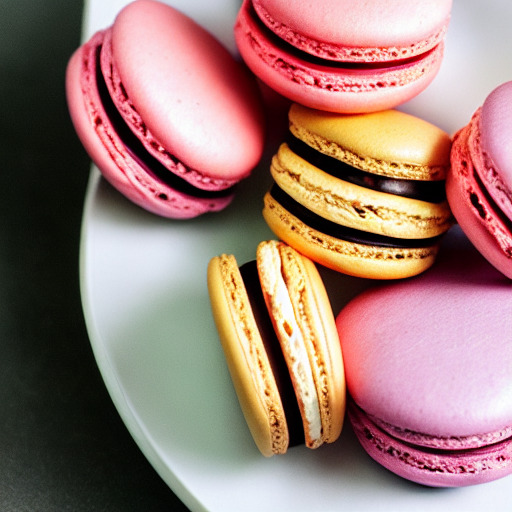}
\label{fig_sd_macaron_1}
}
\subfloat[\textit{“eight macarons”}]{
\includegraphics[width=.31\columnwidth]{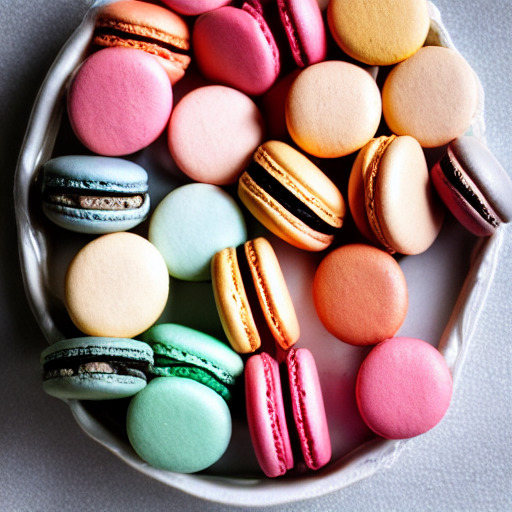}
\label{fig_sd_macaron_8}
}
\subfloat[\textit{“nine macarons”}]{
\includegraphics[width=.31\columnwidth]{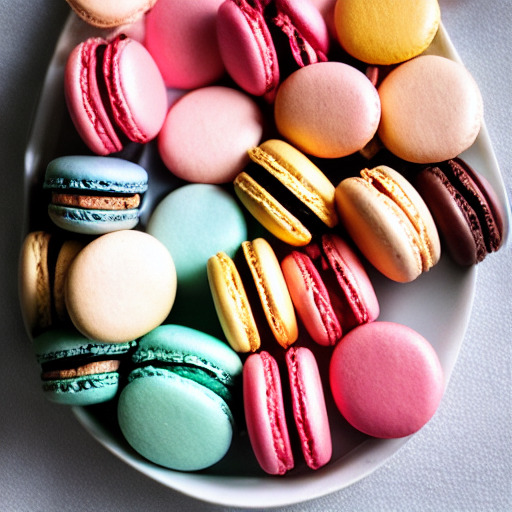}
\label{fig_sd_macaron_9}
}
\subfloat[\textit{“ten macarons”}]{
\includegraphics[width=.31\columnwidth]{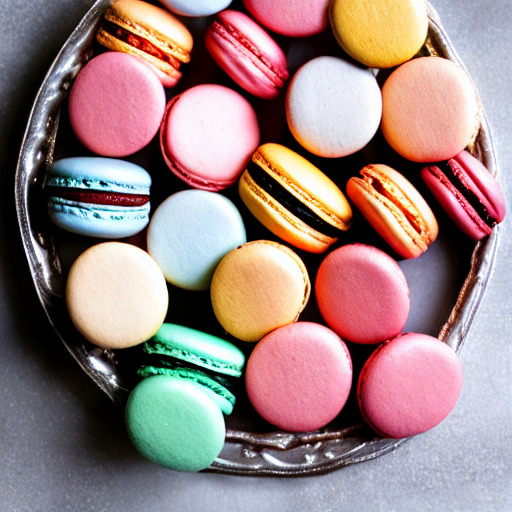}
\label{fig_sd_macaron_10}
}
\subfloat[\textit{“eleven macarons”}]{
\includegraphics[width=.31\columnwidth]{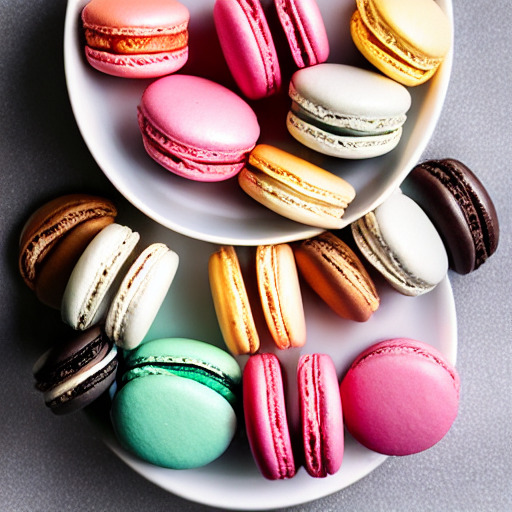}
\label{fig_sd_macaron_11}
}
\subfloat[\textit{“fourteen macarons”}]{
\includegraphics[width=.31\columnwidth]{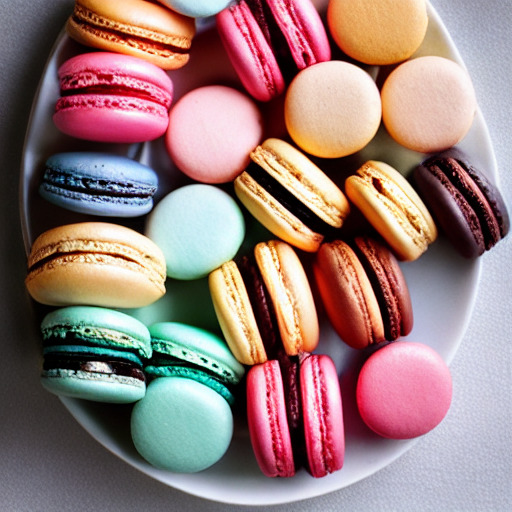}
\label{fig_sd_macaron_14}
}

\subfloat[\textit{“six eggs”}]{
\includegraphics[width=.31\columnwidth]{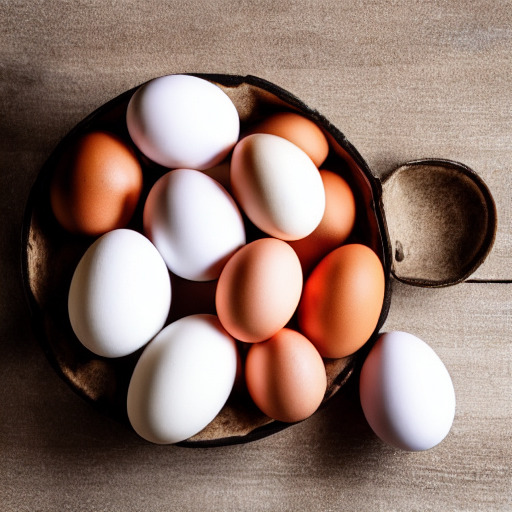}
\label{fig_sd_egg_6}
}
\subfloat[\textit{“seven eggs”}]{
\includegraphics[width=.31\columnwidth]{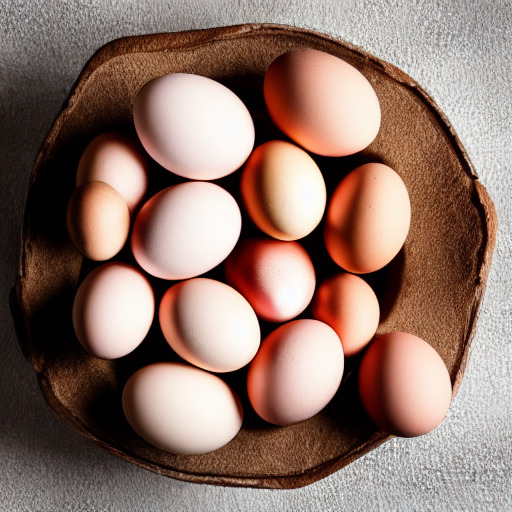}
\label{fig_sd_egg_7}
}
\subfloat[\textit{“eight eggs”}]{
\includegraphics[width=.31\columnwidth]{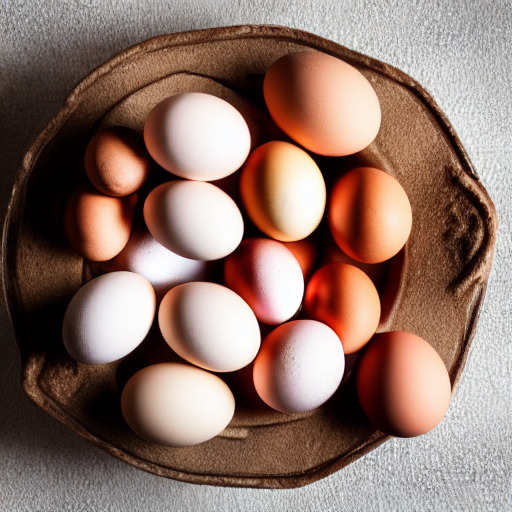}
\label{fig_sd_egg_8}
}
\subfloat[\textit{“nine eggs”}]{
\includegraphics[width=.31\columnwidth]{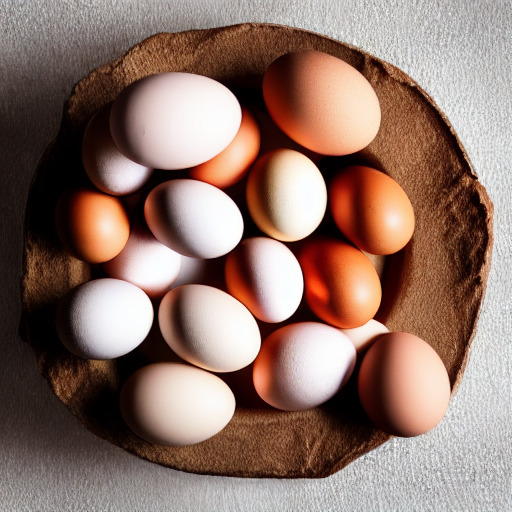}
\label{fig_sd_egg_9}
}
\subfloat[\textit{“ten eggs”}]{
\includegraphics[width=.31\columnwidth]{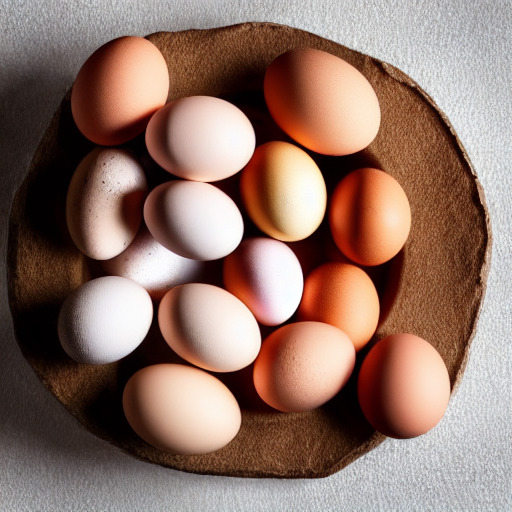}
\label{fig_sd_egg_10}
}
\subfloat[\textit{“eleven eggs”}]{
\includegraphics[width=.31\columnwidth]{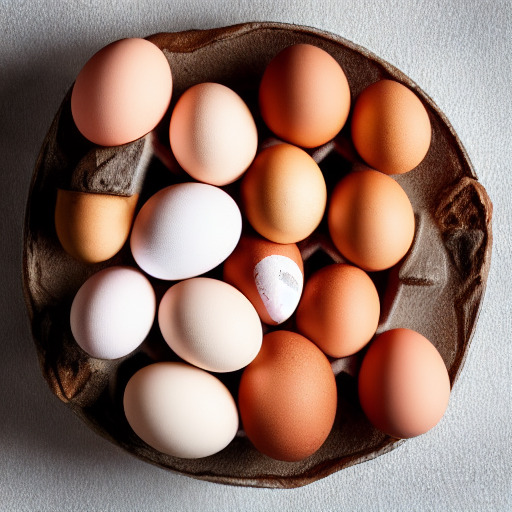}
\label{fig_sd_egg_11}
}

\begin{minipage}{.5\textwidth}
\centering

\bigskip
Attend-and-Excite
\medskip
\end{minipage}

\subfloat[\textit{“a macaron”}]{
\includegraphics[width=.31\columnwidth]{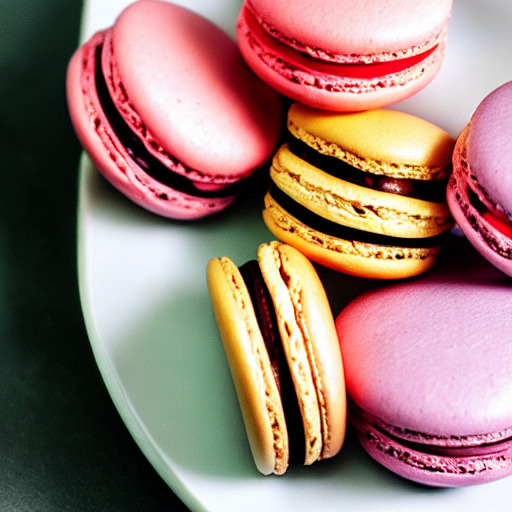}
\label{fig_ae_macaron_1}
}
\subfloat[\textit{“eight macarons”}]{
\includegraphics[width=.31\columnwidth]{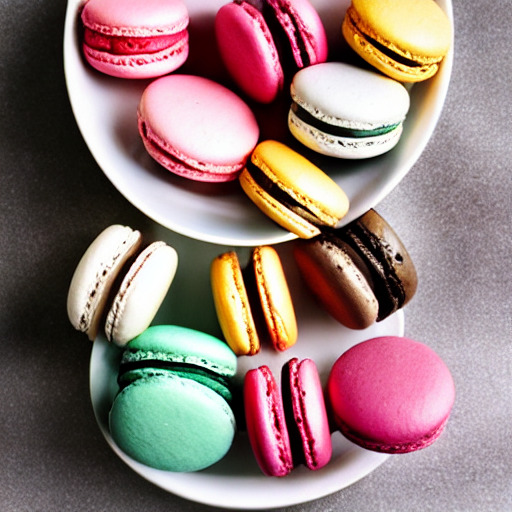}
\label{fig_ae_macaron_8}
}
\subfloat[\textit{“nine macarons”}]{
\includegraphics[width=.31\columnwidth]{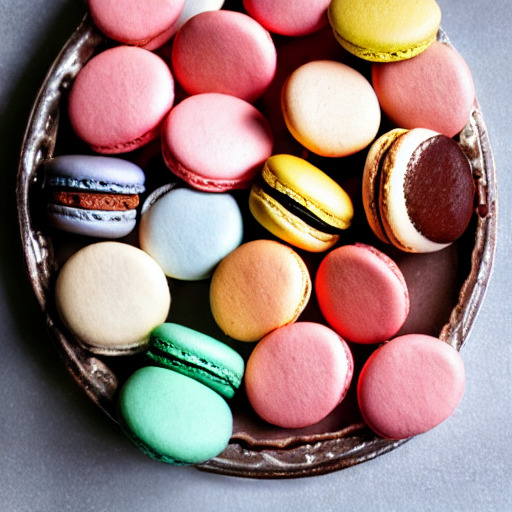}
\label{fig_ae_macaron_9}
}
\subfloat[\textit{“ten macarons”}]{
\includegraphics[width=.31\columnwidth]{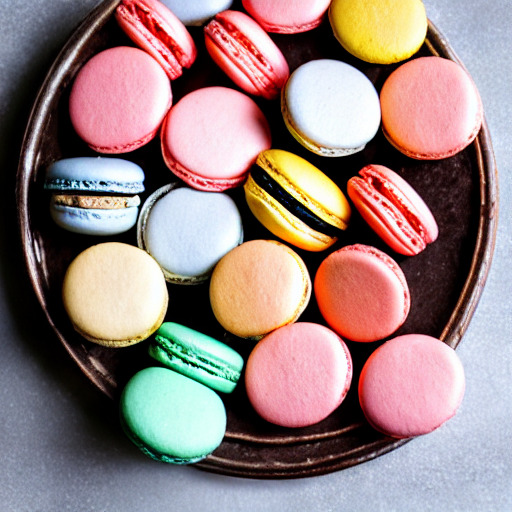}
\label{fig_ae_macaron_10}
}
\subfloat[\textit{“eleven macarons”}]{
\includegraphics[width=.31\columnwidth]{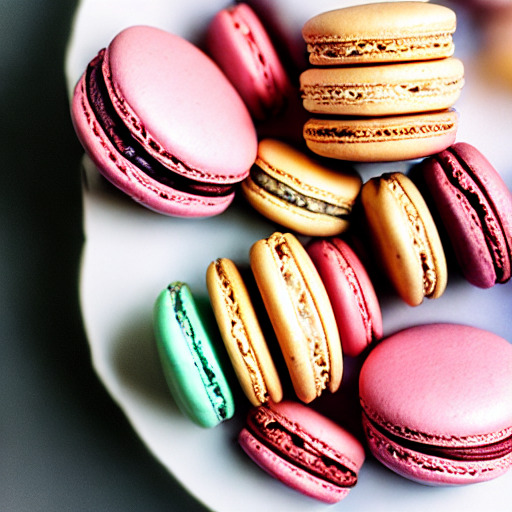}
\label{fig_ae_macaron_11}
}
\subfloat[\textit{“fourteen macarons”}]{
\includegraphics[width=.31\columnwidth]{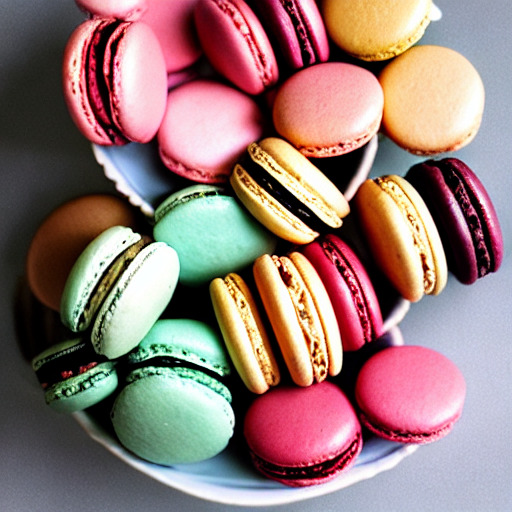}
\label{fig_ae_macaron_14}
}

\subfloat[\textit{“six eggs”}]{
\includegraphics[width=.31\columnwidth]{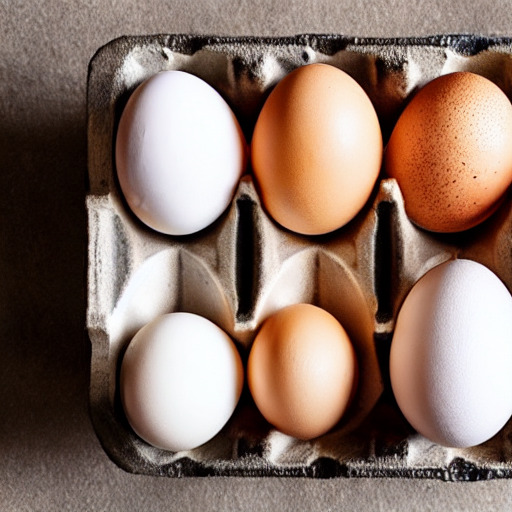}
\label{fig_ae_egg_6}
}
\subfloat[\textit{“seven eggs”}]{
\includegraphics[width=.31\columnwidth]{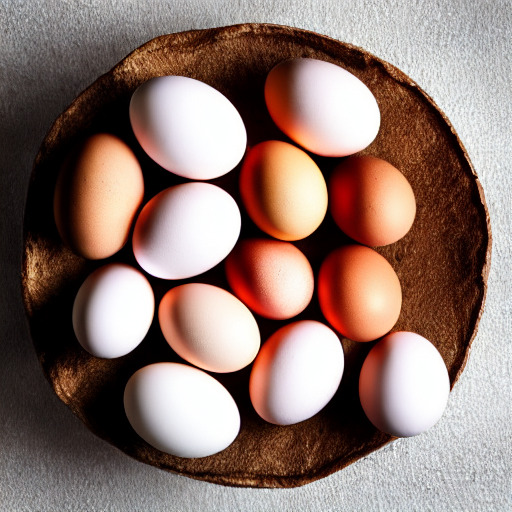}
\label{fig_ae_egg_7}
}
\subfloat[\textit{“eight eggs”}]{
\includegraphics[width=.31\columnwidth]{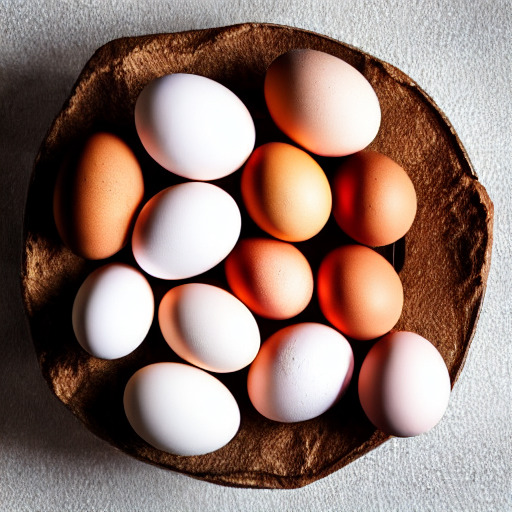}
\label{fig_ae_egg_8}
}
\subfloat[\textit{“nine eggs”}]{
\includegraphics[width=.31\columnwidth]{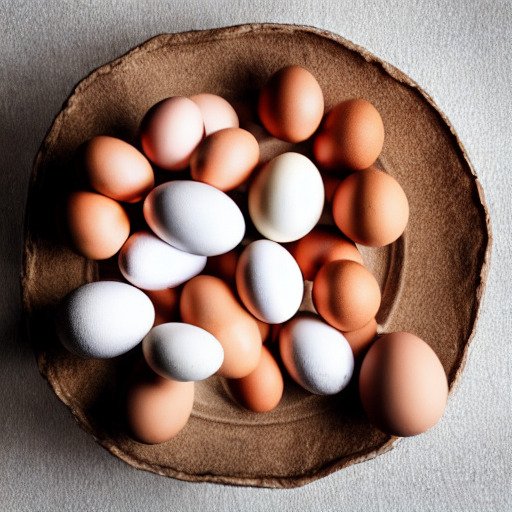}
\label{fig_ae_egg_9}
}
\subfloat[\textit{“ten eggs”}]{
\includegraphics[width=.31\columnwidth]{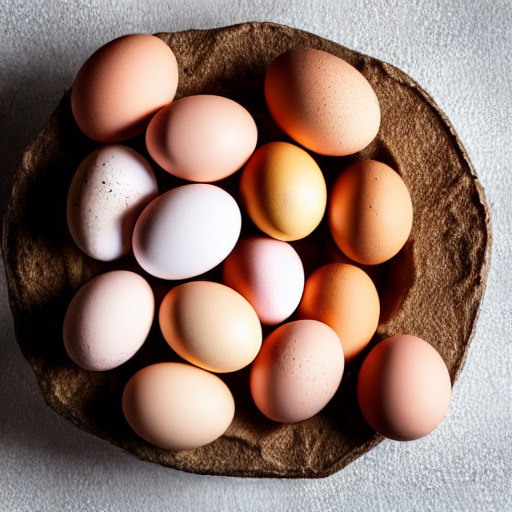}
\label{fig_ae_egg_10}
}
\subfloat[\textit{“eleven eggs”}]{
\includegraphics[width=.31\columnwidth]{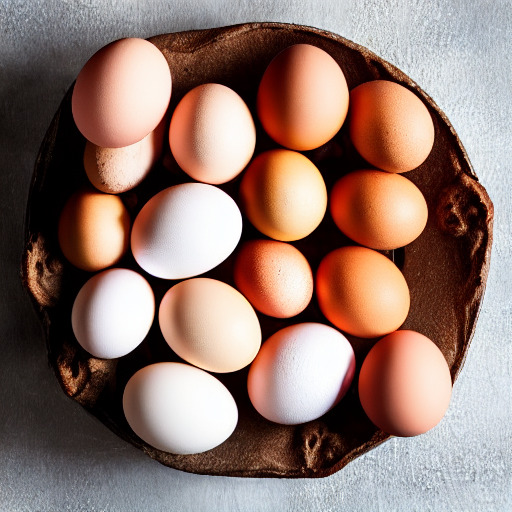}
\label{fig_ae_egg_11}
}

\begin{minipage}{.5\textwidth}
\centering

\bigskip
Ours
\medskip
\end{minipage}

\subfloat[\textit{“a macaron”}]{
\includegraphics[width=.31\columnwidth]{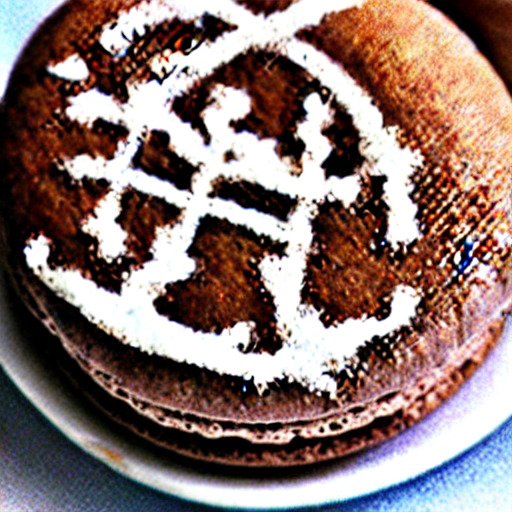}
\label{fig_ours_macaron_1}
}
\subfloat[\textit{“eight macarons”}]{
\includegraphics[width=.31\columnwidth]{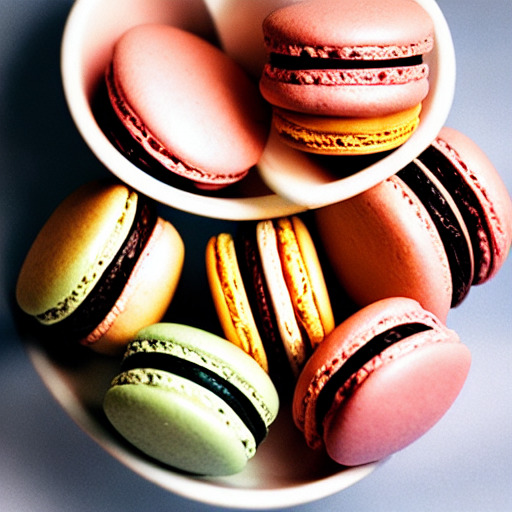}
\label{fig_ours_macaron_8}
}
\subfloat[\textit{“nine macarons”}]{
\includegraphics[width=.31\columnwidth]{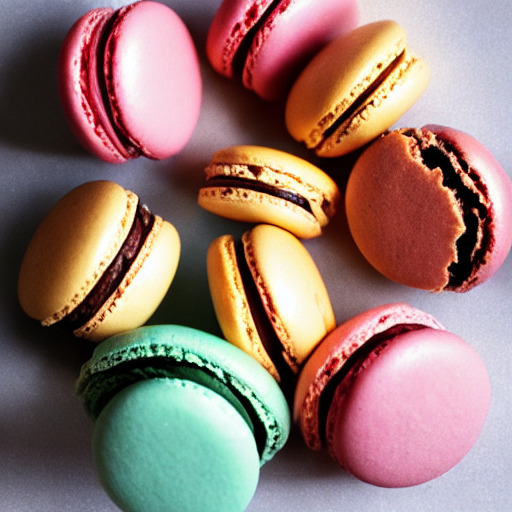}
\label{fig_ours_macaron_9}
}
\subfloat[\textit{“ten macarons”}]{
\includegraphics[width=.31\columnwidth]{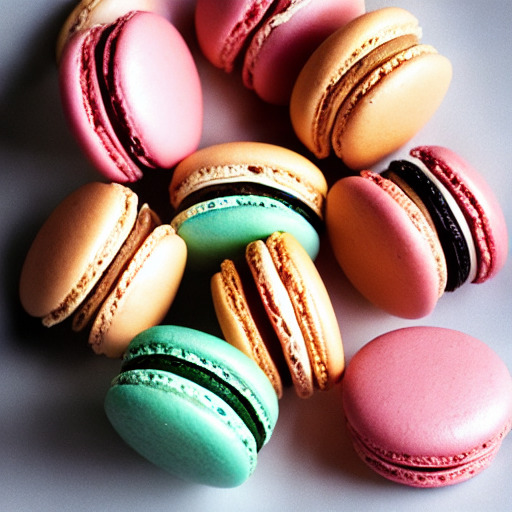}
\label{fig_ours_macaron_10}
}
\subfloat[\textit{“eleven macarons”}]{
\includegraphics[width=.31\columnwidth]{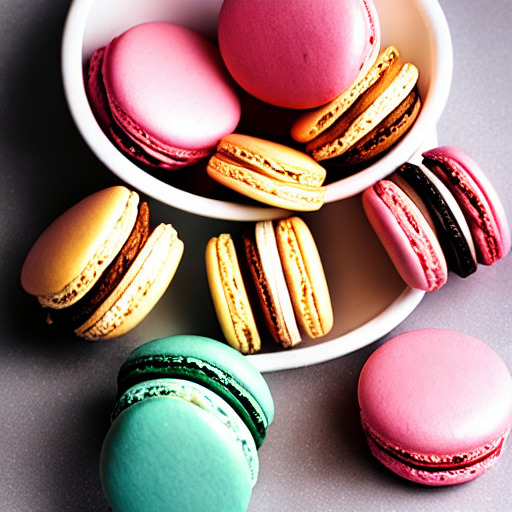}
\label{fig_ours_macaron_11}
}
\subfloat[\textit{“fourteen macarons”}]{
\includegraphics[width=.31\columnwidth]{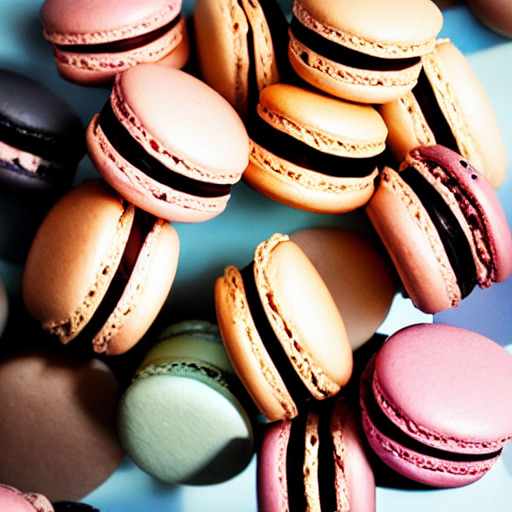}
\label{fig_ours_macaron_14}
}

\subfloat[\textit{“six eggs”}]{
\includegraphics[width=.31\columnwidth]{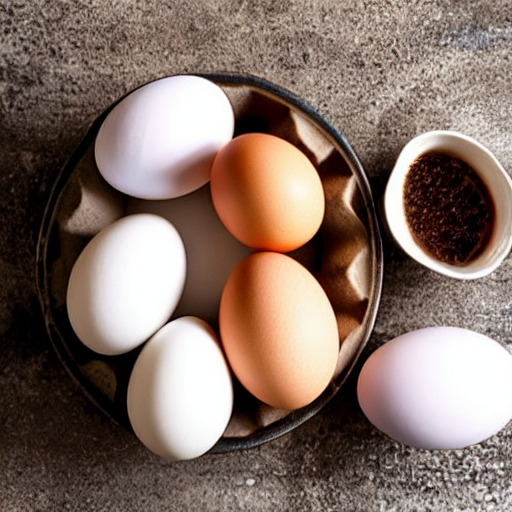}
\label{fig_ours_egg_6}
}
\subfloat[\textit{“seven eggs”}]{
\includegraphics[width=.31\columnwidth]{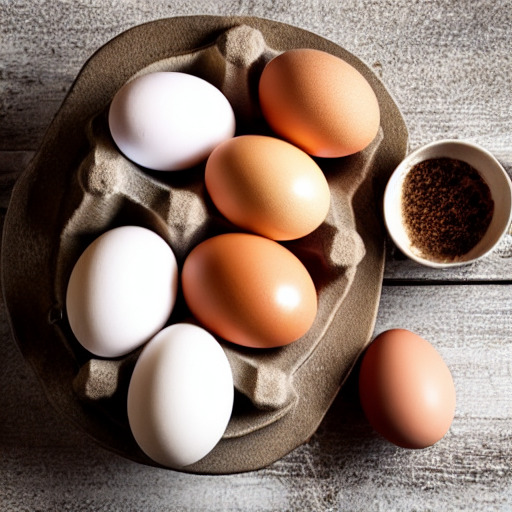}
\label{fig_ours_egg_7}
}
\subfloat[\textit{“eight eggs”}]{
\includegraphics[width=.31\columnwidth]{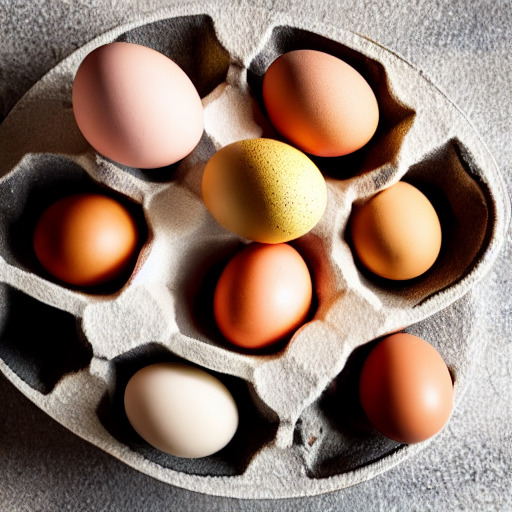}
\label{fig_ours_egg_8}
}
\subfloat[\textit{“nine eggs”}]{
\includegraphics[width=.31\columnwidth]{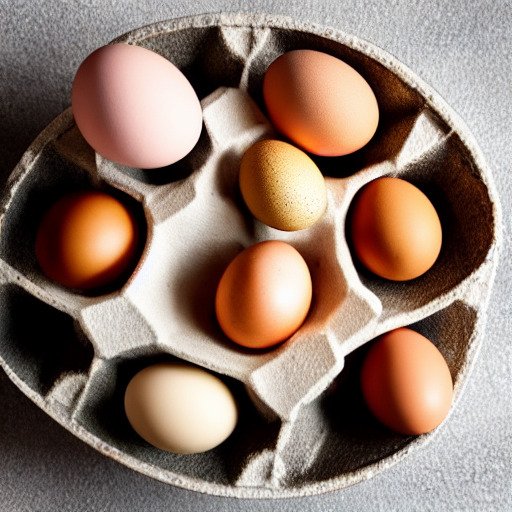}
\label{fig_ours_egg_9}
}
\subfloat[\textit{“ten eggs”}]{
\includegraphics[width=.31\columnwidth]{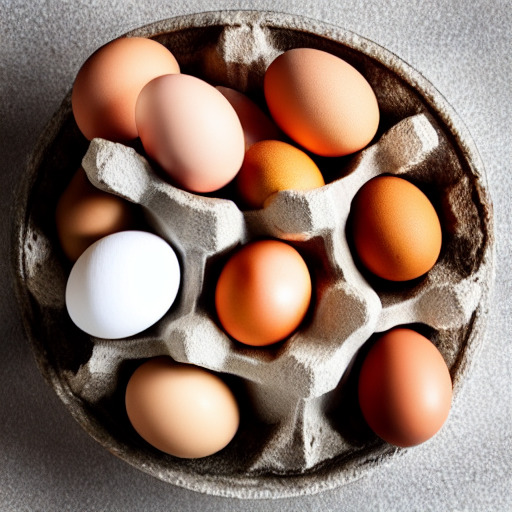}
\label{fig_ours_egg_10}
}
\subfloat[\textit{“eleven eggs”}]{
\includegraphics[width=.31\columnwidth]{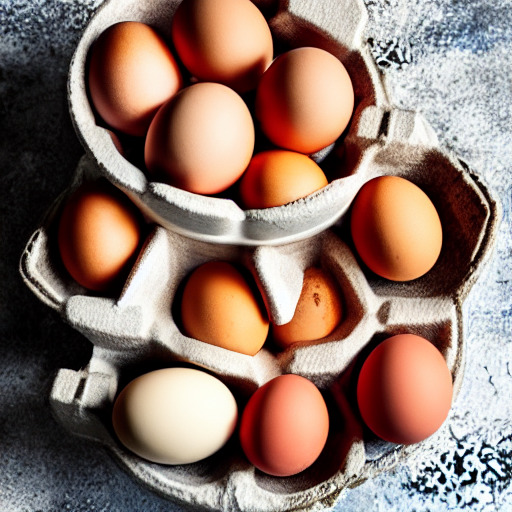}
\label{fig_ours_egg_11}
}

\caption{Additional qualitative results (2)}
\label{fig9}
\end{figure*}

\begin{figure*}[h]
\captionsetup[subfigure]{labelformat=empty}
\centering

\begin{minipage}{.5\textwidth}
\centering
\bigskip
Stable Diffusion
\medskip
\end{minipage}

\subfloat[\textit{“two onions”}]{
\includegraphics[width=.31\columnwidth]{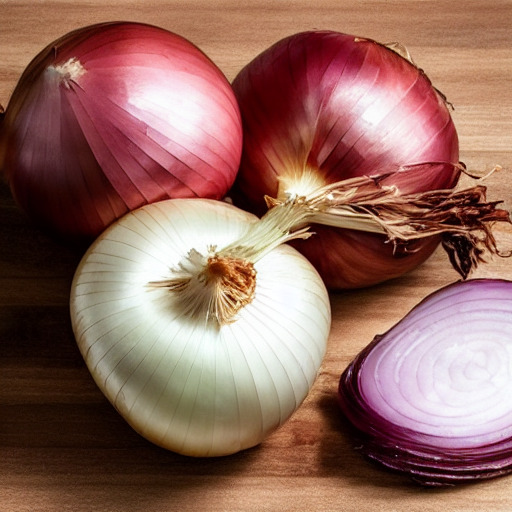}
\label{fig_sd_onion_2}
}
\subfloat[\textit{“three onions”}]{
\includegraphics[width=.31\columnwidth]{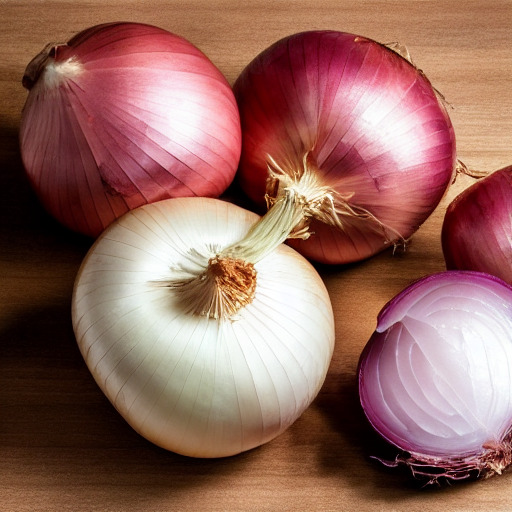}
\label{fig_sd_onion_3}
}
\subfloat[\textit{“six onions”}]{
\includegraphics[width=.31\columnwidth]{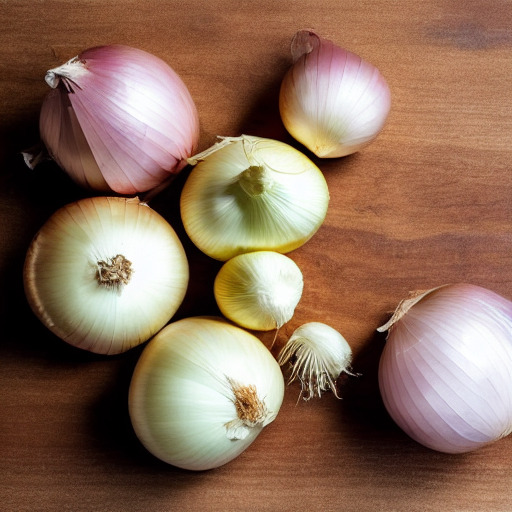}
\label{fig_sd_onion_6}
}
\subfloat[\textit{“eight onions”}]{
\includegraphics[width=.31\columnwidth]{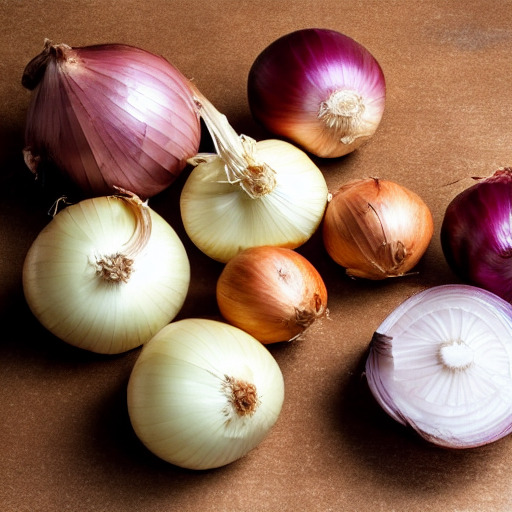}
\label{fig_sd_onion_8}
}
\subfloat[\textit{“nine onions”}]{
\includegraphics[width=.31\columnwidth]{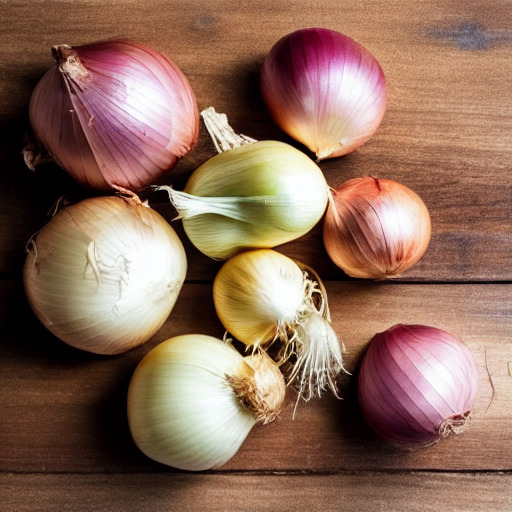}
\label{fig_sd_onion_9}
}
\subfloat[\textit{“eleven onions”}]{
\includegraphics[width=.31\columnwidth]{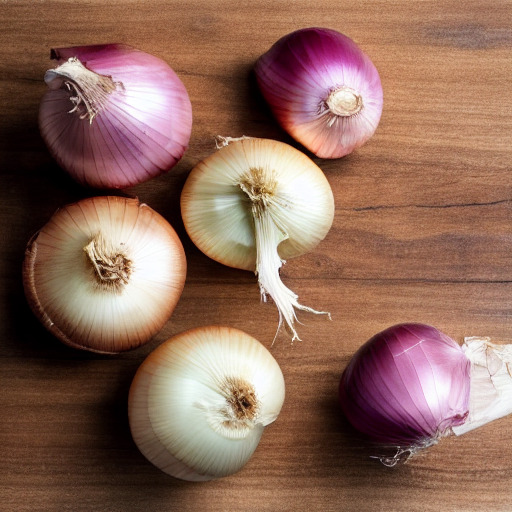}
\label{fig_sd_onion_11}
}

\subfloat[\textit{“a strawberry”}]{
\includegraphics[width=.31\columnwidth]{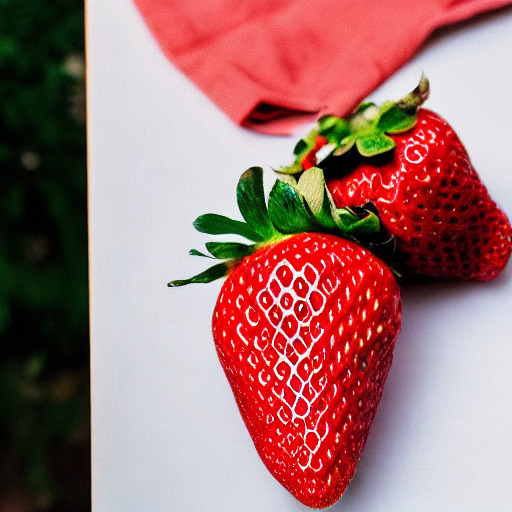}
\label{fig_sd_strawberry_1}
}
\subfloat[\textit{“three strawberries”}]{
\includegraphics[width=.31\columnwidth]{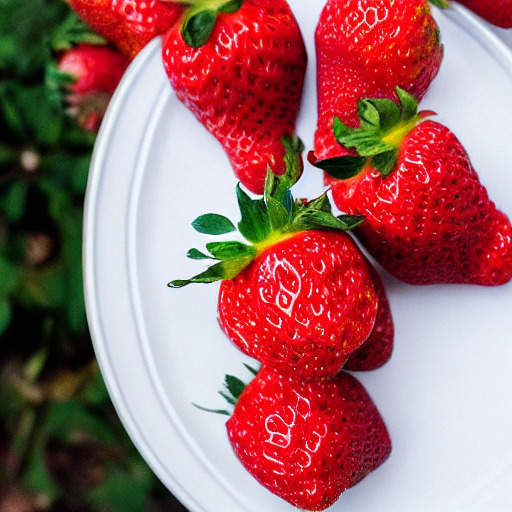}
\label{fig_sd_strawberry_3}
}
\subfloat[\textit{“nine strawberries”}]{
\includegraphics[width=.31\columnwidth]{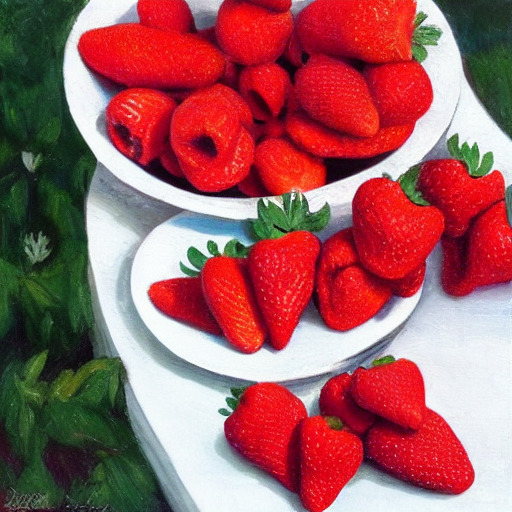}
\label{fig_sd_strawberry_9}
}
\subfloat[\textit{“ten strawberries”}]{
\includegraphics[width=.31\columnwidth]{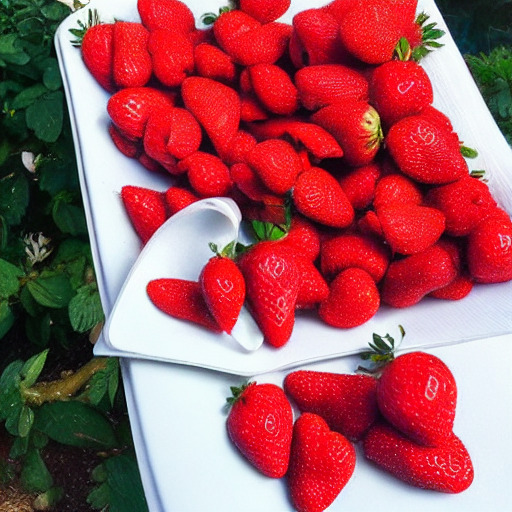}
\label{fig_sd_strawberry_10}
}
\subfloat[\textit{“eleven strawberries”}]{
\includegraphics[width=.31\columnwidth]{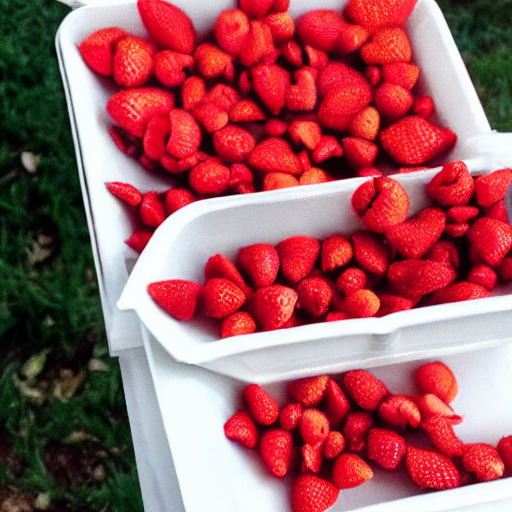}
\label{fig_sd_strawberry_11}
}
\subfloat[\textit{“twelve strawberries”}]{
\includegraphics[width=.31\columnwidth]{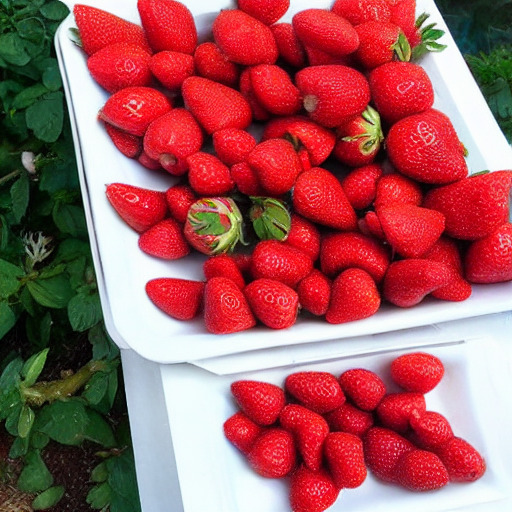}
\label{fig_sd_strawberry_12}
}

\begin{minipage}{.5\textwidth}
\centering

\bigskip
Attend-and-Excite
\medskip
\end{minipage}

\subfloat[\textit{“two onions”}]{
\includegraphics[width=.31\columnwidth]{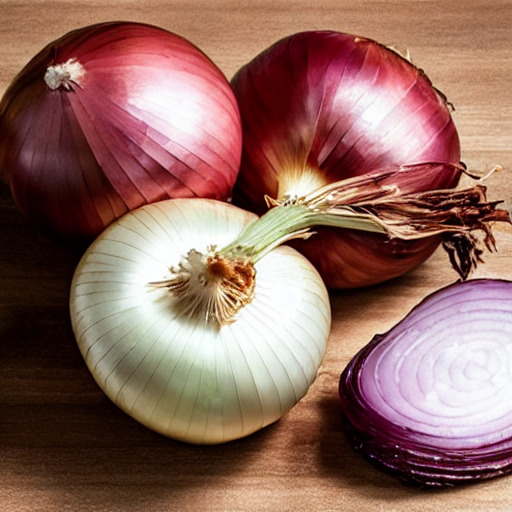}
\label{fig_ae_onion_2}
}
\subfloat[\textit{“three onions”}]{
\includegraphics[width=.31\columnwidth]{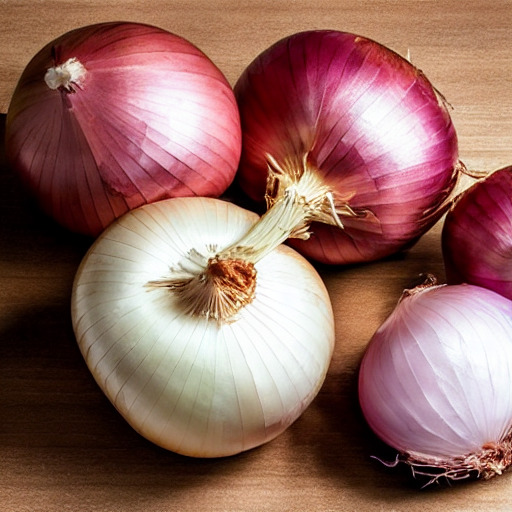}
\label{fig_ae_onion_3}
}
\subfloat[\textit{“six onions”}]{
\includegraphics[width=.31\columnwidth]{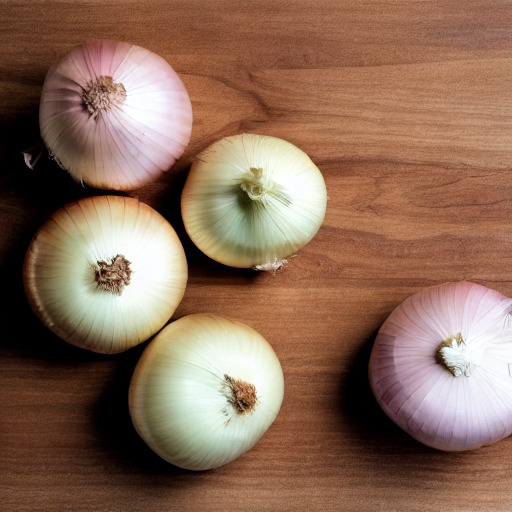}
\label{fig_ae_onion_6}
}
\subfloat[\textit{“eight onions”}]{
\includegraphics[width=.31\columnwidth]{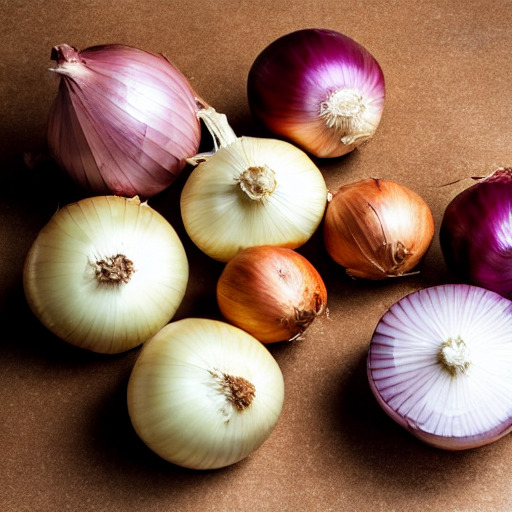}
\label{fig_ae_onion_8}
}
\subfloat[\textit{“nine onions”}]{
\includegraphics[width=.31\columnwidth]{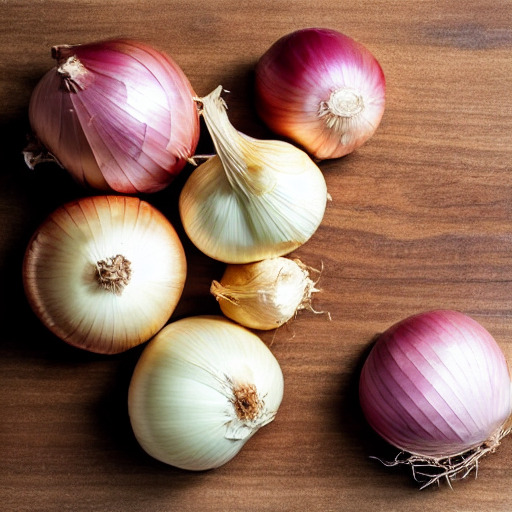}
\label{fig_ae_onion_9}
}
\subfloat[\textit{“eleven onions”}]{
\includegraphics[width=.31\columnwidth]{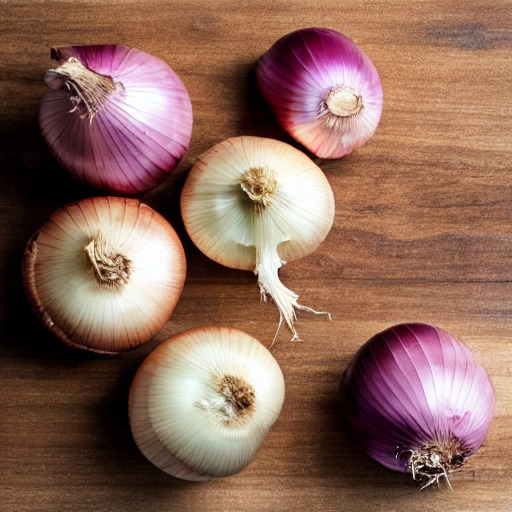}
\label{fig_ae_onion_11}
}

\subfloat[\textit{“a strawberry”}]{
\includegraphics[width=.31\columnwidth]{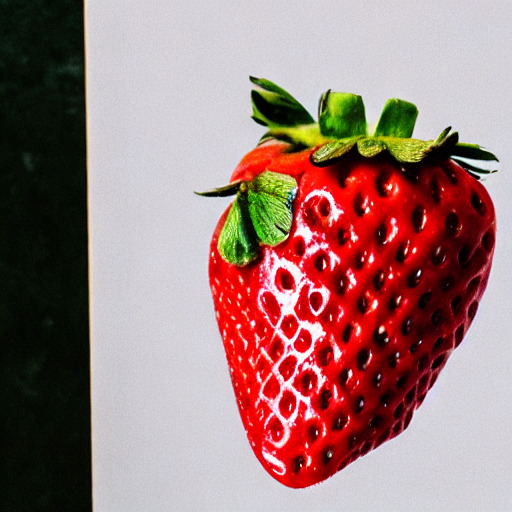}
\label{fig_ae_strawberry_1}
}
\subfloat[\textit{“three strawberries”}]{
\includegraphics[width=.31\columnwidth]{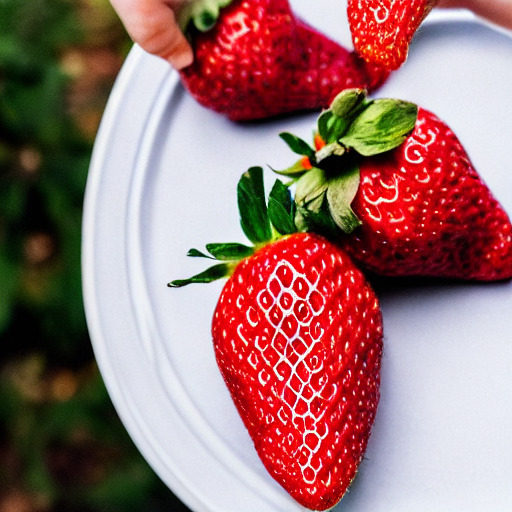}
\label{fig_ae_strawberry_3}
}
\subfloat[\textit{“nine strawberries”}]{
\includegraphics[width=.31\columnwidth]{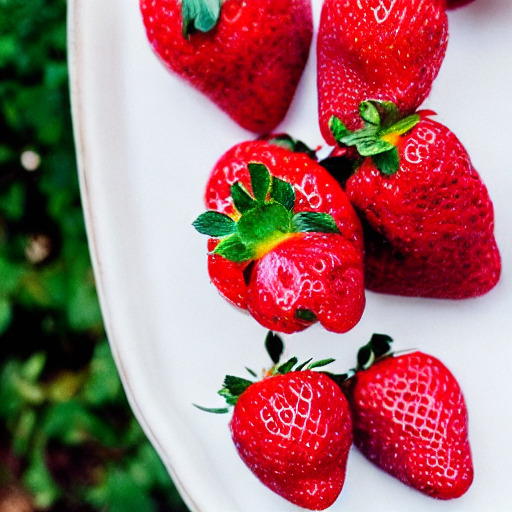}
\label{fig_ae_strawberry_9}
}
\subfloat[\textit{“ten strawberries”}]{
\includegraphics[width=.31\columnwidth]{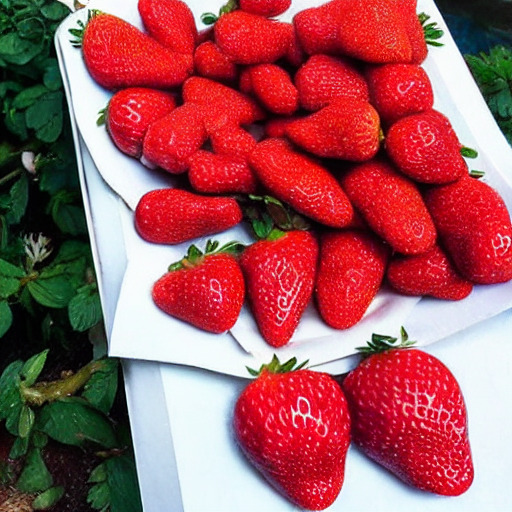}
\label{fig_ae_strawberry_10}
}
\subfloat[\textit{“eleven strawberries”}]{
\includegraphics[width=.31\columnwidth]{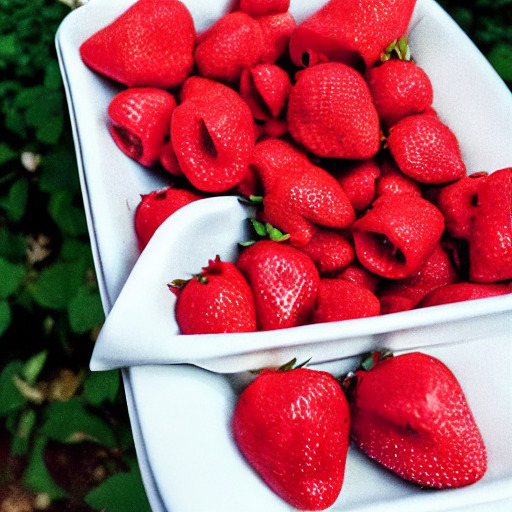}
\label{fig_ae_strawberry_11}
}
\subfloat[\textit{“twelve strawberries”}]{
\includegraphics[width=.31\columnwidth]{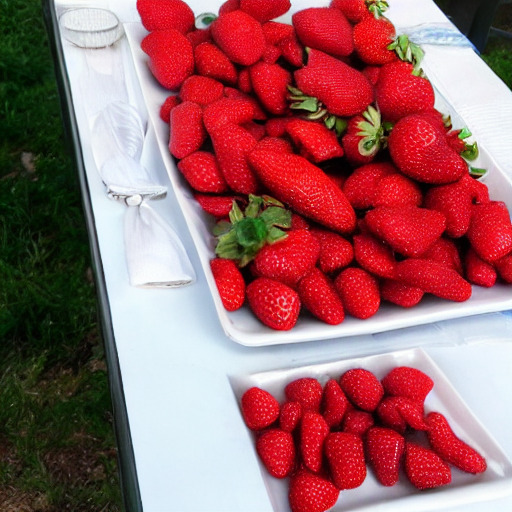}
\label{fig_ae_strawberry_12}
}

\begin{minipage}{.5\textwidth}
\centering

\bigskip
Ours
\medskip
\end{minipage}

\subfloat[\textit{“two onions”}]{
\includegraphics[width=.31\columnwidth]{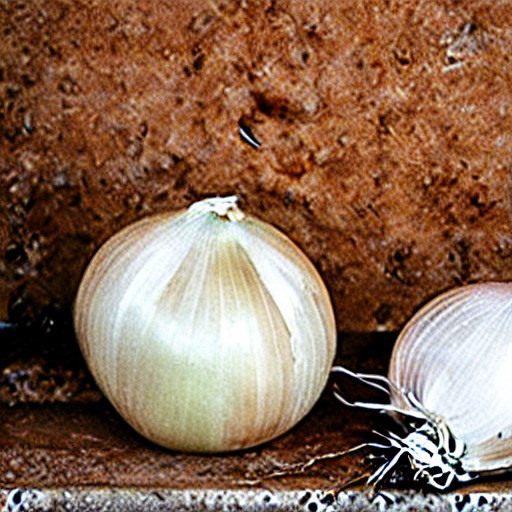}
\label{fig_ours_onion_2}
}
\subfloat[\textit{“three onions”}]{
\includegraphics[width=.31\columnwidth]{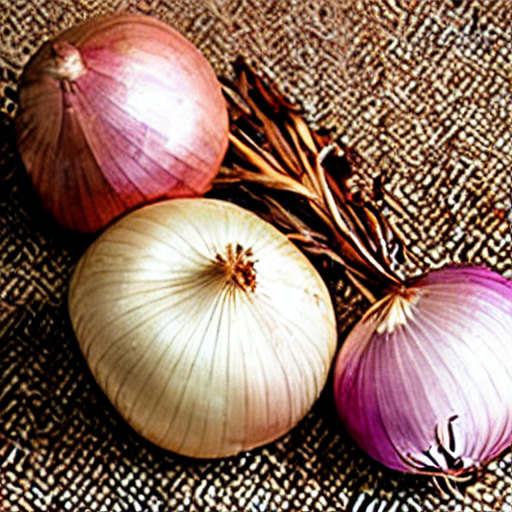}
\label{fig_ours_onion_3}
}
\subfloat[\textit{“six onions”}]{
\includegraphics[width=.31\columnwidth]{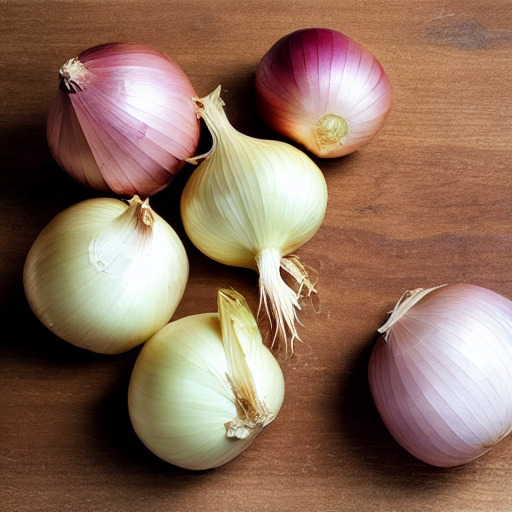}
\label{fig_ours_onion_6}
}
\subfloat[\textit{“eight onions”}]{
\includegraphics[width=.31\columnwidth]{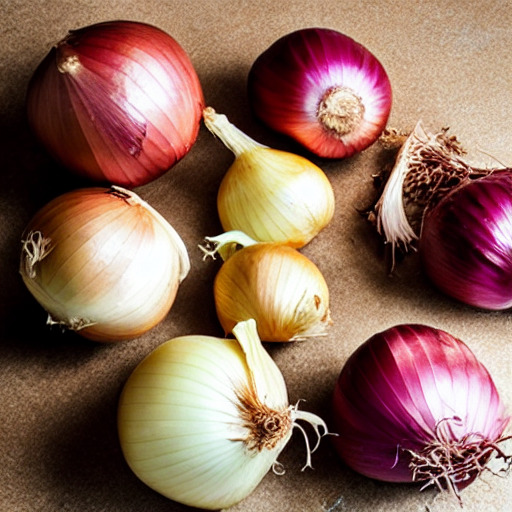}
\label{fig_ours_onion_8}
}
\subfloat[\textit{“nine onions”}]{
\includegraphics[width=.31\columnwidth]{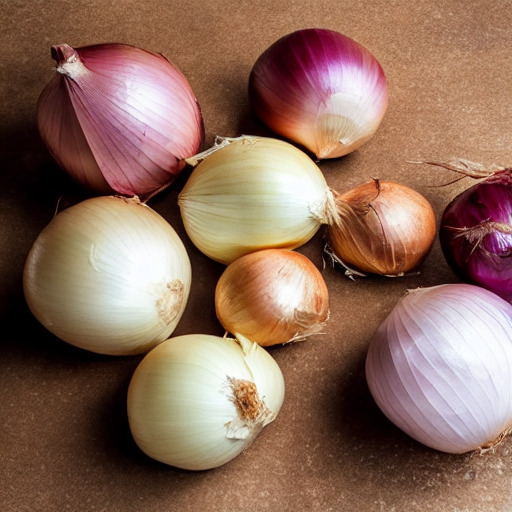}
\label{fig_ours_onion_9}
}
\subfloat[\textit{“eleven onions”}]{
\includegraphics[width=.31\columnwidth]{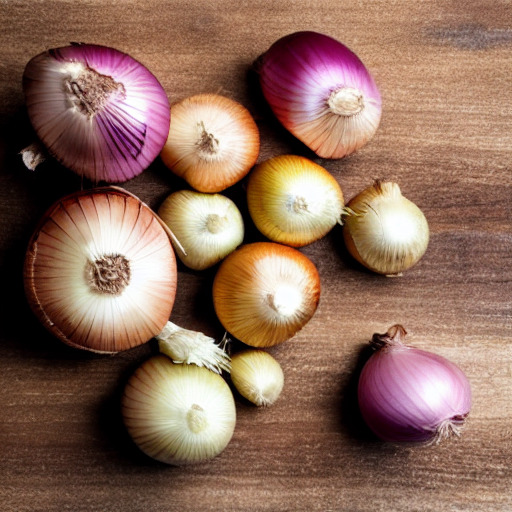}
\label{fig_ours_onion_11}
}

\subfloat[\textit{“a strawberry”}]{
\includegraphics[width=.31\columnwidth]{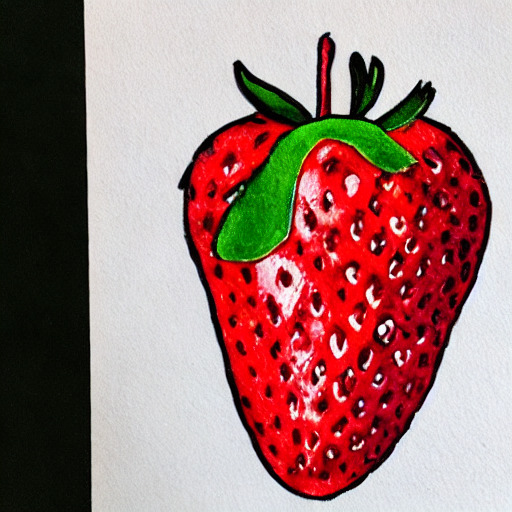}
\label{fig_ours_strawberry_1}
}
\subfloat[\textit{“three strawberries”}]{
\includegraphics[width=.31\columnwidth]{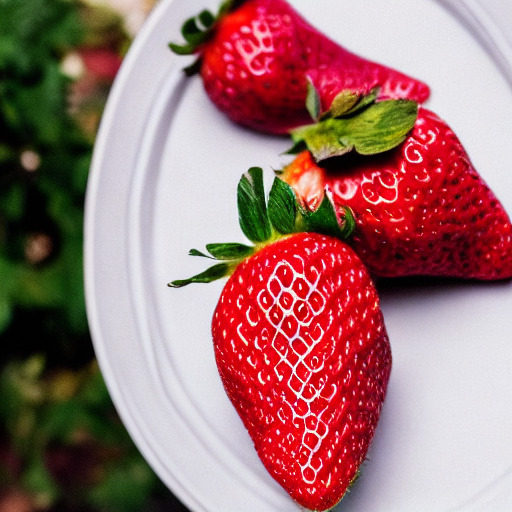}
\label{fig_ours_strawberry_3}
}
\subfloat[\textit{“nine strawberries”}]{
\includegraphics[width=.31\columnwidth]{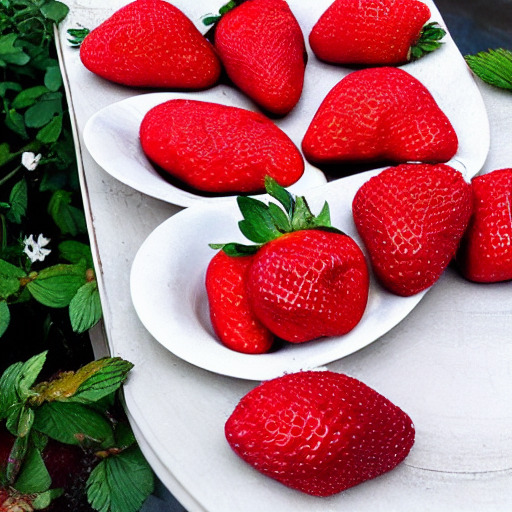}
\label{fig_ours_strawberry_9}
}
\subfloat[\textit{“ten strawberries”}]{
\includegraphics[width=.31\columnwidth]{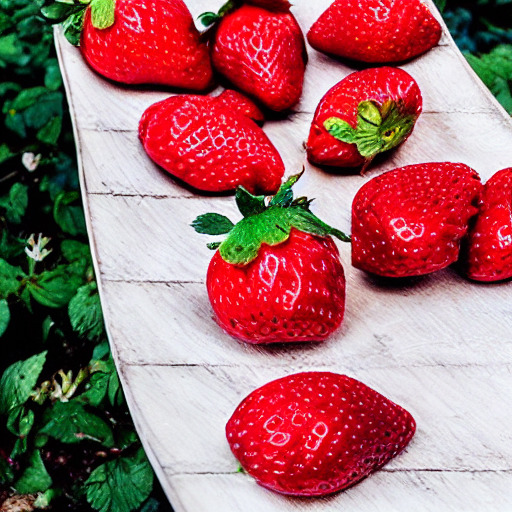}
\label{fig_ours_strawberry_10}
}
\subfloat[\textit{“eleven strawberries”}]{
\includegraphics[width=.31\columnwidth]{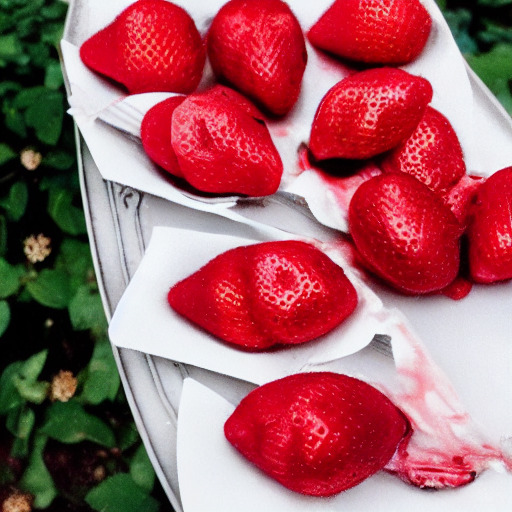}
\label{fig_ours_strawberry_11}
}
\subfloat[\textit{“twelve strawberries”}]{
\includegraphics[width=.31\columnwidth]{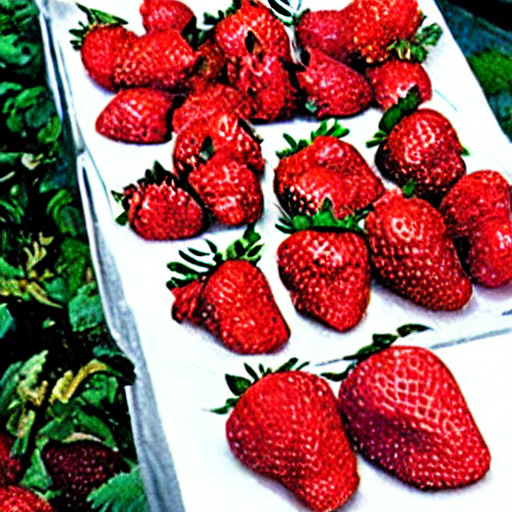}
\label{fig_ours_strawberry_12}
}

\caption{Additional qualitative results (3)}
\label{fig10}
\end{figure*}

\begin{figure*}[t]
\captionsetup[subfigure]{labelformat=empty}
\centering
\begin{minipage}{.45\textwidth}
\centering
\bigskip
Stable Diffusion
\\
\medskip
\textit{“apples and donuts on the table”}
\end{minipage}
\\
\subfloat[generated image]{
\includegraphics[width=.38\columnwidth]{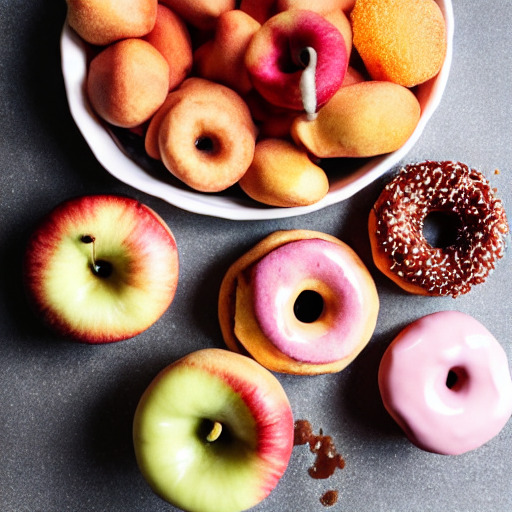}
}
\subfloat[attention map of \textit{“apples”}]{
\includegraphics[width=.38\columnwidth]{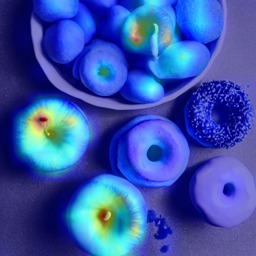}
}
\subfloat[attention map of \textit{“donuts”}]{
\includegraphics[width=.38\columnwidth]{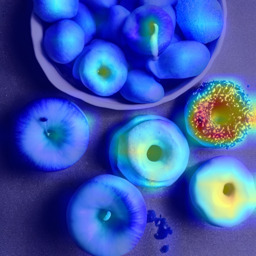}
}
\subfloat[mask of \textit{“apples”}]{
\includegraphics[width=.38\columnwidth]{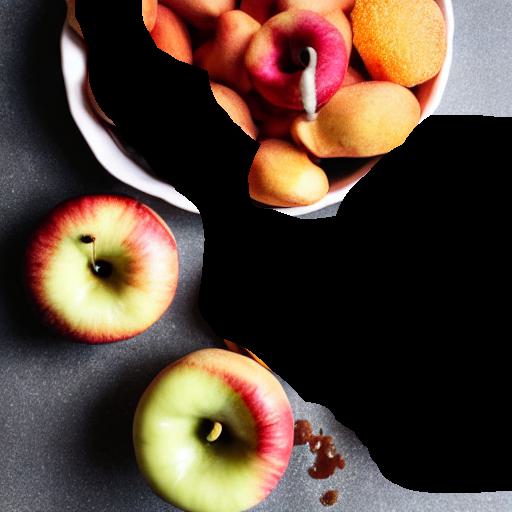}
}
\subfloat[mask of \textit{“donuts”}]{
\includegraphics[width=.38\columnwidth]{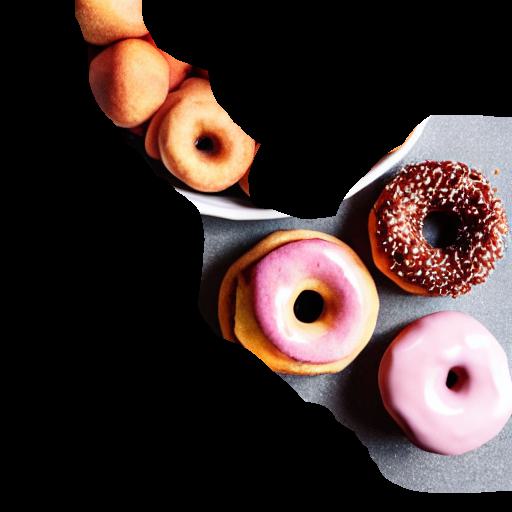}
}

\begin{minipage}{.45\textwidth}
\centering
\bigskip
\textit{“strawberries and eggs on the table”}
\end{minipage}

\subfloat[generated image]{
\includegraphics[width=.38\columnwidth]{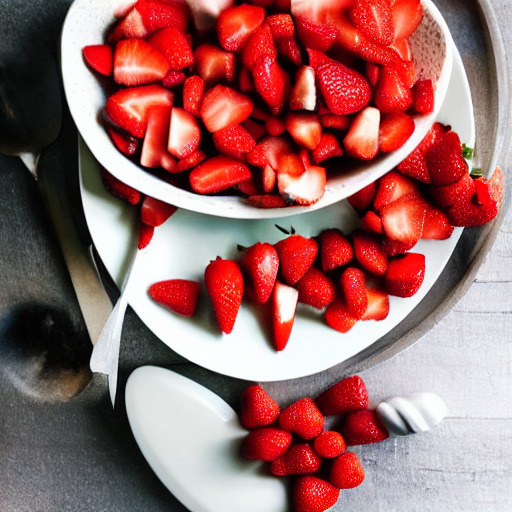}
}
\subfloat[attention map of \textit{“strawberries”}]{
\includegraphics[width=.38\columnwidth]{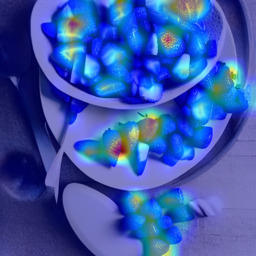}
}
\subfloat[attention map of \textit{“eggs”}]{
\includegraphics[width=.38\columnwidth]{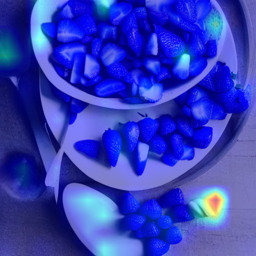}
}
\subfloat[mask of \textit{“strawberries”}]{
\includegraphics[width=.38\columnwidth]{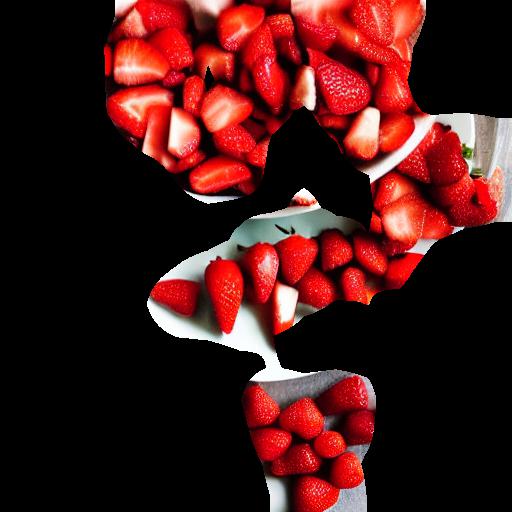}
}
\subfloat[mask of \textit{“eggs”}]{
\includegraphics[width=.38\columnwidth]{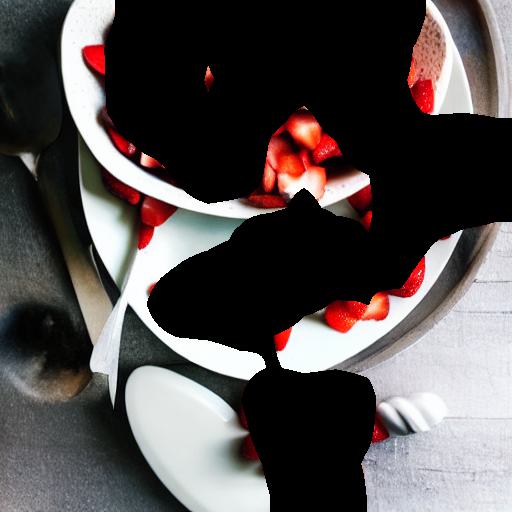}
}

\begin{minipage}{.45\textwidth}
\centering
\bigskip
Ours
\\
\medskip
\textit{“apples and donuts on the table”}
\end{minipage}
\\
\subfloat[generated image]{
\includegraphics[width=.38\columnwidth]{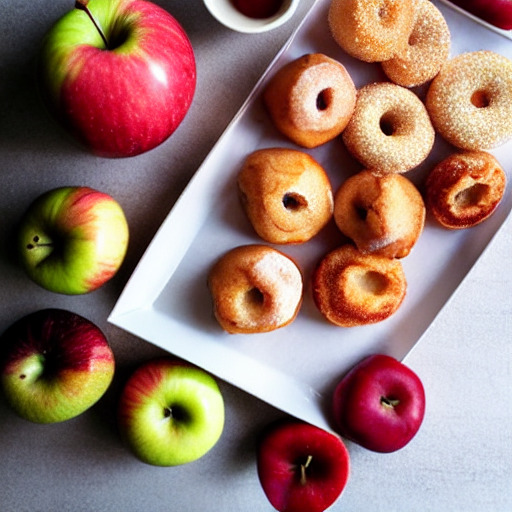}
}
\subfloat[attention map of \textit{“apples”}]{
\includegraphics[width=.38\columnwidth]{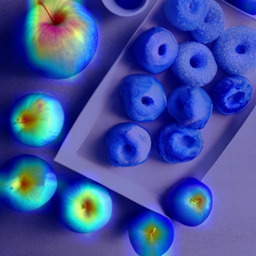}
}
\subfloat[attention map of \textit{“donuts”}]{
\includegraphics[width=.38\columnwidth]{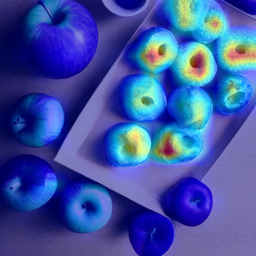}
}
\subfloat[mask of \textit{“apples”}]{
\includegraphics[width=.38\columnwidth]{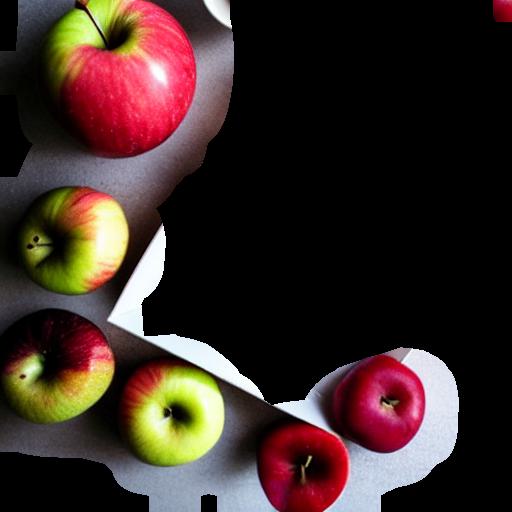}
}
\subfloat[mask of \textit{“donuts”}]{
\includegraphics[width=.38\columnwidth]{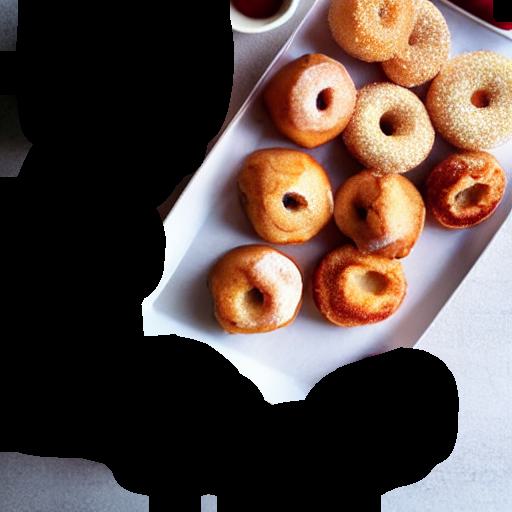}
}

\begin{minipage}{.45\textwidth}
\centering
\bigskip
\textit{“strawberries and eggs on the table”}
\end{minipage}

\subfloat[generated image]{
\includegraphics[width=.38\columnwidth]{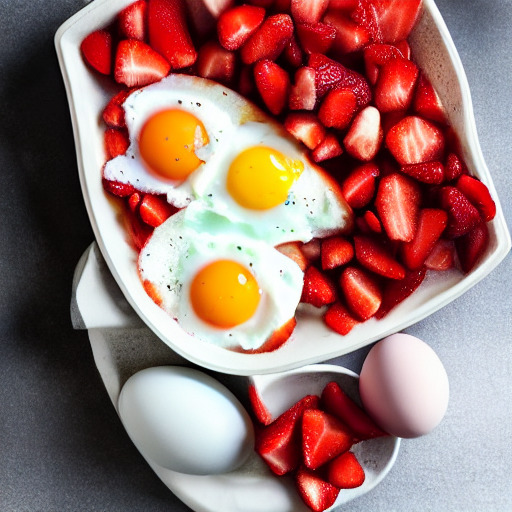}
}
\subfloat[attention map of \textit{“strawberries”}]{
\includegraphics[width=.38\columnwidth]{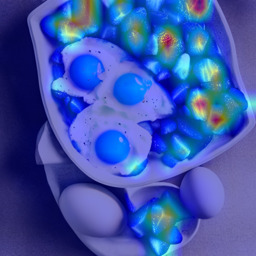}
}
\subfloat[attention map of \textit{“eggs”}]{
\includegraphics[width=.38\columnwidth]{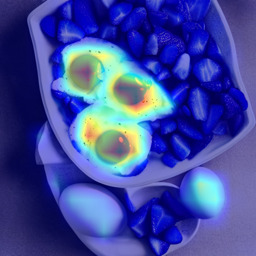}
}
\subfloat[mask of \textit{“strawberries”}]{
\includegraphics[width=.38\columnwidth]{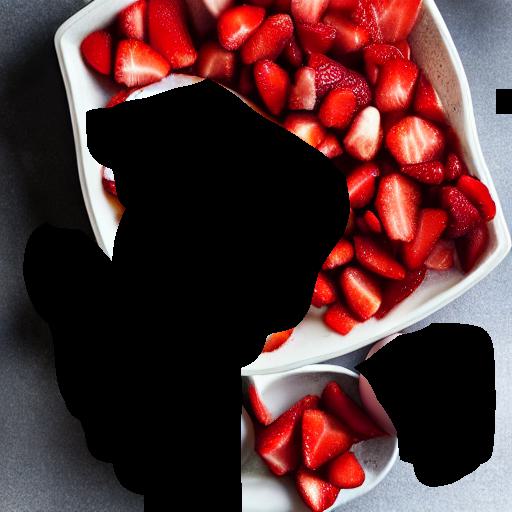}
}
\subfloat[mask of \textit{“eggs”}]{
\includegraphics[width=.38\columnwidth]{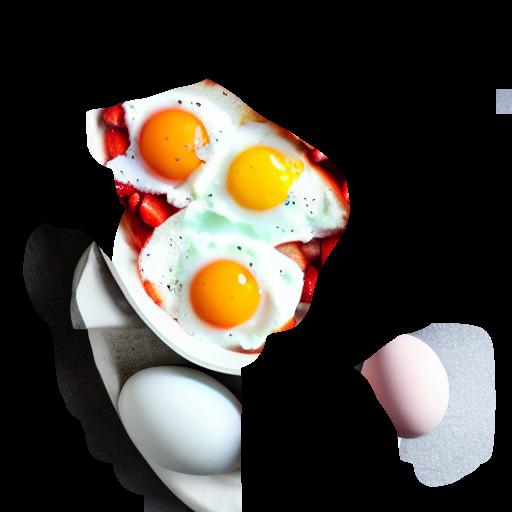}
}

\caption{Additional qualitative results (4)}
\label{fig11}
\end{figure*}

\end{document}